\documentclass{arxiv}

\hypersetup{colorlinks,citecolor=blue,linkcolor=magenta}

\usepackage{times}
\usepackage[american]{babel}

\usepackage{mathtools} 
\usepackage{tikz}

\usepackage{wrapfig}
\usepackage[utf8]{inputenc}
\usepackage[T1]{fontenc}
\usepackage{array,multirow,siunitx,booktabs}
\usepackage{amsfonts}
\usepackage{nicefrac}
\usepackage{microtype}
\usepackage[svgnames]{xcolor}

\usepackage{amssymb}
\newtheorem{assumption}{Assumption}
\usepackage{centernot}
\usepackage{graphicx}
\usepackage{color}
\usepackage{enumitem}
\usepackage{commath}
\usepackage{subcaption}

\DeclareFontFamily{U}{mathx}{\hyphenchar\font45}
\DeclareFontShape{U}{mathx}{m}{n}{
      <5> <6> <7> <8> <9> <10> gen * mathx
      <10.95> mathx10 <12> <14.4> <17.28> <20.74> <24.88> mathx12
      }{}
\DeclareSymbolFont{mathx}{U}{mathx}{m}{n}
\DeclareMathSymbol{\intop}  {1}{mathx}{"B3}

\DeclareFontFamily{U}{mathx}{\hyphenchar\font45}
\DeclareFontShape{U}{mathx}{m}{n}{
      <5> <6> <7> <8> <9> <10>
      <10.95> <12> <14.4> <17.28> <20.74> <24.88>
      mathx10
      }{}
\DeclareSymbolFont{mathx}{U}{mathx}{m}{n}
\DeclareFontSubstitution{U}{mathx}{m}{n}
\DeclareMathAccent{\widecheck}{0}{mathx}{"71}
\DeclareMathAccent{\wideparen}{0}{mathx}{"75}

\newcommand{\wt}{\widetilde}

\let\temp\phi
\let\phi\varphi
\let\varphi\temp

\newcommand{\R}{\mathbb{R}}

\newcommand{\normalN}{\mathcal{N}}

\newcommand{\given}{\,|\,}

\newcommand{\eps}{\varepsilon}

\DeclareMathOperator{\rank}{rank}

\newcommand{\gr}{\mathsf{G}}
\DeclareMathOperator{\pa}{pa}

\newcommand{\AND}{\text{ and }}

\newcommand{\WHERE}{\text{ where }}

\newcommand{\bx}{\mathbf{x}}

\newcommand{\bz}{\mathbf{z}}

\newcommand{\bc}{\mathbf{c}}
\newcommand{\be}{\mathbf{e}}

\newcommand{\beps}{\boldsymbol{\eps}}
\newcommand{\ba}{\mathbf{a}}
\newcommand{\bmu}{\boldsymbol{\mu}}

\newcommand{\lat}{\bz}
\newcommand{\exo}{\beps}
\newcommand{\con}{\bc}
\newcommand{\emb}{\be}
\newcommand{\obs}{\bx}

\newcommand{\width}{w}
\newcommand{\dlat}{k}
\newcommand{\deps}{d}
\newcommand{\dcon}{\deps_{c}}
\newcommand{\demb}{\deps_{e}}
\newcommand{\dobs}{n}

\newcommand{\natom}{n}

\DeclareMathOperator{\enc}{enc}
\DeclareMathOperator{\dec}{dec}

\newcommand{\xto}{\longrightarrow}

\newcommand{\ablation}{CM pooled}
\newcommand{\obscolor}{\tikz[baseline=0ex]{
    \shade[shading=axis, left color=orange, middle color=yellow, right color=green]
    (0,0) rectangle (4em,1.2ex);
  }}
\newcommand{\ivncolor}{\tikz[baseline=0ex]{
    \shade[shading=axis, left color=Aquamarine, middle color=Violet, right color=VioletRed]
    (0,0) rectangle (4em,1.2ex);
  }}

\newcommand{\TableMnistAblation}{
	\begin{table}[htbp]
		\centering
		\caption{Evaluation of the context module, ablations, and baselines in terms of reconstruction and concept learning on MNIST.}
		\label{tab:mnist-ablation}
		\resizebox{\textwidth}{!}{%
			\begin{tabular}{rccccccccc}
\toprule
& \textbf{CM (end-to-end)} & \textbf{CM (fine-tune)} & \textbf{CM (frozen)} & \textbf{base} & \textbf{base pooled} & \textbf{CM pooled} & \textbf{Ada-GVAE} & \textbf{BetaTCVAE} & \textbf{TVAE} \\
\midrule
\multicolumn{ 10 }{c}{\textbf{reconstruction}  (ELBO BPD)} \\
\midrule
\textbf{in-distribution} & 0.259 ± 0.005 & 0.235 ± 0.004 & 0.231 ± 0.003 & \textbf{0.181} ± 0.000 & 0.214 ± 0.003 & 0.228 ± 0.002 & 1.035 ± 0.002 & 1.036 ± 0.012 & 1.037 ± 0.001 \\
\textbf{out-of-distribution} & --- & --- & --- & --- & --- & --- & --- & --- & --- \\
\midrule
\multicolumn{ 10 }{c}{\textbf{concept learning}  (sliced Wasserstein)} \\
\midrule
\textbf{obs} & 0.099 ± 0.006 & 0.085 ± 0.002 & 0.082 ± 0.001 & \textbf{0.067} ± 0.002 & 0.085 ± 0.003 & 0.082 ± 0.003 & 0.521 ± 0.009 & 0.502 ± 0.014 & 0.463 ± 0.020 \\
\textbf{scaled} & \textbf{0.062} ± 0.002 & 0.100 ± 0.007 & 0.109 ± 0.011 & 0.179 ± 0.006 & 0.142 ± 0.010 & 0.165 ± 0.008 & 0.526 ± 0.017 & 0.492 ± 0.012 & 0.490 ± 0.016 \\
\textbf{shear} & 0.119 ± 0.006 & 0.110 ± 0.002 & 0.110 ± 0.003 & \textbf{0.105} ± 0.002 & 0.109 ± 0.002 & 0.109 ± 0.003 & 0.510 ± 0.013 & 0.476 ± 0.012 & 0.488 ± 0.024 \\
\textbf{shift} & 0.113 ± 0.005 & 0.112 ± 0.003 & \textbf{0.105} ± 0.003 & 0.110 ± 0.004 & 0.115 ± 0.002 & 0.116 ± 0.004 & 0.494 ± 0.014 & 0.447 ± 0.012 & 0.501 ± 0.009 \\
\textbf{swel} & 0.126 ± 0.007 & 0.113 ± 0.005 & 0.115 ± 0.002 & \textbf{0.102} ± 0.003 & 0.129 ± 0.009 & 0.114 ± 0.003 & 0.473 ± 0.011 & 0.492 ± 0.007 & 0.481 ± 0.011 \\
\textbf{thic} & \textbf{0.178} ± 0.007 & 0.184 ± 0.002 & 0.193 ± 0.010 & 0.196 ± 0.008 & 0.231 ± 0.014 & 0.207 ± 0.014 & 0.474 ± 0.008 & 0.475 ± 0.012 & 0.472 ± 0.013 \\
\textbf{thin} & \textbf{0.064} ± 0.004 & 0.080 ± 0.006 & 0.092 ± 0.010 & 0.133 ± 0.011 & 0.107 ± 0.006 & 0.126 ± 0.007 & 0.501 ± 0.014 & 0.506 ± 0.013 & 0.504 ± 0.011 \\
\addlinespace
\textbf{concept learning mean} & \textbf{0.109} ± 0.006 & 0.112 ± 0.004 & 0.115 ± 0.006 & 0.127 ± 0.005 & 0.131 ± 0.007 & 0.131 ± 0.006 & 0.500 ± 0.012 & 0.484 ± 0.012 & 0.486 ± 0.015 \\
\midrule
\bottomrule
\end{tabular}
		}
	\end{table}
}

\newcommand{\TableMnistCapacityTwo}{
	\begin{table}[htbp]
		\centering
		\caption{Evaluation of CM (end-to-end) in terms of reconstruction and concept learning on MNIST for different capacities, with depth $h_{exp} = 2$ as $(w_{exp}, w_{c})$ varies.}
		\label{tab:mnist-capacity-2}
		\resizebox{\textwidth}{!}{%
			\begin{tabular}{rcccccc}
	\toprule
	$(w_{exp}, w_{c})$             & \textbf{(15, 22)} & \textbf{(15, 5)}       & \textbf{(22, 22)}      & \textbf{(22, 5)}       & \textbf{(50, 22)}      & \textbf{(50, 5)} \\
	\midrule
	\multicolumn{ 7 }{c}{\textbf{reconstruction}  (ELBO BPD)}                                                                                                                 \\
	\midrule
	\textbf{in-distribution}       & 0.260 ± 0.005     & 0.259 ± 0.005          & 0.255 ± 0.003          & \textbf{0.252} ± 0.003 & 0.265 ± 0.001          & 0.265 ± 0.006    \\
	\textbf{out-of-distribution}   & ---               & ---                    & ---                    & ---                    & ---                    & ---              \\
	\midrule
	\multicolumn{ 7 }{c}{\textbf{concept learning}  (sliced Wasserstein)}                                                                                                     \\
	\midrule
	\textbf{obs}                   & 0.087 ± 0.003     & 0.099 ± 0.006          & \textbf{0.082} ± 0.005 & 0.095 ± 0.006          & 0.086 ± 0.004          & 0.092 ± 0.008    \\
	\textbf{scaled}                & 0.073 ± 0.011     & \textbf{0.062} ± 0.002 & 0.065 ± 0.004          & 0.067 ± 0.005          & 0.068 ± 0.003          & 0.072 ± 0.004    \\
	\textbf{shear}                 & 0.115 ± 0.006     & 0.119 ± 0.006          & \textbf{0.108} ± 0.006 & 0.122 ± 0.006          & 0.116 ± 0.004          & 0.118 ± 0.004    \\
	\textbf{shift}                 & 0.106 ± 0.009     & 0.113 ± 0.005          & 0.108 ± 0.008          & 0.111 ± 0.002          & \textbf{0.096} ± 0.003 & 0.115 ± 0.005    \\
	\textbf{swel}                  & 0.119 ± 0.006     & 0.126 ± 0.007          & \textbf{0.107} ± 0.005 & 0.118 ± 0.006          & 0.111 ± 0.007          & 0.118 ± 0.007    \\
	\textbf{thic}                  & 0.202 ± 0.016     & 0.178 ± 0.007          & 0.180 ± 0.007          & \textbf{0.171} ± 0.014 & 0.176 ± 0.010          & 0.178 ± 0.015    \\
	\textbf{thin}                  & 0.065 ± 0.006     & \textbf{0.064} ± 0.004 & 0.069 ± 0.005          & 0.072 ± 0.010          & 0.073 ± 0.007          & 0.067 ± 0.006    \\
	\addlinespace
	\textbf{concept learning mean} & 0.110 ± 0.008     & 0.109 ± 0.006          & \textbf{0.103} ± 0.006 & 0.108 ± 0.007          & 0.104 ± 0.005          & 0.109 ± 0.007    \\
	\midrule
	\bottomrule
\end{tabular}

		}
	\end{table}
}

\newcommand{\TableMnistCapacityFour}{
	\begin{table}[htbp]
		\centering
		\caption{Evaluation of CM (end-to-end) in terms of reconstruction and concept learning on MNIST for different capacities, with depth $h_{exp} = 4$ as $(w_{exp}, w_{c})$ varies.}
		\label{tab:mnist-capacity-4}
		\resizebox{\textwidth}{!}{%
			\begin{tabular}{rcccccc}
	\toprule
	$(w_{exp}, w_{c})$             & \textbf{(15, 22)}      & \textbf{(15, 5)}       & \textbf{(22, 22)}      & \textbf{(22, 5)}       & \textbf{(50, 22)} & \textbf{(50, 5)}       \\
	\midrule
	\multicolumn{ 7 }{c}{\textbf{reconstruction}  (ELBO BPD)}                                                                                                                       \\
	\midrule
	\textbf{in-distribution}       & 0.269 ± 0.003          & 0.272 ± 0.006          & 0.276 ± 0.004          & \textbf{0.255} ± 0.004 & 0.281 ± 0.005     & 0.279 ± 0.008          \\
	\textbf{out-of-distribution}   & ---                    & ---                    & ---                    & ---                    & ---               & ---                    \\
	\midrule
	\multicolumn{ 7 }{c}{\textbf{concept learning}  (sliced Wasserstein)}                                                                                                           \\
	\midrule
	\textbf{obs}                   & 0.096 ± 0.004          & \textbf{0.088} ± 0.004 & 0.089 ± 0.001          & 0.092 ± 0.004          & 0.090 ± 0.005     & 0.093 ± 0.003          \\
	\textbf{scaled}                & \textbf{0.063} ± 0.003 & 0.071 ± 0.004          & 0.071 ± 0.003          & 0.085 ± 0.018          & 0.067 ± 0.001     & 0.079 ± 0.003          \\
	\textbf{shear}                 & 0.119 ± 0.006          & 0.110 ± 0.004          & 0.116 ± 0.004          & 0.113 ± 0.004          & 0.107 ± 0.004     & \textbf{0.103} ± 0.004 \\
	\textbf{shift}                 & 0.114 ± 0.004          & 0.109 ± 0.006          & 0.108 ± 0.005          & 0.107 ± 0.005          & 0.113 ± 0.006     & \textbf{0.104} ± 0.002 \\
	\textbf{swel}                  & 0.120 ± 0.006          & 0.113 ± 0.007          & \textbf{0.110} ± 0.001 & 0.122 ± 0.006          & 0.116 ± 0.005     & 0.115 ± 0.003          \\
	\textbf{thic}                  & 0.179 ± 0.012          & 0.178 ± 0.014          & \textbf{0.175} ± 0.011 & 0.185 ± 0.011          & 0.188 ± 0.007     & 0.184 ± 0.011          \\
	\textbf{thin}                  & \textbf{0.066} ± 0.005 & 0.076 ± 0.004          & 0.072 ± 0.003          & 0.070 ± 0.003          & 0.076 ± 0.007     & 0.083 ± 0.006          \\
	\addlinespace
	\textbf{concept learning mean} & 0.108 ± 0.005          & 0.106 ± 0.006          & \textbf{0.106} ± 0.004 & 0.111 ± 0.007          & 0.108 ± 0.005     & 0.109 ± 0.005          \\
	\midrule
	\bottomrule
\end{tabular}

		}
	\end{table}
}

\newcommand{\TableMnistBeta}{
	\begin{table}[htbp]
		\centering
		\caption{Evaluation of CM (end-to-end) in terms of reconstruction and concept learning on MNIST for different latent space regularization strengths, as $\beta$ varies.}
		\label{tab:mnist-beta}
		\begin{tabular}{rcccc}
\toprule
& \textbf{0.5} & \textbf{1.0} & \textbf{2.0} & \textbf{4.0} \\
\midrule
\multicolumn{ 5 }{c}{\textbf{reconstruction}  (ELBO BPD)} \\
\midrule
\textbf{in-distribution} & \textbf{0.222} ± 0.002 & 0.259 ± 0.005 & 0.298 ± 0.003 & 0.341 ± 0.007 \\
\textbf{out-of-distribution} & --- & --- & --- & --- \\
\midrule
\multicolumn{ 5 }{c}{\textbf{concept learning}  (sliced Wasserstein)} \\
\midrule
\textbf{obs} & \textbf{0.085} ± 0.003 & 0.099 ± 0.006 & 0.102 ± 0.003 & 0.108 ± 0.005 \\
\textbf{scaled} & 0.077 ± 0.006 & \textbf{0.062} ± 0.002 & 0.077 ± 0.004 & 0.098 ± 0.009 \\
\textbf{shear} & \textbf{0.108} ± 0.003 & 0.119 ± 0.006 & 0.125 ± 0.005 & 0.117 ± 0.004 \\
\textbf{shift} & \textbf{0.105} ± 0.003 & 0.113 ± 0.005 & 0.125 ± 0.004 & 0.130 ± 0.007 \\
\textbf{swel} & \textbf{0.110} ± 0.007 & 0.126 ± 0.007 & 0.129 ± 0.006 & 0.134 ± 0.004 \\
\textbf{thic} & 0.182 ± 0.011 & \textbf{0.178} ± 0.007 & 0.190 ± 0.016 & 0.192 ± 0.008 \\
\textbf{thin} & 0.069 ± 0.006 & \textbf{0.064} ± 0.004 & 0.074 ± 0.005 & 0.117 ± 0.005 \\
\addlinespace
\textbf{concept learning mean} & \textbf{0.105} ± 0.006 & 0.109 ± 0.006 & 0.118 ± 0.006 & 0.128 ± 0.006 \\
\midrule
\bottomrule
\end{tabular}
	\end{table}
}

\newcommand{\TableMnistGroupNorm}{
	\begin{table}[htbp]
		\centering
		\caption{Evaluation of CM (end-to-end) in terms of reconstruction and concept learning on MNIST for different group lasso regularization strengths.}
		\label{tab:mnist-groupnorm}
		\begin{tabular}{rcccc}
	\toprule
	                               & \(\lambda =\) \textbf{0} & \(\lambda =\)\textbf{1} & \(\lambda =\)\textbf{10} & \(\lambda =\)\textbf{100} \\
	\midrule
	\multicolumn{ 5 }{c}{\textbf{reconstruction}  (ELBO BPD)}                                                                                  \\
	\midrule
	\textbf{in-distribution}       & 0.259 ± 0.005            & 0.259 ± 0.003           & 0.253 ± 0.003            & \textbf{0.248} ± 0.002    \\
	\textbf{out-of-distribution}   & ---                      & ---                     & ---                      & ---                       \\
	\midrule
	\multicolumn{ 5 }{c}{\textbf{concept learning}  (sliced Wasserstein)}                                                                      \\
	\midrule
	\textbf{obs}                   & 0.099 ± 0.006            & 0.092 ± 0.005           & \textbf{0.091} ± 0.006   & 0.092 ± 0.006             \\
	\textbf{scaled}                & \textbf{0.062} ± 0.002   & 0.069 ± 0.002           & 0.075 ± 0.003            & 0.118 ± 0.006             \\
	\textbf{shear}                 & 0.119 ± 0.006            & 0.118 ± 0.005           & 0.115 ± 0.008            & \textbf{0.113} ± 0.004    \\
	\textbf{shift}                 & 0.113 ± 0.005            & \textbf{0.112} ± 0.007  & 0.116 ± 0.005            & 0.113 ± 0.003             \\
	\textbf{swel}                  & 0.126 ± 0.007            & \textbf{0.122} ± 0.004  & 0.131 ± 0.007            & 0.127 ± 0.006             \\
	\textbf{thic}                  & \textbf{0.178} ± 0.007   & 0.183 ± 0.008           & 0.216 ± 0.015            & 0.220 ± 0.005             \\
	\textbf{thin}                  & \textbf{0.064} ± 0.004   & 0.072 ± 0.005           & 0.082 ± 0.009            & 0.111 ± 0.008             \\
	\addlinespace
	\textbf{concept learning mean} & \textbf{0.109} ± 0.006   & 0.110 ± 0.005           & 0.118 ± 0.008            & 0.128 ± 0.005             \\
	\midrule
	\bottomrule
\end{tabular}

	\end{table}
}

\newcommand{\TableMnistEllTwoNorm}{
	\begin{table}[htbp]
		\centering
		\caption{Evaluation of CM (end-to-end) in terms of reconstruction and concept learning on MNIST for different \(L_2\)-norm regularization strengths.}
		\label{tab:mnist-l2norm}
		\begin{tabular}{rcccc}
	\toprule
	                               & \(\lambda =\) \textbf{0} & \(\lambda =\)\textbf{1} & \(\lambda =\)\textbf{10} & \(\lambda =\)\textbf{100} \\
	\midrule
	\multicolumn{ 5 }{c}{\textbf{reconstruction}  (ELBO BPD)}                                                                                  \\
	\midrule
	\textbf{in-distribution}       & 0.259 ± 0.005            & 0.254 ± 0.003           & 0.255 ± 0.004            & \textbf{0.253} ± 0.001    \\
	\textbf{out-of-distribution}   & ---                      & ---                     & ---                      & ---                       \\
	\midrule
	\multicolumn{ 5 }{c}{\textbf{concept learning}  (sliced Wasserstein)}                                                                      \\
	\midrule
	\textbf{obs}                   & 0.099 ± 0.006            & 0.091 ± 0.006           & 0.102 ± 0.005            & \textbf{0.086} ± 0.004    \\
	\textbf{scaled}                & 0.062 ± 0.002            & 0.068 ± 0.005           & \textbf{0.061} ± 0.003   & 0.073 ± 0.004             \\
	\textbf{shear}                 & 0.119 ± 0.006            & 0.119 ± 0.007           & 0.127 ± 0.005            & \textbf{0.111} ± 0.006    \\
	\textbf{shift}                 & 0.113 ± 0.005            & 0.111 ± 0.006           & 0.121 ± 0.005            & \textbf{0.108} ± 0.003    \\
	\textbf{swel}                  & 0.126 ± 0.007            & 0.121 ± 0.009           & 0.133 ± 0.005            & \textbf{0.115} ± 0.004    \\
	\textbf{thic}                  & \textbf{0.178} ± 0.007   & 0.186 ± 0.015           & 0.202 ± 0.010            & 0.183 ± 0.005             \\
	\textbf{thin}                  & 0.064 ± 0.004            & 0.072 ± 0.008           & \textbf{0.062} ± 0.003   & 0.074 ± 0.006             \\
	\addlinespace
	\textbf{concept learning mean} & 0.109 ± 0.006            & 0.110 ± 0.008           & 0.115 ± 0.005            & \textbf{0.107} ± 0.005    \\
	\midrule
	\bottomrule
\end{tabular}

	\end{table}
}

\newcommand{\TableQuadAblation}{
	\begin{table}[htbp]
		\centering
		\caption{Evaluation of the context module, ablations, and baselines in terms of reconstruction, concept learning, and composition on \texttt{quad} (independent).}
		\label{tab:quad-ablation}
		\resizebox{\textwidth}{!}{%
			\begin{tabular}{rccccccccc}
\toprule
& \textbf{CM (end-to-end)} & \textbf{CM (fine-tune)} & \textbf{CM (frozen)} & \textbf{base} & \textbf{base pooled} & \textbf{CM pooled} & \textbf{Ada-GVAE} & \textbf{BetaTCVAE} & \textbf{TVAE} \\
\midrule
\multicolumn{ 10 }{c}{\textbf{reconstruction}  (ELBO BPD)} \\
\midrule
\textbf{in-distribution} & 0.519 ± 0.022 & 0.518 ± 0.006 & 0.559 ± 0.011 & \textbf{0.444} ± 0.000 & 0.446 ± 0.002 & 0.840 ± 0.072 & 2.017 ± 0.006 & 1.973 ± 0.001 & 2.048 ± 0.008 \\
\textbf{out-of-distribution} & 0.720 ± 0.069 & 0.675 ± 0.039 & 0.774 ± 0.039 & 3.095 ± 0.295 & \textbf{0.487} ± 0.009 & 0.830 ± 0.042 & 2.092 ± 0.007 & 1.980 ± 0.003 & 2.115 ± 0.019 \\
\midrule
\multicolumn{ 10 }{c}{\textbf{concept learning}  (sliced Wasserstein)} \\
\midrule
\textbf{obs} & 0.056 ± 0.003 & \textbf{0.051} ± 0.004 & 0.078 ± 0.005 & 0.061 ± 0.004 & 0.100 ± 0.007 & 0.110 ± 0.013 & 0.332 ± 0.007 & 0.317 ± 0.009 & 0.349 ± 0.007 \\
\textbf{quad1} & 0.062 ± 0.003 & \textbf{0.056} ± 0.005 & 0.075 ± 0.003 & 0.233 ± 0.004 & 0.201 ± 0.014 & 0.266 ± 0.020 & 0.342 ± 0.009 & 0.335 ± 0.012 & 0.324 ± 0.013 \\
\textbf{quad2} & 0.061 ± 0.005 & \textbf{0.058} ± 0.007 & 0.086 ± 0.009 & 0.234 ± 0.014 & 0.196 ± 0.012 & 0.254 ± 0.017 & 0.346 ± 0.007 & 0.323 ± 0.006 & 0.334 ± 0.010 \\
\textbf{quad3} & \textbf{0.063} ± 0.002 & 0.064 ± 0.005 & 0.085 ± 0.006 & 0.251 ± 0.013 & 0.210 ± 0.008 & 0.244 ± 0.020 & 0.333 ± 0.013 & 0.343 ± 0.010 & 0.334 ± 0.011 \\
\textbf{quad4} & 0.065 ± 0.007 & \textbf{0.057} ± 0.004 & 0.100 ± 0.011 & 0.227 ± 0.010 & 0.227 ± 0.005 & 0.246 ± 0.011 & 0.347 ± 0.016 & 0.334 ± 0.011 & 0.324 ± 0.016 \\
\textbf{size} & 0.075 ± 0.007 & \textbf{0.073} ± 0.006 & 0.101 ± 0.008 & 0.157 ± 0.009 & 0.122 ± 0.001 & 0.177 ± 0.020 & 0.327 ± 0.004 & 0.307 ± 0.010 & 0.336 ± 0.009 \\
\textbf{orientation} & 0.057 ± 0.003 & \textbf{0.053} ± 0.003 & 0.077 ± 0.006 & 0.060 ± 0.003 & 0.104 ± 0.009 & 0.114 ± 0.016 & 0.316 ± 0.011 & 0.349 ± 0.013 & 0.334 ± 0.010 \\
\addlinespace
\textbf{concept learning mean} & 0.063 ± 0.004 & \textbf{0.059} ± 0.005 & 0.086 ± 0.007 & 0.175 ± 0.008 & 0.166 ± 0.008 & 0.202 ± 0.017 & 0.335 ± 0.010 & 0.330 ± 0.010 & 0.334 ± 0.011 \\
\midrule
\multicolumn{ 10 }{c}{\textbf{composition}  (sliced Wasserstein)} \\
\midrule
\textbf{(quad1, quad2)} & \textbf{0.100} ± 0.028 & 0.108 ± 0.016 & 0.137 ± 0.014 & 0.311 ± 0.009 & 0.276 ± 0.014 & 0.343 ± 0.018 & 0.346 ± 0.012 & 0.346 ± 0.004 & 0.334 ± 0.007 \\
\textbf{(quad1, quad3)} & \textbf{0.078} ± 0.016 & 0.107 ± 0.011 & 0.146 ± 0.013 & 0.323 ± 0.011 & 0.273 ± 0.007 & 0.349 ± 0.008 & 0.367 ± 0.016 & 0.372 ± 0.009 & 0.337 ± 0.008 \\
\textbf{(quad1, quad4)} & 0.094 ± 0.032 & \textbf{0.091} ± 0.010 & 0.179 ± 0.014 & 0.346 ± 0.017 & 0.277 ± 0.017 & 0.325 ± 0.023 & 0.377 ± 0.018 & 0.343 ± 0.012 & 0.346 ± 0.010 \\
\textbf{(quad1, size)} & 0.101 ± 0.010 & \textbf{0.087} ± 0.005 & 0.111 ± 0.010 & 0.266 ± 0.012 & 0.226 ± 0.013 & 0.284 ± 0.011 & 0.343 ± 0.005 & 0.331 ± 0.009 & 0.336 ± 0.009 \\
\textbf{(quad1, orientation)} & 0.060 ± 0.003 & \textbf{0.059} ± 0.004 & 0.080 ± 0.002 & 0.236 ± 0.012 & 0.194 ± 0.017 & 0.240 ± 0.004 & 0.367 ± 0.016 & 0.360 ± 0.007 & 0.356 ± 0.007 \\
\textbf{(quad2, quad3)} & \textbf{0.067} ± 0.007 & 0.098 ± 0.017 & 0.148 ± 0.014 & 0.338 ± 0.017 & 0.283 ± 0.014 & 0.325 ± 0.016 & 0.350 ± 0.010 & 0.349 ± 0.012 & 0.350 ± 0.008 \\
\textbf{(quad2, quad4)} & \textbf{0.068} ± 0.004 & 0.078 ± 0.010 & 0.194 ± 0.018 & 0.302 ± 0.013 & 0.291 ± 0.015 & 0.332 ± 0.012 & 0.382 ± 0.007 & 0.342 ± 0.005 & 0.340 ± 0.013 \\
\textbf{(quad2, size)} & 0.099 ± 0.008 & \textbf{0.083} ± 0.008 & 0.102 ± 0.007 & 0.248 ± 0.008 & 0.197 ± 0.004 & 0.281 ± 0.021 & 0.334 ± 0.007 & 0.328 ± 0.015 & 0.336 ± 0.006 \\
\textbf{(quad2, orientation)} & 0.063 ± 0.004 & \textbf{0.058} ± 0.003 & 0.090 ± 0.012 & 0.225 ± 0.011 & 0.214 ± 0.010 & 0.274 ± 0.013 & 0.329 ± 0.011 & 0.348 ± 0.010 & 0.333 ± 0.008 \\
\addlinespace
\textbf{composition mean} & \textbf{0.081} ± 0.013 & 0.086 ± 0.009 & 0.132 ± 0.012 & 0.288 ± 0.012 & 0.248 ± 0.012 & 0.306 ± 0.014 & 0.355 ± 0.011 & 0.346 ± 0.009 & 0.341 ± 0.008 \\
\midrule
\bottomrule
\end{tabular}
		}
	\end{table}
}

\newcommand{\TableQuadCapacityTwo}{
	\begin{table}[htbp]
		\centering
		\caption{Evaluation of CM (end-to-end) in terms of reconstruction, concept learning, and composition on \texttt{quad} (independent) for different capacities, with depth $h_{exp} = 2$ as $(w_{exp}, w_{c})$ varies.}
		\label{tab:quad-capacity-2}
		\resizebox{\textwidth}{!}{%
			\begin{tabular}{rcccccc}
	\toprule
	$(w_{exp)}, w_{c)})$           & \textbf{(16, 16)}      & \textbf{(16, 32)}      & \textbf{(16, 8)}       & \textbf{(32, 16)}      & \textbf{(32, 32)} & \textbf{(32, 8)} \\
	\midrule
	\multicolumn{ 7 }{c}{\textbf{reconstruction}  (ELBO BPD)}                                                                                                                 \\
	\midrule
	\textbf{in-distribution}       & \textbf{0.494} ± 0.014 & 0.497 ± 0.012          & 0.494 ± 0.010          & 0.512 ± 0.023          & 0.557 ± 0.016     & 0.519 ± 0.022    \\
	\textbf{out-of-distribution}   & 0.676 ± 0.024          & 0.665 ± 0.025          & \textbf{0.598} ± 0.025 & 0.626 ± 0.030          & 0.752 ± 0.067     & 0.720 ± 0.069    \\
	\midrule
	\multicolumn{ 7 }{c}{\textbf{concept learning}  (sliced Wasserstein)}                                                                                                     \\
	\midrule
	\textbf{obs}                   & 0.055 ± 0.005          & 0.056 ± 0.002          & 0.056 ± 0.004          & \textbf{0.053} ± 0.003 & 0.063 ± 0.005     & 0.056 ± 0.003    \\
	\textbf{quad1}                 & \textbf{0.057} ± 0.004 & 0.061 ± 0.005          & 0.059 ± 0.004          & 0.058 ± 0.004          & 0.067 ± 0.007     & 0.062 ± 0.003    \\
	\textbf{quad2}                 & 0.061 ± 0.006          & \textbf{0.053} ± 0.003 & 0.060 ± 0.003          & 0.056 ± 0.005          & 0.063 ± 0.005     & 0.061 ± 0.005    \\
	\textbf{quad3}                 & 0.058 ± 0.004          & 0.058 ± 0.005          & 0.065 ± 0.007          & \textbf{0.055} ± 0.003 & 0.066 ± 0.006     & 0.063 ± 0.002    \\
	\textbf{quad4}                 & 0.060 ± 0.007          & 0.063 ± 0.004          & 0.063 ± 0.006          & \textbf{0.055} ± 0.003 & 0.068 ± 0.006     & 0.065 ± 0.007    \\
	\textbf{size}                  & 0.072 ± 0.008          & 0.072 ± 0.007          & 0.076 ± 0.005          & \textbf{0.071} ± 0.005 & 0.082 ± 0.011     & 0.075 ± 0.007    \\
	\textbf{orientation}           & 0.061 ± 0.006          & 0.055 ± 0.002          & 0.058 ± 0.003          & \textbf{0.055} ± 0.003 & 0.062 ± 0.006     & 0.057 ± 0.003    \\
	\addlinespace
	\textbf{concept learning mean} & 0.061 ± 0.006          & 0.060 ± 0.004          & 0.062 ± 0.005          & \textbf{0.057} ± 0.004 & 0.067 ± 0.007     & 0.063 ± 0.004    \\
	\midrule
	\multicolumn{ 7 }{c}{\textbf{composition}  (sliced Wasserstein)}                                                                                                          \\
	\midrule
	\textbf{(quad1, quad2)}        & 0.091 ± 0.010          & 0.086 ± 0.010          & \textbf{0.066} ± 0.006 & 0.076 ± 0.014          & 0.131 ± 0.017     & 0.100 ± 0.028    \\
	\textbf{(quad1, quad3)}        & 0.064 ± 0.008          & 0.083 ± 0.006          & 0.075 ± 0.012          & \textbf{0.059} ± 0.005 & 0.102 ± 0.010     & 0.078 ± 0.016    \\
	\textbf{(quad1, quad4)}        & 0.082 ± 0.019          & 0.069 ± 0.006          & \textbf{0.064} ± 0.007 & 0.076 ± 0.012          & 0.092 ± 0.013     & 0.094 ± 0.032    \\
	\textbf{(quad1, size)}         & \textbf{0.084} ± 0.007 & 0.098 ± 0.009          & 0.094 ± 0.005          & 0.107 ± 0.013          & 0.117 ± 0.011     & 0.101 ± 0.010    \\
	\textbf{(quad1, orientation)}  & 0.059 ± 0.004          & 0.065 ± 0.006          & 0.060 ± 0.004          & \textbf{0.055} ± 0.003 & 0.068 ± 0.011     & 0.060 ± 0.003    \\
	\textbf{(quad2, quad3)}        & 0.083 ± 0.017          & 0.080 ± 0.014          & 0.076 ± 0.015          & \textbf{0.066} ± 0.008 & 0.083 ± 0.012     & 0.067 ± 0.007    \\
	\textbf{(quad2, quad4)}        & 0.079 ± 0.010          & 0.074 ± 0.007          & 0.071 ± 0.010          & \textbf{0.065} ± 0.009 & 0.077 ± 0.008     & 0.068 ± 0.004    \\
	\textbf{(quad2, size)}         & 0.100 ± 0.014          & \textbf{0.085} ± 0.011 & 0.087 ± 0.006          & 0.093 ± 0.006          & 0.116 ± 0.011     & 0.099 ± 0.008    \\
	\textbf{(quad2, orientation)}  & 0.062 ± 0.006          & 0.057 ± 0.003          & 0.060 ± 0.005          & \textbf{0.057} ± 0.003 & 0.067 ± 0.008     & 0.063 ± 0.004    \\
	\addlinespace
	\textbf{composition mean}      & 0.078 ± 0.011          & 0.077 ± 0.008          & \textbf{0.072} ± 0.008 & 0.073 ± 0.008          & 0.095 ± 0.011     & 0.081 ± 0.013    \\
	\midrule
	\bottomrule
\end{tabular}

		}
	\end{table}
}

\newcommand{\TableQuadCapacityFour}{
	\begin{table}[htbp]
		\centering
		\caption{Evaluation of CM (end-to-end) in terms of reconstruction, concept learning, and composition on \texttt{quad} (independent) for different capacities, with depth $h_{exp} = 4$ as $(w_{exp}, w_{c})$ varies.}
		\label{tab:quad-capacity-4}
		\resizebox{\textwidth}{!}{%
			\begin{tabular}{rcccccc}
	\toprule
	$(w_{exp}, w_{c})$             & \textbf{(16, 16)}      & \textbf{(16, 32)}      & \textbf{(16, 8)}       & \textbf{(32, 16)} & \textbf{(32, 32)} & \textbf{(32, 8)}       \\
	\midrule
	\multicolumn{ 7 }{c}{\textbf{reconstruction}  (ELBO BPD)}                                                                                                                  \\
	\midrule
	\textbf{in-distribution}       & \textbf{0.511} ± 0.008 & 0.516 ± 0.015          & 0.573 ± 0.045          & 0.590 ± 0.009     & 0.613 ± 0.020     & 0.580 ± 0.017          \\
	\textbf{out-of-distribution}   & 0.761 ± 0.040          & 0.750 ± 0.033          & 0.674 ± 0.049          & 0.706 ± 0.035     & 0.759 ± 0.081     & \textbf{0.636} ± 0.017 \\
	\midrule
	\multicolumn{ 7 }{c}{\textbf{concept learning}  (sliced Wasserstein)}                                                                                                      \\
	\midrule
	\textbf{obs}                   & 0.059 ± 0.004          & 0.052 ± 0.002          & \textbf{0.052} ± 0.003 & 0.053 ± 0.002     & 0.059 ± 0.006     & 0.065 ± 0.004          \\
	\textbf{quad1}                 & 0.061 ± 0.005          & \textbf{0.054} ± 0.004 & 0.061 ± 0.005          & 0.063 ± 0.002     & 0.081 ± 0.013     & 0.075 ± 0.005          \\
	\textbf{quad2}                 & \textbf{0.057} ± 0.004 & 0.057 ± 0.002          & 0.058 ± 0.003          & 0.064 ± 0.004     & 0.080 ± 0.014     & 0.071 ± 0.005          \\
	\textbf{quad3}                 & 0.059 ± 0.006          & \textbf{0.058} ± 0.003 & 0.064 ± 0.003          & 0.066 ± 0.004     & 0.078 ± 0.014     & 0.073 ± 0.002          \\
	\textbf{quad4}                 & 0.056 ± 0.004          & \textbf{0.053} ± 0.002 & 0.064 ± 0.003          & 0.058 ± 0.004     & 0.075 ± 0.012     & 0.071 ± 0.002          \\
	\textbf{size}                  & \textbf{0.069} ± 0.003 & 0.081 ± 0.005          & 0.077 ± 0.008          & 0.097 ± 0.004     & 0.111 ± 0.016     & 0.087 ± 0.006          \\
	\textbf{orientation}           & 0.059 ± 0.004          & \textbf{0.055} ± 0.003 & 0.061 ± 0.005          & 0.060 ± 0.003     & 0.071 ± 0.012     & 0.071 ± 0.003          \\
	\addlinespace
	\textbf{concept learning mean} & 0.060 ± 0.004          & \textbf{0.059} ± 0.003 & 0.063 ± 0.004          & 0.066 ± 0.003     & 0.079 ± 0.012     & 0.073 ± 0.004          \\
	\midrule
	\multicolumn{ 7 }{c}{\textbf{composition}  (sliced Wasserstein)}                                                                                                           \\
	\midrule
	\textbf{(quad1, quad2)}        & 0.071 ± 0.008          & 0.090 ± 0.010          & \textbf{0.059} ± 0.004 & 0.089 ± 0.009     & 0.091 ± 0.016     & 0.074 ± 0.005          \\
	\textbf{(quad1, quad3)}        & 0.062 ± 0.006          & \textbf{0.062} ± 0.004 & 0.080 ± 0.023          & 0.075 ± 0.005     & 0.086 ± 0.014     & 0.079 ± 0.006          \\
	\textbf{(quad1, quad4)}        & 0.112 ± 0.025          & 0.072 ± 0.010          & \textbf{0.063} ± 0.009 & 0.077 ± 0.004     & 0.086 ± 0.010     & 0.083 ± 0.005          \\
	\textbf{(quad1, size)}         & \textbf{0.083} ± 0.003 & 0.119 ± 0.008          & 0.102 ± 0.007          & 0.106 ± 0.003     & 0.133 ± 0.008     & 0.112 ± 0.005          \\
	\textbf{(quad1, orientation)}  & 0.059 ± 0.003          & \textbf{0.057} ± 0.003 & 0.062 ± 0.006          & 0.064 ± 0.003     & 0.079 ± 0.011     & 0.074 ± 0.004          \\
	\textbf{(quad2, quad3)}        & 0.077 ± 0.011          & 0.072 ± 0.002          & \textbf{0.061} ± 0.004 & 0.083 ± 0.008     & 0.109 ± 0.028     & 0.076 ± 0.006          \\
	\textbf{(quad2, quad4)}        & 0.092 ± 0.016          & 0.083 ± 0.017          & \textbf{0.065} ± 0.004 & 0.081 ± 0.005     & 0.102 ± 0.021     & 0.075 ± 0.007          \\
	\textbf{(quad2, size)}         & \textbf{0.091} ± 0.006 & 0.108 ± 0.009          & 0.101 ± 0.016          & 0.128 ± 0.012     & 0.135 ± 0.021     & 0.108 ± 0.007          \\
	\textbf{(quad2, orientation)}  & \textbf{0.058} ± 0.004 & 0.062 ± 0.003          & 0.064 ± 0.002          & 0.067 ± 0.005     & 0.081 ± 0.015     & 0.073 ± 0.005          \\
	\addlinespace
	\textbf{composition mean}      & 0.078 ± 0.009          & 0.080 ± 0.007          & \textbf{0.073} ± 0.008 & 0.086 ± 0.006     & 0.100 ± 0.016     & 0.084 ± 0.005          \\
	\midrule
	\bottomrule
\end{tabular}

		}
	\end{table}
}

\newcommand{\TableQuadBeta}{
	\begin{table}[htbp]
		\centering
		\caption{Evaluation of CM (end-to-end) in terms of reconstruction, concept learning, and composition on \texttt{quad} (independent) for different latent space regularization strengths, as $\beta$ varies.}
		\label{tab:quad-beta}
		\begin{tabular}{rcccc}
\toprule
& \textbf{0.5} & \textbf{1.0} & \textbf{2.0} & \textbf{4.0} \\
\midrule
\multicolumn{ 5 }{c}{\textbf{reconstruction}  (ELBO BPD)} \\
\midrule
\textbf{in-distribution} & \textbf{0.474} ± 0.003 & 0.519 ± 0.022 & 0.599 ± 0.020 & 0.640 ± 0.029 \\
\textbf{out-of-distribution} & \textbf{0.632} ± 0.028 & 0.720 ± 0.069 & 0.821 ± 0.047 & 0.865 ± 0.061 \\
\midrule
\multicolumn{ 5 }{c}{\textbf{concept learning}  (sliced Wasserstein)} \\
\midrule
\textbf{obs} & 0.050 ± 0.001 & 0.056 ± 0.003 & \textbf{0.048} ± 0.002 & 0.054 ± 0.004 \\
\textbf{quad1} & \textbf{0.051} ± 0.003 & 0.062 ± 0.003 & 0.055 ± 0.002 & 0.064 ± 0.005 \\
\textbf{quad2} & 0.052 ± 0.002 & 0.061 ± 0.005 & \textbf{0.052} ± 0.003 & 0.064 ± 0.011 \\
\textbf{quad3} & 0.059 ± 0.002 & 0.063 ± 0.002 & \textbf{0.047} ± 0.002 & 0.064 ± 0.009 \\
\textbf{quad4} & 0.054 ± 0.002 & 0.065 ± 0.007 & \textbf{0.050} ± 0.002 & 0.058 ± 0.006 \\
\textbf{size} & \textbf{0.061} ± 0.003 & 0.075 ± 0.007 & 0.066 ± 0.005 & 0.082 ± 0.005 \\
\textbf{orientation} & \textbf{0.050} ± 0.002 & 0.057 ± 0.003 & 0.051 ± 0.002 & 0.068 ± 0.008 \\
\addlinespace
\textbf{concept learning mean} & 0.054 ± 0.002 & 0.063 ± 0.004 & \textbf{0.053} ± 0.003 & 0.065 ± 0.007 \\
\midrule
\multicolumn{ 5 }{c}{\textbf{composition}  (sliced Wasserstein)} \\
\midrule
\textbf{(quad1, quad2)} & \textbf{0.059} ± 0.005 & 0.100 ± 0.028 & 0.080 ± 0.005 & 0.082 ± 0.015 \\
\textbf{(quad1, quad3)} & \textbf{0.070} ± 0.008 & 0.078 ± 0.016 & 0.071 ± 0.007 & 0.082 ± 0.014 \\
\textbf{(quad1, quad4)} & 0.076 ± 0.010 & 0.094 ± 0.032 & \textbf{0.073} ± 0.012 & 0.123 ± 0.027 \\
\textbf{(quad1, size)} & \textbf{0.099} ± 0.007 & 0.101 ± 0.010 & 0.118 ± 0.008 & 0.115 ± 0.010 \\
\textbf{(quad1, orientation)} & \textbf{0.052} ± 0.002 & 0.060 ± 0.003 & 0.057 ± 0.003 & 0.073 ± 0.005 \\
\textbf{(quad2, quad3)} & \textbf{0.056} ± 0.000 & 0.067 ± 0.007 & 0.074 ± 0.008 & 0.094 ± 0.021 \\
\textbf{(quad2, quad4)} & 0.087 ± 0.018 & 0.068 ± 0.004 & 0.101 ± 0.027 & \textbf{0.067} ± 0.009 \\
\textbf{(quad2, size)} & \textbf{0.086} ± 0.015 & 0.099 ± 0.008 & 0.121 ± 0.018 & 0.131 ± 0.011 \\
\textbf{(quad2, orientation)} & \textbf{0.055} ± 0.003 & 0.063 ± 0.004 & 0.056 ± 0.006 & 0.068 ± 0.009 \\
\addlinespace
\textbf{composition mean} & \textbf{0.071} ± 0.008 & 0.081 ± 0.013 & 0.083 ± 0.010 & 0.093 ± 0.013 \\
\midrule
\bottomrule
\end{tabular}
	\end{table}
}

\newcommand{\TableQuadGroupNorm}{
	\begin{table}[htbp]
		\centering
		\caption{Evaluation of CM (end-to-end) in terms of reconstruction, concept learning, and composition on \texttt{quad} (independent) for different group norm regularization strengths.}
		\label{tab:quad-groupnorm}
		\begin{tabular}{rcccc}
	\toprule
	                               & \(\lambda =\) \textbf{0} & \(\lambda =\)\textbf{1} & \(\lambda =\)\textbf{10} & \(\lambda =\)\textbf{100} \\
	\midrule
	\multicolumn{ 5 }{c}{\textbf{reconstruction}  (ELBO BPD)}                                                                                  \\
	\midrule
	\textbf{in-distribution}       & 0.519 ± 0.022            & 0.511 ± 0.023           & 0.528 ± 0.030            & \textbf{0.492} ± 0.009    \\
	\textbf{out-of-distribution}   & 0.720 ± 0.069            & \textbf{0.703} ± 0.052  & 0.868 ± 0.147            & 0.993 ± 0.181             \\
	\midrule
	\multicolumn{ 5 }{c}{\textbf{concept learning}  (sliced Wasserstein)}                                                                      \\
	\midrule
	\textbf{obs}                   & 0.056 ± 0.003            & \textbf{0.049} ± 0.001  & 0.054 ± 0.004            & 0.055 ± 0.002             \\
	\textbf{quad1}                 & 0.062 ± 0.003            & 0.058 ± 0.004           & \textbf{0.056} ± 0.005   & 0.060 ± 0.001             \\
	\textbf{quad2}                 & 0.061 ± 0.005            & 0.057 ± 0.006           & 0.057 ± 0.003            & \textbf{0.055} ± 0.004    \\
	\textbf{quad3}                 & 0.063 ± 0.002            & 0.058 ± 0.004           & \textbf{0.058} ± 0.003   & 0.061 ± 0.006             \\
	\textbf{quad4}                 & 0.065 ± 0.007            & 0.054 ± 0.003           & \textbf{0.052} ± 0.003   & 0.056 ± 0.003             \\
	\textbf{size}                  & 0.075 ± 0.007            & 0.076 ± 0.007           & 0.073 ± 0.007            & \textbf{0.069} ± 0.005    \\
	\textbf{orientation}           & 0.057 ± 0.003            & 0.057 ± 0.003           & \textbf{0.055} ± 0.006   & 0.065 ± 0.006             \\
	\addlinespace
	\textbf{concept learning mean} & 0.063 ± 0.004            & 0.058 ± 0.004           & \textbf{0.058} ± 0.004   & 0.060 ± 0.004             \\
	\midrule
	\multicolumn{ 5 }{c}{\textbf{composition}  (sliced Wasserstein)}                                                                           \\
	\midrule
	\textbf{(quad1, quad2)}        & 0.100 ± 0.028            & \textbf{0.061} ± 0.008  & 0.092 ± 0.023            & 0.101 ± 0.038             \\
	\textbf{(quad1, quad3)}        & \textbf{0.078} ± 0.016   & 0.080 ± 0.019           & 0.101 ± 0.017            & 0.134 ± 0.019             \\
	\textbf{(quad1, quad4)}        & 0.094 ± 0.032            & \textbf{0.081} ± 0.017  & 0.127 ± 0.040            & 0.132 ± 0.043             \\
	\textbf{(quad1, size)}         & 0.101 ± 0.010            & \textbf{0.098} ± 0.006  & 0.115 ± 0.011            & 0.127 ± 0.025             \\
	\textbf{(quad1, orientation)}  & 0.060 ± 0.003            & \textbf{0.055} ± 0.004  & 0.057 ± 0.004            & 0.063 ± 0.005             \\
	\textbf{(quad2, quad3)}        & \textbf{0.067} ± 0.007   & 0.072 ± 0.008           & 0.075 ± 0.008            & 0.123 ± 0.042             \\
	\textbf{(quad2, quad4)}        & \textbf{0.068} ± 0.004   & 0.087 ± 0.029           & 0.075 ± 0.006            & 0.145 ± 0.024             \\
	\textbf{(quad2, size)}         & 0.099 ± 0.008            & \textbf{0.097} ± 0.004  & 0.111 ± 0.019            & 0.108 ± 0.022             \\
	\textbf{(quad2, orientation)}  & 0.063 ± 0.004            & 0.059 ± 0.004           & \textbf{0.058} ± 0.004   & 0.065 ± 0.008             \\
	\addlinespace
	\textbf{composition mean}      & 0.081 ± 0.013            & \textbf{0.077} ± 0.011  & 0.090 ± 0.015            & 0.111 ± 0.025             \\
	\midrule
	\bottomrule
\end{tabular}

	\end{table}
}

\newcommand{\TableQuadEllTwoNorm}{
	\begin{table}[htbp]
		\centering
		\caption{Evaluation of CM (end-to-end) in terms of reconstruction, concept learning, and composition on \texttt{quad} (independent) for different \(L_2\)-norm regularization strengths.}
		\label{tab:quad-l2norm}
		\begin{tabular}{rcccc}
	\toprule
	                               & \(\lambda =\) \textbf{0} & \(\lambda =\)\textbf{1} & \(\lambda =\)\textbf{10} & \(\lambda =\)\textbf{100} \\
	\midrule
	\multicolumn{ 5 }{c}{\textbf{reconstruction}  (ELBO BPD)}                                                                                  \\
	\midrule
	\textbf{in-distribution}       & 0.519 ± 0.022            & 0.529 ± 0.029           & 0.550 ± 0.029            & \textbf{0.507} ± 0.024    \\
	\textbf{out-of-distribution}   & 0.720 ± 0.069            & \textbf{0.662} ± 0.058  & 0.676 ± 0.023            & 0.685 ± 0.040             \\
	\midrule
	\multicolumn{ 5 }{c}{\textbf{concept learning}  (sliced Wasserstein)}                                                                      \\
	\midrule
	\textbf{obs}                   & 0.056 ± 0.003            & \textbf{0.054} ± 0.001  & 0.055 ± 0.002            & 0.054 ± 0.001             \\
	\textbf{quad1}                 & 0.062 ± 0.003            & 0.058 ± 0.003           & 0.061 ± 0.003            & \textbf{0.053} ± 0.004    \\
	\textbf{quad2}                 & 0.061 ± 0.005            & \textbf{0.054} ± 0.003  & 0.062 ± 0.002            & 0.056 ± 0.002             \\
	\textbf{quad3}                 & 0.063 ± 0.002            & 0.057 ± 0.002           & 0.059 ± 0.005            & \textbf{0.055} ± 0.005    \\
	\textbf{quad4}                 & 0.065 ± 0.007            & 0.053 ± 0.003           & 0.058 ± 0.002            & \textbf{0.052} ± 0.004    \\
	\textbf{size}                  & 0.075 ± 0.007            & 0.073 ± 0.009           & 0.076 ± 0.007            & \textbf{0.070} ± 0.004    \\
	\textbf{orientation}           & 0.057 ± 0.003            & 0.058 ± 0.004           & 0.063 ± 0.004            & \textbf{0.056} ± 0.005    \\
	\addlinespace
	\textbf{concept learning mean} & 0.063 ± 0.004            & 0.058 ± 0.003           & 0.062 ± 0.003            & \textbf{0.057} ± 0.004    \\
	\midrule
	\multicolumn{ 5 }{c}{\textbf{composition}  (sliced Wasserstein)}                                                                           \\
	\midrule
	\textbf{(quad1, quad2)}        & 0.100 ± 0.028            & 0.070 ± 0.007           & 0.069 ± 0.003            & \textbf{0.058} ± 0.003    \\
	\textbf{(quad1, quad3)}        & 0.078 ± 0.016            & 0.073 ± 0.014           & 0.096 ± 0.021            & \textbf{0.063} ± 0.005    \\
	\textbf{(quad1, quad4)}        & 0.094 ± 0.032            & 0.101 ± 0.029           & 0.081 ± 0.012            & \textbf{0.079} ± 0.019    \\
	\textbf{(quad1, size)}         & 0.101 ± 0.010            & 0.095 ± 0.011           & 0.098 ± 0.004            & \textbf{0.093} ± 0.007    \\
	\textbf{(quad1, orientation)}  & 0.060 ± 0.003            & \textbf{0.056} ± 0.004  & 0.061 ± 0.002            & 0.058 ± 0.003             \\
	\textbf{(quad2, quad3)}        & \textbf{0.067} ± 0.007   & 0.071 ± 0.009           & 0.080 ± 0.006            & 0.071 ± 0.009             \\
	\textbf{(quad2, quad4)}        & 0.068 ± 0.004            & 0.072 ± 0.009           & 0.067 ± 0.003            & \textbf{0.063} ± 0.006    \\
	\textbf{(quad2, size)}         & 0.099 ± 0.008            & \textbf{0.084} ± 0.006  & 0.097 ± 0.008            & 0.089 ± 0.005             \\
	\textbf{(quad2, orientation)}  & 0.063 ± 0.004            & \textbf{0.058} ± 0.003  & 0.061 ± 0.003            & 0.061 ± 0.003             \\
	\addlinespace
	\textbf{composition mean}      & 0.081 ± 0.013            & 0.076 ± 0.010           & 0.079 ± 0.007            & \textbf{0.071} ± 0.007    \\
	\midrule
	\bottomrule
\end{tabular}

	\end{table}
}

\newcommand{\TableQuadDependentAblation}{
	\begin{table}[htbp]
		\centering
		\caption{Evaluation of the context module, ablations, and baselines in terms of reconstruction, concept learning, and composition on \texttt{quad} (dependent).}
		\label{tab:quad-dep-ablation}
		\resizebox{\textwidth}{!}{%
			\begin{tabular}{rccccccccc}
\toprule
& \textbf{CM (end-to-end)} & \textbf{CM (fine-tune)} & \textbf{CM (frozen)} & \textbf{base} & \textbf{base pooled} & \textbf{CM pooled} & \textbf{Ada-GVAE} & \textbf{BetaTCVAE} & \textbf{TVAE} \\
\midrule
\multicolumn{ 10 }{c}{\textbf{reconstruction}  (ELBO BPD)} \\
\midrule
\textbf{in-distribution} & 0.828 ± 0.004 & 0.840 ± 0.005 & 0.889 ± 0.017 & 0.767 ± 0.001 & \textbf{0.746} ± 0.000 & 0.752 ± 0.000 & 2.443 ± 0.002 & 2.408 ± 0.000 & 2.455 ± 0.002 \\
\textbf{out-of-distribution} & 0.863 ± 0.006 & 0.908 ± 0.005 & 0.959 ± 0.011 & 0.815 ± 0.005 & \textbf{0.746} ± 0.001 & 0.753 ± 0.001 & 2.450 ± 0.002 & 2.406 ± 0.000 & 2.463 ± 0.002 \\
\midrule
\multicolumn{ 10 }{c}{\textbf{concept learning}  (sliced Wasserstein)} \\
\midrule
\textbf{obs} & 0.107 ± 0.005 & \textbf{0.103} ± 0.008 & 0.116 ± 0.003 & 0.110 ± 0.005 & 0.104 ± 0.006 & 0.113 ± 0.004 & 0.365 ± 0.007 & 0.375 ± 0.008 & 0.357 ± 0.005 \\
\textbf{quad1} & \textbf{0.100} ± 0.004 & 0.102 ± 0.004 & 0.107 ± 0.008 & 0.158 ± 0.003 & 0.163 ± 0.004 & 0.137 ± 0.007 & 0.373 ± 0.004 & 0.352 ± 0.004 & 0.374 ± 0.008 \\
\textbf{quad2} & 0.103 ± 0.004 & \textbf{0.100} ± 0.006 & 0.110 ± 0.009 & 0.179 ± 0.007 & 0.165 ± 0.008 & 0.173 ± 0.009 & 0.373 ± 0.004 & 0.359 ± 0.003 & 0.366 ± 0.005 \\
\textbf{quad3} & \textbf{0.127} ± 0.017 & 0.127 ± 0.008 & 0.139 ± 0.008 & 0.200 ± 0.010 & 0.214 ± 0.009 & 0.213 ± 0.015 & 0.379 ± 0.005 & 0.374 ± 0.006 & 0.372 ± 0.009 \\
\textbf{quad4} & 0.145 ± 0.010 & 0.140 ± 0.007 & \textbf{0.131} ± 0.011 & 0.305 ± 0.012 & 0.277 ± 0.010 & 0.278 ± 0.015 & 0.401 ± 0.015 & 0.386 ± 0.008 & 0.388 ± 0.013 \\
\textbf{size} & 0.109 ± 0.006 & \textbf{0.105} ± 0.006 & 0.117 ± 0.004 & 0.116 ± 0.004 & 0.121 ± 0.007 & 0.119 ± 0.004 & 0.361 ± 0.006 & 0.362 ± 0.005 & 0.367 ± 0.005 \\
\textbf{orientation} & 0.123 ± 0.003 & \textbf{0.109} ± 0.007 & 0.124 ± 0.004 & 0.120 ± 0.009 & 0.123 ± 0.008 & 0.130 ± 0.007 & 0.369 ± 0.006 & 0.353 ± 0.002 & 0.365 ± 0.007 \\
\addlinespace
\textbf{concept learning mean} & 0.116 ± 0.007 & \textbf{0.112} ± 0.007 & 0.121 ± 0.007 & 0.170 ± 0.007 & 0.167 ± 0.007 & 0.166 ± 0.009 & 0.375 ± 0.007 & 0.366 ± 0.005 & 0.370 ± 0.007 \\
\midrule
\multicolumn{ 10 }{c}{\textbf{composition}  (sliced Wasserstein)} \\
\midrule
\textbf{(quad1, quad2)} & 0.111 ± 0.008 & \textbf{0.097} ± 0.007 & 0.123 ± 0.011 & 0.229 ± 0.003 & 0.214 ± 0.005 & 0.196 ± 0.013 & 0.366 ± 0.006 & 0.381 ± 0.010 & 0.367 ± 0.007 \\
\textbf{(quad1, quad3)} & 0.134 ± 0.008 & \textbf{0.117} ± 0.003 & 0.140 ± 0.008 & 0.227 ± 0.009 & 0.228 ± 0.008 & 0.198 ± 0.011 & 0.390 ± 0.009 & 0.367 ± 0.011 & 0.374 ± 0.004 \\
\textbf{(quad1, quad4)} & \textbf{0.133} ± 0.007 & 0.163 ± 0.010 & 0.163 ± 0.008 & 0.284 ± 0.015 & 0.292 ± 0.006 & 0.276 ± 0.013 & 0.403 ± 0.012 & 0.376 ± 0.013 & 0.361 ± 0.006 \\
\textbf{(quad1, size)} & 0.108 ± 0.003 & \textbf{0.106} ± 0.004 & 0.123 ± 0.007 & 0.172 ± 0.009 & 0.170 ± 0.009 & 0.157 ± 0.011 & 0.360 ± 0.008 & 0.357 ± 0.008 & 0.367 ± 0.003 \\
\textbf{(quad1, orientation)} & 0.110 ± 0.007 & \textbf{0.108} ± 0.007 & 0.125 ± 0.006 & 0.167 ± 0.006 & 0.157 ± 0.007 & 0.143 ± 0.007 & 0.365 ± 0.010 & 0.371 ± 0.007 & 0.369 ± 0.004 \\
\textbf{(quad2, quad3)} & 0.158 ± 0.007 & \textbf{0.150} ± 0.013 & 0.164 ± 0.005 & 0.222 ± 0.008 & 0.200 ± 0.011 & 0.195 ± 0.010 & 0.367 ± 0.007 & 0.378 ± 0.006 & 0.379 ± 0.008 \\
\textbf{(quad2, quad4)} & 0.130 ± 0.011 & 0.122 ± 0.009 & \textbf{0.119} ± 0.010 & 0.299 ± 0.008 & 0.282 ± 0.025 & 0.279 ± 0.008 & 0.403 ± 0.010 & 0.404 ± 0.013 & 0.382 ± 0.014 \\
\textbf{(quad2, size)} & 0.126 ± 0.009 & \textbf{0.109} ± 0.006 & 0.120 ± 0.005 & 0.191 ± 0.011 & 0.166 ± 0.010 & 0.181 ± 0.010 & 0.368 ± 0.004 & 0.365 ± 0.005 & 0.363 ± 0.004 \\
\textbf{(quad2, orientation)} & 0.135 ± 0.004 & \textbf{0.120} ± 0.010 & 0.133 ± 0.006 & 0.208 ± 0.011 & 0.195 ± 0.020 & 0.198 ± 0.010 & 0.368 ± 0.005 & 0.366 ± 0.005 & 0.351 ± 0.006 \\
\addlinespace
\textbf{composition mean} & 0.127 ± 0.007 & \textbf{0.121} ± 0.008 & 0.134 ± 0.007 & 0.222 ± 0.009 & 0.212 ± 0.011 & 0.203 ± 0.010 & 0.377 ± 0.008 & 0.374 ± 0.009 & 0.368 ± 0.006 \\
\midrule
\bottomrule
\end{tabular}
		}
	\end{table}
}

\newcommand{\TableThreeDIdentDisent}{
	\begin{table}[htbp]
		\centering
		\caption{Evaluation of the baselines in terms of reconstruction, concept learning, and composition on 3DIdent.}
		\label{tab:3didnt-disent}
		\begin{tabular}{rccc}
\toprule
& \textbf{Ada-GVAE} & \textbf{BetaTCVAE} & \textbf{TVAE} \\
\midrule
\multicolumn{ 4 }{c}{\textbf{reconstruction}  (ELBO BPD)} \\
\midrule
\textbf{in-distribution} & 3.267 ± 0.001 & \textbf{3.261} ± 0.000 & 3.282 ± 0.005 \\
\textbf{out-of-distribution} & 3.274 ± 0.001 & \textbf{3.264} ± 0.001 & 3.291 ± 0.006 \\
\midrule
\multicolumn{ 4 }{c}{\textbf{concept learning}  (sliced Wasserstein)} \\
\midrule
\textbf{obs} & 0.267 ± 0.011 & 0.257 ± 0.012 & \textbf{0.245} ± 0.003 \\
\textbf{bg} & \textbf{0.232} ± 0.012 & 0.254 ± 0.009 & 0.262 ± 0.006 \\
\textbf{obj} & \textbf{0.243} ± 0.007 & 0.263 ± 0.010 & 0.259 ± 0.007 \\
\textbf{sl} & 0.253 ± 0.013 & 0.254 ± 0.007 & \textbf{0.250} ± 0.009 \\
\addlinespace
\textbf{concept learning mean} & \textbf{0.249} ± 0.011 & 0.257 ± 0.010 & 0.254 ± 0.006 \\
\midrule
\multicolumn{ 4 }{c}{\textbf{composition}  (sliced Wasserstein)} \\
\midrule
\textbf{(bg, obj)} & \textbf{0.252} ± 0.014 & 0.261 ± 0.011 & 0.256 ± 0.010 \\
\textbf{(bg, sl)} & \textbf{0.227} ± 0.009 & 0.255 ± 0.012 & 0.241 ± 0.013 \\
\textbf{(obj, sl)} & 0.274 ± 0.010 & \textbf{0.247} ± 0.005 & 0.269 ± 0.009 \\
\addlinespace
\textbf{composition mean} & \textbf{0.251} ± 0.011 & 0.254 ± 0.009 & 0.256 ± 0.011 \\
\midrule
\bottomrule
\end{tabular}
	\end{table}
}

\graphicspath{{figures/}}

\title{Intervening to Learn and Compose\\ Causally Disentangled Representations}

\clearauthor{%
 \Name{Alex Markham}  \Email{\href{mailto:alex.markham@causal.dev}{alex.markham@causal.dev}}\\
 \addr University of Copenhagen
 \AND
 \Name{Isaac Hirsch}\textsuperscript{\textdagger}
 \AND
 \Name{Jeri A. Chang}\textsuperscript{\textdagger}
 \AND
 \Name{Liam Solus}\\
 \addr KTH Royal Institute of Technology
 \AND
 \Name{Bryon Aragam}\textsuperscript{\textdagger}\\
 \addr \textsuperscript{\textdagger}University of Chicago
}

\begin{document}
\maketitle

\begin{abstract}
	In designing generative models, it is commonly believed that in order to learn useful latent structure, we face a fundamental tension between expressivity and structure.
	In this paper we challenge this view by proposing a new approach to training arbitrarily expressive generative models that simultaneously learn causally disentangled concepts.
	This is accomplished by adding a simple \emph{context module} to an arbitrarily complex black-box model, which learns to process concept information by implicitly inverting linear representations from the model's encoder.
	Inspired by the notion of intervention in a causal model, our module selectively modifies its architecture during training, allowing it to learn a compact joint model over different contexts.
	We show how adding this module leads to causally disentangled representations that can be composed for out-of-distribution generation on both real and simulated data.
	The resulting models can be trained end-to-end or fine-tuned from pre-trained models.
	To further validate our proposed approach, we prove a new identifiability result that extends existing work on identifying structured representations.
\end{abstract}


\section{Introduction}

Generative models have transformed information processing and demonstrated remarkable capacities for creativity in a variety of tasks ranging from vision to language to audio.
The success of these models has been largely driven by modular, differentiable architectures based on deep neural networks that learn useful representations for downstream tasks.
Recent years have seen increased interest in understanding and exploring the representations produced by these models through evolving lines of work on structured representation learning, identifiability and interpretability, disentanglement, and causal generative models.
This work is motivated in part by the desire to produce performant generative models that also capture meaningful, semantic latent spaces that enable out-of-distribution (OOD) generation.

A key driver behind this work is the tradeoff between flexibility and structure, or expressivity and interpretability: Conventional wisdom suggests that to learn structured, interpretable representations, model capacity must be constrained, sacrificing flexibility and expressivity.
This intuition is supported by a growing body of work on nonlinear ICA \citep{hyvarinen1999nonlinear}, disentanglement \citep{bengio2013deep}, and causal representation learning \citep{scholkopf2021toward}.
On the practical side, methods that have been developed to learn structured latent spaces tend to be bespoke to specific data types and models, and typically impose limitations on the flexibility and expressivity of the underlying models.
Moreover, the most successful methods for learning structured representations typically impose fixed, known structure \emph{a priori}, as opposed to learning this structure from data.

\begin{figure*}[t]
	\centering
	\includegraphics[width=.6\textwidth]{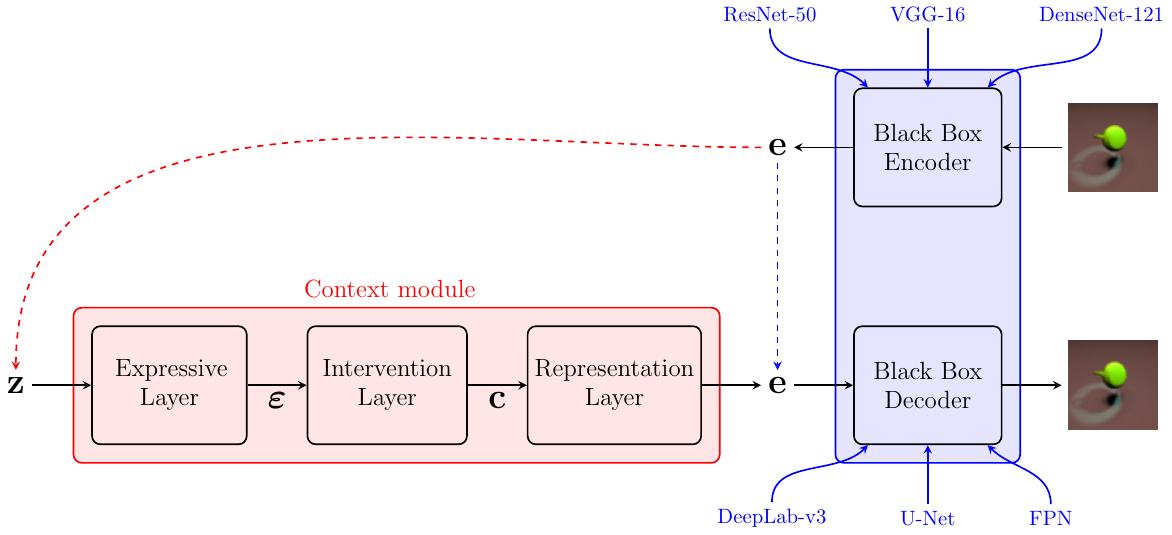}\hfil
	\includegraphics[width=.2\textwidth,trim=6 6 6 6,clip]{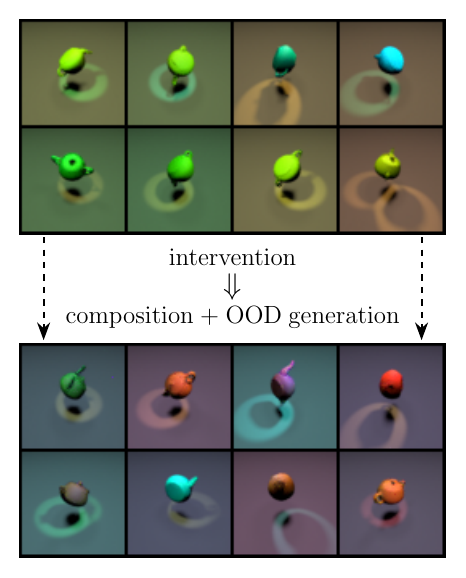}
	\caption{Overview (see~\eqref{eq:decoder:layers} for notation map).
		Given a black-box encoder-decoder architecture (blue), we propose to prepend a \emph{context module} (red) to the decoder.
		(left) Instead of passing the output of the encoder directly to the decoder (blue box + blue arrow), the embeddings $\be$ are passed through the context module, consisting of three distinct layers.
		The output of this module is then passed into the decoder.
		(right) The model learns to compose different concepts OOD, e.g., object and background colour, to values that never appear together in the training data.}
	\label{fig:main}
\end{figure*}

At the same time, there is also a growing body of work that suggests generative models already learn surprisingly structured latent spaces
\citep[e.g.][]{mikolov2013linguistic,szegedy2013intriguing,radford2015unsupervised}.
This in turn suggests that existing models are already ``close'' to capturing useful structure, and perhaps only small modifications are needed.
Since we already have performant models that achieve state-of-the-art results in generation and prediction, do we need to re-invent the wheel to achieve these goals?
Our hypothesis is that we should be able to leverage the expressiveness of these models to build new models that learn latent structure from scratch.
We emphasize that our goal is not to explain or interpolate the latent space of a pre-trained model, but rather to leverage known architectures to train a \emph{new} model, either end-to-end or by fine-tuning an existing model.

In this paper, we adopt this perspective: We start with an arbitrarily complex black-box model, upon which we make no assumptions, then augment this model to learn causally disentangled representations.
The idea is that the black-box model is already known to perform well on downstream tasks (e.g., generation, prediction), and so is flexible enough to capture complex patterns in data.
We introduce a modular, end-to-end differentiable architecture for learning causally disentangled representations \citep{bengio2013deep,scholkopf2021toward} that augments the original model with a simple module that retains its existing capabilities while enabling concept identification and intervention as well as composing together multiple concept interventions for OOD generation (Figure~\ref{fig:main}).
The resulting latent structure is learned entirely from data, with no fixed structure imposed \emph{a priori}.
The motivation is to provide a framework for taking existing performant architectures and to train a new model that performs just as well, but with the added benefit of learning structured representations without imposing specific prior knowledge or structure.
Since the model, unlike previous approaches, does not impose rigid latent structure, we theoretically validate the approach with a novel identifiability result for structured representations recovered by the model.
For a detailed discussion of related work, see Appendix~\ref{sec:related}.

\paragraph{Contributions}
More specifically, we make the following contributions:
\begin{enumerate}[itemsep=-4pt]
	\item We introduce a \emph{context module} that is attached to a decoder to learn concept representations by splitting each concept into a tensor slice, where each slice represents an interventional context in a reduced form structural equation model (SEM).
	      This module can be attached to any decoder and trained end-to-end or fine-tuned (Figure~\ref{fig:main}).
	\item We evaluate this module through quantitative and qualitative experiments on OOD generation, which is distinct from and significantly more challenging than OOD reconstruction as in prior work \citep[e.g.][]{xu2022compositional,mo2024compositional,montero2022lost,schott2022visual}.
	      We perform experiments on OOD generation on several real datasets as well as carefully controlled simulations.
	\item We illustrate the adaptivity of our module to different architectures \citep[including NVAE,][]{vahdat2020nvae} and against multiple ablations that control model capacity, data contexts, training protocols, and hyperparameters, ensuring that any differences in performance are directly attributable to the context module.
	\item We fine-tune existing, pre-trained models after attaching the context module, giving substantial computational savings compared to end-to-end training while improving the learned representations and lending OOD generation abilities to pre-trained models.
\end{enumerate}
As a matter of independent interest, we prove an identifiability result under concept interventions (Theorem~\ref{theorem:ident}), and discuss how the proposed architecture can be interpreted as an approximation to this model.
We also introduce \texttt{quad}, a simple simulated visual environment for evaluating causal disentanglement, compositional abilities, and OOD generation.
Due to space limitations, full technical proofs, along with detailed literature review and experimental details, can be found in the appendix.

\section{Preliminaries}
\label{sec:prelim}

Let $\bx$ denote the observations, e.g., pixels in an image, and $\bz$ denote hidden variables, e.g., latent variables that are to be inferred from the pixels.
A typical generative model consists of a decoder $p_{\theta}(\bx\given\bz)$ and an encoder $q_{\varphi}(\bz\given \bx)$.
After specifying a prior $p_{\theta}(\bz)$, this defines a likelihood by
\begin{align}
	\label{eq:lvm}
	p_{\theta}(\bx)
	= \int p_{\theta}(\bx\given\bz)p_{\theta}(\bz)\dif{\bz}.
\end{align}
Both the decoder and encoder are specified by deep neural networks that are trained end-to-end via standard techniques \citep{kingma2013auto,rezende2014stochastic,rezende2015variational}.
In this work, we focus on variational autoencoders (VAEs) since they are often use to represent structured and semantic latent spaces, which is our focus.
Unlike traditional structured VAEs, we do not impose a specific structure \emph{a priori}.

\paragraph{Causal disentanglement}
\label{sec:cd}
Our goal is to learn distinct latent factors that can be composed together.
This is closely related to causal disentanglement, which we recall informally here:
\begin{definition}[\citealp{bengio2013deep,thomas_disentangling_2017}]\label{defn:disentangled}
	A \emph{causally disentangled} latent space consists of separately controllable (intervenable) latent factors.
\end{definition}
\noindent
In particular, we want to manipulate and intervene on latent factors in a causally meaningful way; crucially this allows us to compose distinct concepts together without needing to learn the full causal graph in the latent space.
This notion of disentanglement is also different from the axis-alignment notion \citep{kim2018disentangling,chen2018isolating}.
The ``causal'' aspect comes from interpreting ``separately controllable'' to mean ``able to intervene on independent causal mechanisms'', as suggested by \citet{locatello2019challenging}; see also \citet{scholkopf2021toward}.
See Appendix~\ref{app:ident-prop-model} for formal details.
Rather than attempt to reconstruct the full causal DAG model (as in \citealp{yang2021causalvae,seigal2022linear,buchholz2023learning}),
we aim for the more practical goal of learning different interventional contexts through distributional invariances;
thus our approach strikes a balance between easy-to-learn unstructured latent spaces and hard-to-learn but desirable fully structured causal DAG latent spaces.

\paragraph{Measuring causal disentanglement}
\label{sec:ood}

A standard approach to quantifying causal disentanglement is to evaluate the recovery of the latent causal graph; since our explicit goal is to avoid learning this graph, we propose a different metric that is more suitable to downstream applications.
To measure causal disentanglement, we use OOD generation on novel interventions as a metric.
We train models so that they never see examples of certain latent factors composed together (equivalently, multiple simultaneous interventions on distinct factors), so these examples are genuinely OOD.
By withholding these examples during training and building a model that is capable of intervening on multiple factors, we can compare the quality of OOD samples generated from our model to the held-out OOD samples.
In our experiments, we use the sliced Wasserstein (SW) metric~\eqref{eq:sw}; see Appendix~\ref{app:evaluation-metrics} for details.

\paragraph{OOD reconstruction vs.~OOD generation}
Previous work has evaluated the OOD capabilities of generative models by using OOD reconstruction as a metric \citep[e.g.][]{xu2022compositional,mo2024compositional,montero2022lost,schott2022visual}.
There is a crucial difference between OOD reconstruction and OOD generation:
By OOD generation, we mean the ability to conditionally generate and combine novel interventions on existing concepts.
For example, suppose during training the model only sees small, red objects along with big, blue objects.
At test time, we wish to generate OOD samples of big, red objects or small, blue objects that were never seen during training.

The crucial difference is that while we can \emph{always} attempt to reconstruct \emph{any} sample (i.e., OOD or not) and evaluate its reconstruction error, it is not always possible to have the model \emph{generate} a new OOD sample.
Unless the model learns specific latent factors corresponding to concepts such as size or colour, we cannot control (i.e., intervene) the size or colour of random samples from the model.

Thus, OOD generation not only evaluates the ability of a model to compose learned concepts in new ways, but also the ability of a model to identify and capture underlying concepts of interest.
Accordingly, we argue that OOD generation is more appropriate to evaluate causal disentanglement since it captures \emph{both} the model's ability to learn concepts as distinct latent factors \emph{as well as} compose them together in novel ways.

\paragraph{Set-up and approach}
Our set-up is the following: We have a black-box encoder and decoder, on which we make no assumptions other than the encoder outputs a latent code corresponding to $\bx$.
In the notation of~\eqref{eq:lvm}, this corresponds to $\bz$, however, to distinguish the black-box embeddings from our model, we will denote the black-box embeddings hereafter by $\be$.
Our plan is to work solely with the embeddings $\be$ and learn how to extract linear concept representations from $\be$ such that distinct concepts can be intervened upon and composed together to create novel, OOD samples.

Our approach is motivated by the following incongruence: On the one hand, it is well-established that the latent spaces of generative models are typically entangled and semantically misaligned, fail to generalize OOD, and suffer from posterior collapse (which has been tied to latent variable nonidentifiability, \citealp{wang2021posterior}).
On the other hand, they still learn structured latent spaces that can be interpolated, are ``nearly'' identifiable \citep{willetts2021don,reizinger2022embrace}, and represent abstract concepts linearly \citep{mikolov2013linguistic,szegedy2013intriguing,radford2015unsupervised}.
So, to learn latent structure, we simply need to extract the right concept representations from the already performant embeddings of existing architectures and learn how to compose them together.

\section{Architecture}

We start with a black-box encoder-decoder architecture along with $\dcon$ concepts of interest, denoted by $\bc=(\bc_{1},\ldots,\bc_{\dcon})$.
Our objectives are three-fold:
\begin{enumerate}[itemsep=-3pt]
	\item To extract concepts from black-box embeddings $\be$ as linear projections $C\be$;
	\item To compose concepts together in a single, transparent model that captures how different concepts are related;
	\item To accomplish both of these objectives without restricting the black-box encoder or decoder architectures, and without sacrificing sampling quality.
\end{enumerate}
A key intuition is that composition can be interpreted as a type of intervention in the latent space.
This is sensible since interventions in a causal model are a type of nontrivial distribution shift.
Thus, the problem takes on a causal flavour which we leverage to build our architecture.
The difficulty with this from the causal modeling perspective is that encoding structural assignments and/or causal mechanisms directly into a feed-forward neural network is tricky, because edges between nodes within the same layer are not allowed in a feed-forward network but are required for the usual DAG representation of a causal model.
To bypass this, we use the \emph{reduced form} of a causal model which has a clear representation as a bipartite directed graph, with arrows only from exogenous to endogenous variables, making it conducive to being embedded into a neural network.
The tradeoff is that this does not learn a causal DAG, however it still enables the model to perform interventions directly in the latent space and to leverage invariances between interventional contexts.

\subsection{Overview}

Before outlining the architectural details, we provide a high-level overview of the main idea.
A traditional decoder transforms the embeddings $\be$ into the observed variables $\bx$, and the encoder operates in reverse by encoding $\bx$ into $\be$.
Thus,
\begin{align*}
	\bx\overset{\textup{encoder}}{\xrightarrow{\hspace{2em}}}\be\overset{\textup{decoder}}{\xrightarrow{\hspace{2em}}}\widehat{\bx}.
\end{align*}
Based on an extensive body of empirical work that shows concepts are linearly represented~\citep[][see Appendix~\ref{sec:related} for more discussion]{mikolov2013linguistic,szegedy2013intriguing,radford2015unsupervised}, we represent concepts as linear projections of the embeddings $\be$: Each concept $\bc_{j}$ can be approximated as $\bc_{j}\approx C_{j}\be$.
This is modeled via a linear layer $\bc\to\be$ that implicitly inverts this relationship in the decoder.

We further model the relationships between concepts with a linear SEM:
\begin{align}
	\label{eq:concept:sem}
	\bc_{j}
	= \sum_{k=1}^{\dcon}\alpha_{kj}\bc_{k}+\beps_{k}, \quad
	\alpha_{kj}\in\R.
\end{align}
By reducing this SEM and solving for $\bc$, we deduce that
\begin{align}
	\label{eq:reduced:sem}
	\bc = A_{0}\beps,
	\WHERE
	\left\{
	\begin{aligned}
		\bc
		 & = (\bc_{1},\ldots,\bc_{\dcon})     \\
		\beps
		 & = (\beps_{1},\ldots,\beps_{\dcon})
	\end{aligned}
	\right..
\end{align}
This is known as the \emph{reduced form} of the SEM~\eqref{eq:concept:sem}.
We model this with a linear layer $\beps\to\bc$, where the weights in this layer correspond to the matrix $A_{0}$.
This layer will be used to encode the SEM between the concepts, which will be used to implement causal interventions.

\begin{remark}
	Due to the reduced form SEM above, our approach \emph{does not} and \emph{cannot} model the structural causal model encoded by the $\alpha_{kj}$.
	What is important is that the reduced form $\bc = A_{0}\beps$ still encodes causal invariances and interventions, which is enough in our setting, without directly estimating a causal graph.
\end{remark}

In principle, since the exogenous variables $\beps_{j}$ are independent, we could treat $\beps$ as the input latent space to the generative model.
Doing this, however, incurs two costs: 1) To conform to standard practice, $\beps$ would have to follow an isotropic Gaussian prior, and 2) It enforces artificial constraints on the latent dimension $\dim(\beps)$.
For this reason, we use a second expressive layer $\bz\to\beps$ that gradually transforms $\bz\sim\normalN(0,I)$ into $\beps$.
This allows for $\dim(\bz)$ to be larger and more expressive than $\dcon$ (in practice, we set $\dim(\bz)$ to be a multiple of $\dim(\beps)$), and for $\beps$ to be potentially non-Gaussian.

The final decoder architecture can be decomposed at a high-level as follows (Figure~\ref{fig:main}):
\begin{align}
	\label{eq:decoder:layers}
	\lat\xto\exo\xto\con\xto\emb\xto\obs.
\end{align}
Implementation details for each of these layers can be found in the next section.

\subsection{Details}
\label{sec:details}

We assume given a black-box encoder-decoder pair, denoted by $\enc(\bx)$ and $\dec(\bz)$, respectively.
We modify the decoder by appending a \emph{context module} to the decoder.
The context module has three layers: A \emph{representation layer}, an \emph{intervention layer}, and an \emph{expressive layer}.
\begin{enumerate}[itemsep=-5pt]
	\item The first \emph{representation layer}, as the name suggests, learns to represent concepts by implicitly inverting their linear representations $\bc_{j}=C_j\be$ from the embeddings of the encoder.
	      This is a linear layer between $\bc\to\be$  in~\eqref{eq:decoder:layers}.
	\item The \emph{intervention layer} embeds these concept representations into the reduced form SEM~\eqref{eq:reduced:sem}.
	      Each concept has its own context where it has been intervened upon.
	      A key part of this layer is how it can be used to learn and enforce interventional semantics through this shared SEM.
	      This layer corresponds to the second layer $\beps\to\bc$ in~\eqref{eq:decoder:layers}.
	\item The \emph{expressive layer} is used to reduce the (potentially very large) input latent dimension of independent Gaussian inputs down to a smaller space of non-Gaussian exogenous noise variables for the intervention layer, corresponding to the first layer $\bz\to\beps$ in~\eqref{eq:decoder:layers}.
	      This is implemented as independent, deep MLPs that gradually reduce the dimension in each layer:
	      If it is desirable to preserve Gaussianity for the exogenous variables, linear activations can be used in place of nonlinear activations (e.g., ReLU).
\end{enumerate}
\noindent
Because the intervention layer is modeled after an SEM, it is straightforward to perform latent concept interventions using the calculus of interventions in an SEM.

\begin{remark}
	Crucially, no part of this SEM is fixed or known---everything is trained end-to-end.
	In particular, we do not assume a known causal DAG or even a known causal order.
	This stands in contrast to previous work on causal generative models \citep[e.g.][]{kocaoglu2017causalgan,yang2021causalvae}.
\end{remark}

\paragraph{Concept interventions}
Assume for now that the concepts are each one-dimensional with $\dim(\bc_{j})=\dim(\beps_{j})=1$; extensions to multi-dimensional concepts are straightforward and explained below.
The structural coefficients $\alpha_{kj}\in\R$ capture direct causal effects between concepts, with $\alpha_{kj}\ne0$ indicating the presence of an edge $\bc_{k}\to\bc_{j}$.
An intervention on the $j$th concept requires deleting each incoming edge; i.e., setting $\alpha_{\cdot j}=0$, updating the outgoing edges from $\beps_{j}$, as well as replacing $\beps_{j}$ with a new $\beps_{j}^\prime$, which results from updating the incoming edges to $\beps_{j}$ in the preceding layer.
Thus, during training, when we intervene on $\bc_{j}$, we zero out the row $\alpha_{\cdot j}$ while replacing the column $\alpha_{j\cdot}$ with a new column $\beta_{j\cdot}$ that is trained specifically for this interventional setting.
The result is $\dcon+1$ intervention-specific layers: $A_{0}$ for the observational setting, and $A_{1},\ldots,A_{\dcon}$ for each interventional setting.
This yields a three-way tensor where each slice of this tensor captures a different context that corresponds to intervening on different concepts.
At inference time, to generate interventional samples we simply swap out the $A_{0}$ for $A_{j}$.
Moreover, multi-target interventions can be sampled by zeroing out multiple rows and substituting multiple $\beta_{j\cdot}$'s and $\beps_{j}$'s.

\paragraph{Multi-dimensional concepts}
Everything above goes through if we allow additional flexibility with $\dim(\bc_{j})>1$ and $\dim(\beps_{j})>1$, potentially even with $\dim(\bc_{j})\ne\dim(\beps_{j})$.
In practice, we implement this via two width parameters $\width_{\eps}=\dim(\beps_{j})$ and $\width_c=\dim(\bc_{j})$.
The design choice of enforcing uniform dimensionality is not necessary, but is made here since in our experiments there did not seem to be substantial advantages to choosing nonuniform widths.
With these modifications, $\alpha_{kj}$ becomes a $\width_{\eps}\times\width_{c}$ matrix; instead of substituting rows and columns above, we now substitute slices in the obvious way.
The three-way tensor $A$ is now $(\dcon+1)\times\width_{\eps}\dcon\times\width_{c}\dcon$-dimensional.

\paragraph{Identifiability}
An appealing aspect of this architecture is that it can be viewed as an approximation to an identifiable model over causally disentangled concepts.
Formally, we assume given observations $\bx$, concepts $\bc$, and a collection of embeddings $\be$ such that $\bx=f(\be)$.
Additionally assume a noisy version of the linear representation hypothesis, i.e., $\bc_j=C_j\be+\eps_j$ with $\eps_j\sim\normalN(0,\Omega_j)$.
Then we have the following identifiability result:
\begin{theorem}[Identifiability]
	\label{theorem:ident}
	Assume that the rows of each $C_j$ come from a linearly independent set and $f$ is injective and differentiable.
	Then, given single-node interventions on each concept $\bc_j$, the representations $C_j$ and latent concept distribution $p(\bc)$ are identifiable.
\end{theorem}
For a formal statement, see Theorem~\ref{theorem:formal} in Appendix~\ref{app:ident-prop-model}.
As opposed to solving a causal representation learning problem, the causal semantics used in this paper have been ported into a generative model that give solutions to the problem of learning latent \emph{distribution} structure, as in Theorem~\ref{theorem:ident}.
We note also that the proof of Theorem~\ref{theorem:ident} itself is nontrivial, and involves tracing the effects of interventions through the linear representations;
since our focus is on implementation and practical aspects, we defer further discussion to Appendix~\ref{app:ident-prop-model}.

\subsection{Summary of architecture}

The context module thus proposed offers the following appealing desiderata in practice:
\begin{enumerate}[itemsep=-3pt]
	\item \textbf{Black-box.} There is no coupling between the embedding dimension and the number of concepts, or the complexity of the SEM.
	      This allows for \emph{arbitrary} black-box encoder-decoder architectures to be used to learn embeddings, from which a causal model is then trained on top of.
	      This can be trained end-to-end, fine-tuned after pre-training, or even partially frozen, allowing for substantial computational savings.
	\item \textbf{Causality.} The intervention layer is a genuine causal model that provides rigorous causal semantics that allow sampling from arbitrary concept interventions, including interventions that have not been seen during training.
	\item \textbf{Identifiability.} The architecture is based on an approximation to an identifiable concept model (Theorem~\ref{theorem:ident}), which provides formal justification for the intervention layer as well as reproducibility assurances.
	\item \textbf{Flexibility.} The context module is itself arbitrarily flexible, so there is no risk of information loss in representing concepts with the embeddings $\be$.
	      Of course, information loss is possible if we compress this layer too much (e.g., by choosing $\dcon$, $\dlat$, $\width_{\eps}$, or $\width_{c}$ too small), but this is a design choice and not a constraint of the module itself.
\end{enumerate}
\noindent
Thus the only tradeoff between representational capacity and causal semantics is design-based: The architecture itself imposes no constraints.
The causal model can be arbitrarily flexible and chosen independent of the encoder-decoder pair, which are also allowed to be arbitrary.

\section{Empirical evaluation}
\label{sec:empirical-evaluation}

Open-source Python code and instructions for reproducing the results are available at \url{https://github.com/Alex-Markham/context-module}.
We evaluate the proposed context module on four different tasks:
\begin{enumerate}[itemsep=-3pt]
	\item \textbf{Concept learning} (Section~\ref{sec:id-gen-eval}): \emph{How well does the context module learn representations that correspond to each concept?}
	      To measure this, we generate interventional samples by intervening on a single concept, and directly compare these interventional samples to held-out samples using the sliced Wasserstein (SW) metric~\eqref{eq:sw}.
	\item \textbf{Composition} (Section~\ref{sec:ood-gen-eval}): \emph{How well does the context module learn to \emph{compose} concepts together in novel ways that have not been seen during training?}
	      To measure this, we generate samples by intervening on multiple concepts, and directly compare these new samples to held-out samples using the SW metric.
	      By design, interventions on multiple concepts are never seen during training, so this is genuinely OOD.
	\item \textbf{Reconstruction} (Section~\ref{sec:reconstr-eval}): \emph{How well does the context module reconstruct samples from each context?}
	      This can be evaluated using standard measures of generation quality; we use bits per dimension (BPD) for comparison with prior work.
	      Since this is computed across all contexts, this captures the model's ability to reconstruct samples from diverse contexts.
	\item \textbf{Fine-tuning} (Section~\ref{sec:fine-tuning-eval}): \emph{Can pre-trained models leverage the context module via fine-tuning to improve their learned representations and enable compositional generation?}
	      Instead of training end-to-end, we start with a pre-trained model and fine-tune it with the context module attached.
	      This is evaluated on the concept learning, composition, and reconstruction tasks.
\end{enumerate}

\noindent
For concept learning and composition, including OOD generation (Section~\ref{sec:prelim}), we report the sliced Wasserstein (SW) distance \citep{bonneel2015sliced,flamary2021pot} between a held-out sample from the ground-truth distribution and a sample generated (not reconstructed) by the trained model.
For reconstruction, we report the standard ELBO loss \citep{kingma2013auto} in bits per dimension.

\paragraph{Methods, ablations, baselines}
We implement three variants of the proposed context module: 1) Full end-to-end training; 2) Fine-tuning all weights from a pre-trained base model; 3) Fine-tuning only the context module weights while leaving the pre-trained base model frozen.
We evaluate our approach by augmenting two base architectures---a lightweight convolutional \(\beta\)-VAE \citep{higgins2017beta} and a deep hierarchical network, NVAE \citep{vahdat2020nvae}---with our context module (CM).
This allows us to compare directly to ablations using the base architectures without the context module: 1) \textbf{base}, which uses only observational data; 2) \textbf{base pooled}, which pools all contexts together into a single dataset; and 3) \textbf{CM pooled}, which pools all contexts together into a single context using the context module.
Thus, for each dataset, every aspect of the ablation is controlled: architecture, hyperparameters, random seeds, training protocol, etc.
The only difference is whether or not the context module is inserted and how many contexts are used.
We can therefore distinguish performance differences due to data diversity and/or architectural complexity.
We additionally include comparisons to three other methods: Ada-GVAE~\citep{adagvae}, \(\beta\)TCVAE~\citep{chen2018isolating}, and TVAE~\citep{tvae}.

\paragraph{Datasets}
We compare these models to our context module on four datasets:
3DIdent~\citep{zimmermann2021contrastive}, Morpho MNIST~\citep{castro2019morpho}, and two versions of a new semi-synthetic environment called \texttt{quad}---one with independent concepts and another with (causally) dependent concepts.
While 3DIdent and Morpho MNIST are established baselines, \texttt{quad} is a novel visual environment with intervenable latents (colour, shape, size, orientation) that explicitly enables sampling from ground-truth OOD composed contexts.
See Appendix~\ref{app:quad} for more details on this dataset.
We introduce \texttt{quad} in order to carefully evaluate model performance in a controlled environment over different random seeds (and hence assess statistical variability).
Thus, our evaluations use \texttt{quad} to assess reproducibility and statistical significance, whereas 3DIdent and MNIST are more appropriate for testing our approach on downstream tasks such as image generation.

\paragraph{}
Due to space constraints, we report only a representative slice of the experiments here;
full details and additional experiments can be found in Appendices~\ref{app:quad}--\ref{app:additional-results}.
This includes detailed performance breakdowns across contexts; the effects of regularization and architectural capacity; and additional figures.

\subsection{Concept learning}
\label{sec:id-gen-eval}

Table~\ref{tab:one-table-to-rule-them-all}, in the \textbf{Concept} columns, shows a quantitative evaluation of concept learning on the \texttt{quad}, MNIST, and 3DIdent datasets.
We draw three main conclusions from this table, by comparing our context module variants to each of the other methods (rows) in the ablations and baselines.

First, comparing the three \textbf{CM} rows to the ablations in Table~\ref{tab:one-table-to-rule-them-all} (specifically, \textbf{base} and \textbf{base pooled}), we see that the context module excels at learning individual concepts across all datasets and models, despite the base models achieving slightly better average reconstruction performance (see Section~\ref{sec:reconstr-eval}).
Moreover, Figure~\ref{fig:combo_comparison} shows that there is essentially no perceptual difference between samples from the base models and the context module.
Additionally, across multiple training runs over different random seeds, our context module consistently outperforms the base models across \texttt{quad} concepts---in this case the simpler model/data allows more replicates to be run to evaluate statistical significance and standard errors.

Second, comparing the \textbf{CM} rows to the \textbf{base} row of Table~\ref{tab:one-table-to-rule-them-all}, we see that the context module outperforms the base model trained only on the observational (rather than pooled) data.
This confirms that the context module is indeed successfully exploiting learned invariances across the different contexts, providing it with performance and computational resource advantages compared to models trained independently on the different contexts.

\begin{table}[t]
	\centering
	\caption{Context module (CM) vs.~fine-tuning, ablations, and baselines on three tasks across four datasets.
		Best performance per task/dataset in bold.
		Base model indicated by $^{*} =$ lightweight VAE, $^{**} =$ NVAE.}
	\label{tab:one-table-to-rule-them-all}
	\resizebox{\textwidth}{!}{\begin{tabular}{c r *{3}{c} *{3}{c} *{3}{c} *{3}{c}}
\toprule
& & \multicolumn{3}{c}{\textbf{\texttt{quad}* (independent)}} & \multicolumn{3}{c}{\textbf{\texttt{quad}* (dependent)}} & \multicolumn{3}{c}{\textbf{MNIST**}} & \multicolumn{3}{c}{\textbf{3DIdent**}} \\
\cmidrule(lr){3-5} \cmidrule(lr){6-8} \cmidrule(lr){9-11} \cmidrule(lr){12-14}
& & \textbf{Concept} & \textbf{Compo} & \textbf{Recon} & \textbf{Concept} & \textbf{Compo} & \textbf{Recon} & \textbf{Concept} & \textbf{Compo} & \textbf{Recon} & \textbf{Concept} & \textbf{Compo} & \textbf{Recon} \\
\midrule
\multirow{3}{*}{\rotatebox{90}{\scriptsize\textbf{ours}}} & \textbf{CM (end-to-end)} & 0.063 & \textbf{0.081} & 0.519 & 0.116 & 0.127 & 0.828 & \textbf{0.035} & - & \textbf{0.135} & \textbf{0.014} & \textbf{0.051} & 0.528 \\
 & \textbf{CM (fine-tune)} & \textbf{0.059} & 0.086 & 0.518 & \textbf{0.112} & \textbf{0.121} & 0.840 & 0.037 & - & 0.136 & 0.031 & 0.063 & 0.549 \\
 & \textbf{CM (frozen)} & 0.086 & 0.132 & 0.559 & 0.121 & 0.134 & 0.889 & 0.052 & - & 0.161 & 0.048 & 0.063 & 0.609 \\
\addlinespace\midrule\addlinespace
\multirow{3}{*}{\rotatebox{90}{\scriptsize\textbf{ablations}}} & \textbf{base} & 0.175 & 0.288 & \textbf{0.444} & 0.170 & 0.222 & 0.767 & 0.102 & - & 0.271 & 0.056 & 0.092 & 4.966 \\
 & \textbf{base pooled} & 0.166 & 0.248 & 0.446 & 0.167 & 0.212 & \textbf{0.746} & 0.105 & - & 0.139 & 0.049 & 0.056 & 0.559 \\
 & \textbf{CM pooled} & 0.202 & 0.306 & 0.840 & 0.166 & 0.203 & 0.752 & 0.099 & - & 0.143 & 0.041 & 0.068 & \textbf{0.516} \\
\addlinespace\addlinespace
\multirow{3}{*}{\rotatebox{90}{\scriptsize\textbf{baselines}}} & \textbf{Ada-GVAE} & 0.335 & 0.355 & 2.017 & 0.375 & 0.377 & 2.443 & 0.500 & - & 1.035 & 0.249 & 0.251 & 3.267 \\
 & \textbf{BetaTCVAE} & 0.330 & 0.346 & 1.973 & 0.366 & 0.374 & 2.408 & 0.484 & - & 1.036 & 0.257 & 0.254 & 3.261 \\
 & \textbf{TVAE} & 0.334 & 0.341 & 2.048 & 0.370 & 0.368 & 2.455 & 0.486 & - & 1.037 & 0.254 & 0.256 & 3.282 \\
\bottomrule
\end{tabular}}
\end{table}

Third, comparing the \textbf{CM} rows to the ablation in \textbf{\ablation}---which pools all the data into a single context in order to circumvent the context module's intervention mechanism while leaving the rest of the context module architecture intact---we see that \textbf{CM} consistently outperforms this ablation.
This demonstrates that the increased performance of the context module is due our module's intervention mechanism and its resulting ability to better leverage context-separated data, as opposed to spuriously benefiting from a different architecture or increased complexity.

Finally, for a qualitative evaluation, we look at Figure~\ref{fig:combo_comparison}, in the {\small\textsf{Concept 1}} and {\small\textsf{Concept 2}} rows, to see concept learning in CM+NVAE across three different datasets.
For example, looking at the 3DIdent results in Figure~\ref{fig:combo_comparison}~(middle), the {\small\textsf{Observation}} row shows the learned observational distribution of images to contain background colours and object colours both in the orange--green range (\obscolor).
The {\small\textsf{Concept 1}} row, produced by intervening on the learned concept of background colour, shows that the background colour is shifted to the blue--red range while other features, like object colour, remain invariant with the observational samples.
Analogously, the {\small\textsf{Concept 2}} row shows a shift of object colour to the blue--red range (\ivncolor) while the other features remain invariant.
These illustrate that CM+NVAE successfully learned the concepts of background colour and object colour, in line with the notion of causal disentanglement in Definition~\ref{defn:disentangled}.

\subsection{Concept composition}
\label{sec:ood-gen-eval}

\begin{figure}[t]
	\centering
	\includegraphics[height=0.28\linewidth]{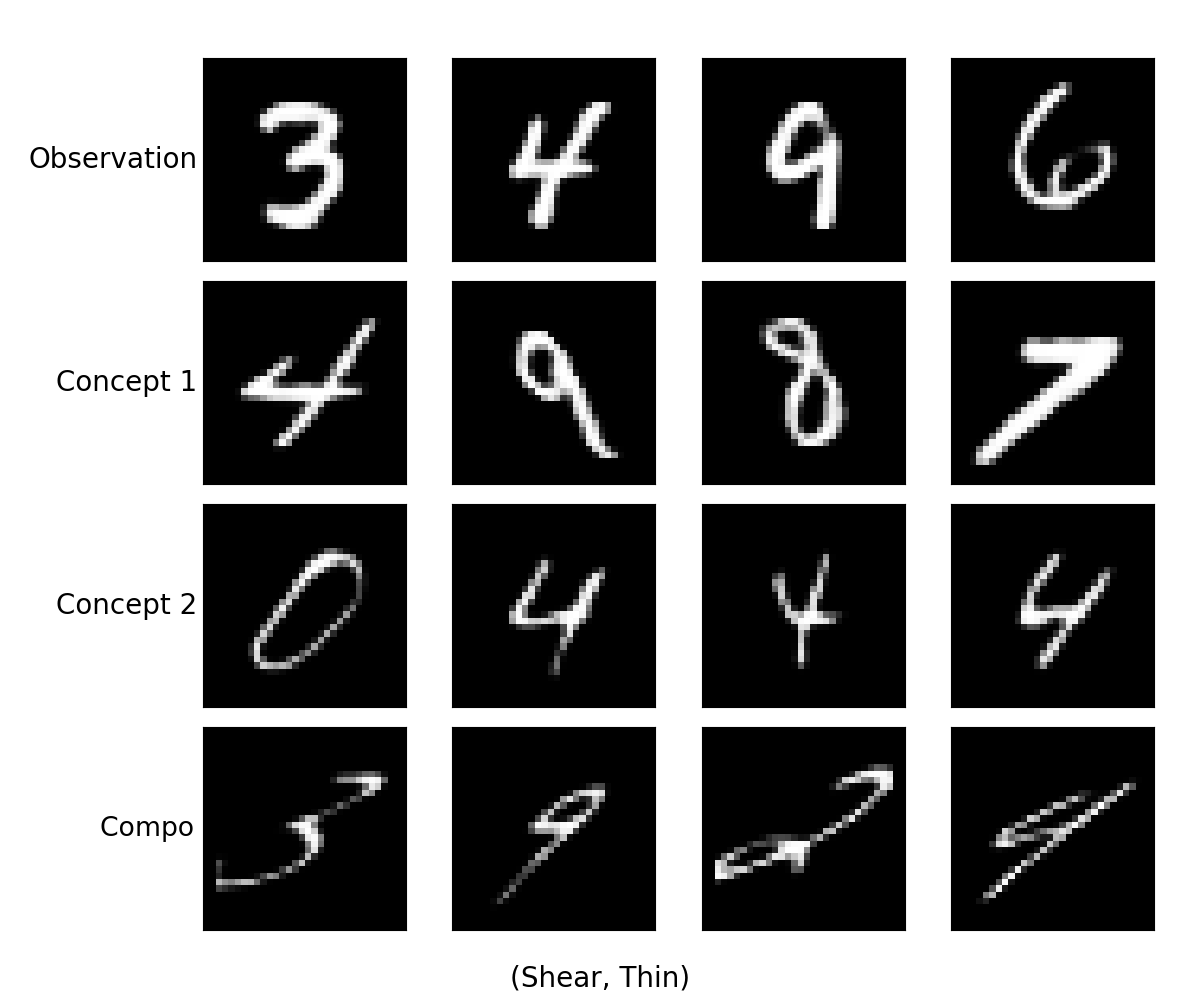}\hspace{-0em}
	\includegraphics[height=0.28\linewidth]{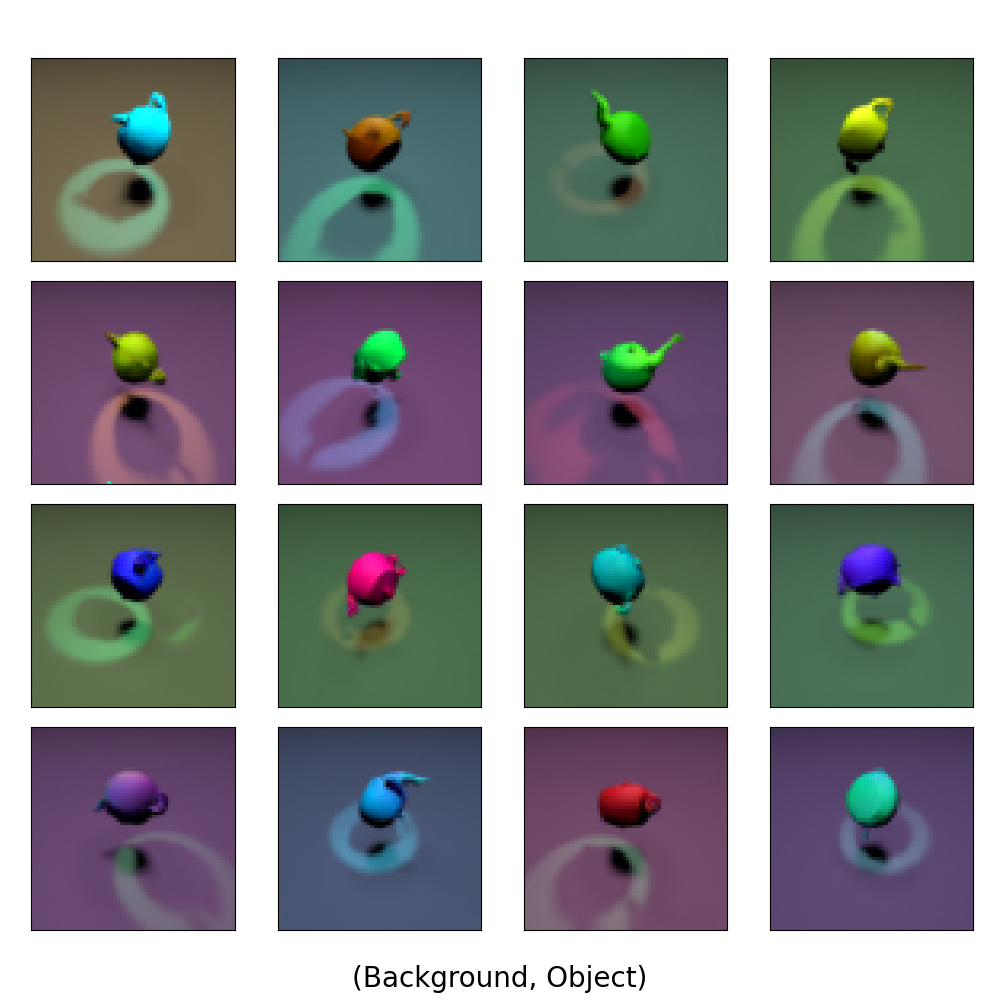}\hspace{0.5em}
	\includegraphics[height=0.28\linewidth]{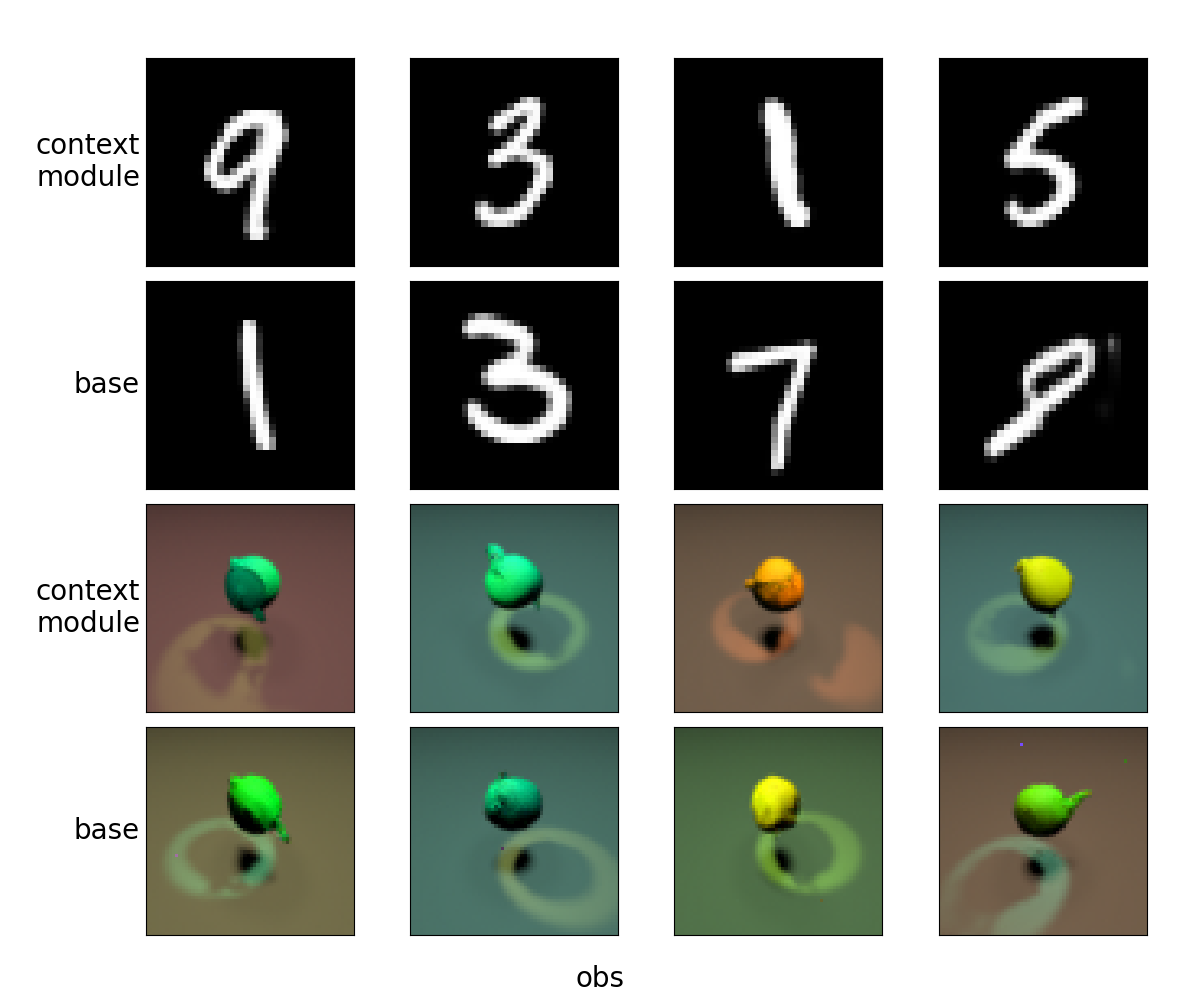}\\[0.5ex]
	\makebox[0.56\linewidth][c]{\hspace{-0.5em}(a)}
	\makebox[0.28\linewidth][c]{\hspace{4.1em}(b)}
	\caption{(a) Examples of concept learning and composition in MNIST (left) and 3DIdent (right) using CM (end-to-end), our context module augmenting the base NVAE---these images are \emph{generated}, not reconstructed.
	The first row shows generated observational samples;
	The second and third rows show samples of \emph{learned concepts}, generated by intervening on individual concepts, e.g., `{\small\textsf{Concept 1}}' is `{\small\textsf{scaled}}' for MNIST;
	The final row shows OOD \emph{composition}, generated by simultaneously intervening on pairs of learned concepts.
	(b) Observational samples from the context module vs.~the base model show no perceptual loss.
	}
	\label{fig:combo_comparison}
\end{figure}

Table~\ref{tab:one-table-to-rule-them-all}, in the \textbf{Compo} columns, provides a quantitative evaluation of concept composition on the \texttt{quad} and 3DIdent datasets, while the {\small\textsf{Compo}} row of Figure~\ref{fig:combo_comparison} provides a qualitative evaluation across the MNIST and 3DIdent datasets.
Since Morpho MNIST does not include ground truth compositional samples, we are not able to evaluate OOD generation on this dataset.
In each evaluation, the model uses the intervention layer (Section~\ref{sec:details}) to intervene on pairs of learned concepts, i.e., to compose two concepts together.
By holding out from training examples where concepts are composed, we are assured that the model has not seen any of these compositions.
The only way to compose these concepts together is through the implicit causal model (via the reduced form SEM~\eqref{eq:reduced:sem}) that is learned during training.
We evaluate model performance with the SW metric~\eqref{eq:sw}, comparing concept compositions from the model against ground-truth hold out compositions.

The same patterns in performance hold here for concept composition as we saw for concept learning in Section~\ref{sec:id-gen-eval}---namely, our context module consistently achieves better performance than baselines and ablations, demonstrating successful concept composition due to the module's intervention mechanism.
For example, looking again at the 3DIdent results in Figure~\ref{fig:combo_comparison}~(middle), we see that the colour range of both the background and the object are shifted from the orange--green range (\obscolor) to the blue--red range (\ivncolor).

These experiments demonstrate the context module's ability to learn causally disentangled representations through both concept learning and composition.

\subsection{Reconstruction}
\label{sec:reconstr-eval}

The \textbf{Recon} columns of Table~\ref{tab:one-table-to-rule-them-all} compare the validation reconstruction performance of our approach against the alternatives.
We draw two conclusions from these results.
First, adding the context module incurs only a minor quantitative cost in reconstruction performance, while granting additional concept learning and composition abilities explored in Sections~\ref{sec:id-gen-eval}-\ref{sec:ood-gen-eval}.
Second, Figure~\ref{fig:combo_comparison} demonstrates that despite this small quantitative cost, the perceptual quality of the samples is still very high and indistinguishable from the \textbf{base} model.
This validates the central motivation behind the context module, namely that it can imbue existing black-box models with additional concept learning and causal composition abilities \emph{without} sacrificing perceptual metrics or downstream performance.

\subsection{Fine-tuning}
\label{sec:fine-tuning-eval}

Finally, an important aspect of our method is that it need not be trained end-to-end but can instead be successfully fine-tuned from a frozen base mode.
The second (\textbf{CM (fine-tune)}) and third (\textbf{CM (frozen)}) rows of Table~\ref{tab:one-table-to-rule-them-all} evaluate the context module on concept leaning and composition using two fine-tuning variants.
Both variants start with a pre-trained model and then insert the context module and resume training, either only updating the context module weights in the \textbf{frozen} variant, or also fine-tuning the non-context module weights in the \textbf{fine-tune} variant.

We see that the fine-tuned models consistently outperform both the \textbf{base} variants as well as the baselines, despite minor fluctuations in performance compared to end-to-end training (\textbf{CM (end-to-end)}).
Hence, both fine-tuning variants offer computational savings (ranging from 2--10\(\times\) in GPU hours) compared to the full end-to-end training while still substantially outperforming all other baselines and ablations in Table~\ref{tab:one-table-to-rule-them-all}.
Choosing between end-to-end training and the two variants allows for a tradeoff between computational cost and performance, all with improved representations and OOD generation abilities compared to the alternatives.

\section{Conclusion and Limitations}
\label{sec:concl-limit}

Designing generative models with structured latent spaces that enable intervention, composition, and OOD generation is an outstanding challenge.
Existing approaches either sacrifice structure for flexibility, or flexibility for structure.
We argue that such a tradeoff is not necessary and that arbitrarily flexible models can indeed learn useful structure.

To this end, we propose a simple context module that can be appended to a black-box decoder that learns causally disentangled latent spaces.
The context module enables genuinely OOD, compositional generation without hardcoding prior knowledge or structure into the model, and it avoids the difficult problem of latent causal discovery.

Our experiments provide both quantitative and qualitative evidence for this.
While existing evaluations have focused on reconstruction as a metric to quantify OOD performance, we argue that OOD generation is in fact more appropriate since it also measures the ability of a model to learn useful concepts that can be manipulated at inference time.

Since our module imposes no hard constraints, it may be useful in combination with other modules.
Understanding how the context module interacts with other architectures is a natural next step and a promising direction for future work.
We also encourage the community to explore alternative metrics for evaluating genuine OOD generalization.

Finally, understanding the cost and tradeoffs of variational approximations compared to, say, exact maximum likelihood estimation is important, especially in causal applications.
Indeed, this can make estimating the exact causal graph much more challenging.
This motivates our fundamental design principle: to use a ``causal enough'' architecture to obtain composability and OOD generation while otherwise leveraging already empirically proven non-causal architectures.

\subsubsection*{Acknowledgments}
The work by AM was supported by Novo Nordisk Foundation Grant NNF20OC0062897.
This research was supported in part through the computational resources and staff contributions provided for the Mercury high performance computing cluster at The University of Chicago Booth School of Business which is supported by the Office of the Dean.

\bibliography{main}
\graphicspath{{figures/}}

\clearpage
\appendix

\begin{center}
	\Large\textbf{
		Supplementary material: \\
		Intervening to learn and compose\\causally disentangled representations}
\end{center}

\section{Related work}
\label{sec:related}

Our work is closely related to several parallel lines of work on structured generative models, causal representation learning, linear representations, and OOD generalization.
Below we compare and contrast our contributions against this literature.

\paragraph{Structured generative models}
To provide structure such as hierarchical, graphical, causal, and disentangled structures as well as other inductive biases in the latent space, there has been a trend towards building \emph{structured} generative models that directly impose this structure \emph{a priori}.
Early work looked at incorporating fixed, known structure into generative models, such as autoregressive, graphical, and hierarchical structure \citep{germain2015made,johnson2016composing,sonderby2016ladder,webb2018faithful,weilbach2020structured,ding2021gans,moutonintegrating}.
This was later translated into known \emph{causal} structure \citep{kocaoglu2017causalgan,markham2023neuro}.
When the latent structure is unknown, several techniques have been developed to learn useful (not necessarily causal) structure from data \citep{li2018learning,he2019variational,wehenkel2021graphical,kivva2022identifiability,moran2022identifiable}.
More recently, based on growing interest in disentangled \citep{bengio2013deep} and/or causal \citep{scholkopf2021toward} representation learning, methods that learn causal structure have been developed \citep{markham2020meas,moraffah2020causal,yang2021causalvae,ashman2022causal,shen2022weakly,pmlr-v202-kaltenpoth23a}.
In contrast to this line of work, which emphasizes hard graphical constraints, we neither learn nor impose any fixed graphical structure.
Instead, we emphasize performance on concept learning and composition as downstream tasks.

\paragraph{Causal disentanglement and representation learning}
Causal representation learning \citep{scholkopf2021toward} is a rapidly developing area that involves, among other goals, two key objectives: 1) Learning disentangled latent factors along with 2) Learning latent causal structure.
The former is what we refer to as \emph{causal disentanglement} (see Definition~\ref{defn:disentangled} and its causal interpretation).
The latter problem has seen tremendous theoretical progress in recent years~\citep{brehmer2022weakly, shen2022weakly, lachapelle2022disentanglement, moran2022identifiable, kivva2022identifiability,buchholz2023learning, gresele2021independent, ahuja2022interventional,von2023nonparametric,morioka2023causal,liu2023identifiable,mameche2023learning,yao2023multi,varici2023score,sturma2023unpaired,xu2024sparsity,zhang2024generalsetting,li2024causal,varici2024general,bing2024identifying,ahuja2024multidomain,talon2024towards,yao2024unifying}.
Recent work has also pushed in the direction of identifying concepts \citep{leemann2023post,rajendran2024learning,fokkema2025concepts}.
Our work draws inspiration from these lines of work, which articulate precise conditions under which latent causal discovery is possible \emph{in principle}.
By contrast, our focus is on methodological aspects of causal disentanglement \emph{in practice}.
While the former is a notoriously difficult task, the latter only needs to exploit learned causal invariances: In particular, causal disentanglement is possible without learning latent causal structure.

\paragraph{Linear representations}
Generative models are known to represent concepts linearly in embedding space \citep[e.g.][]{mikolov2013linguistic,szegedy2013intriguing,radford2015unsupervised}; see also \cite{levy2014neural, arora2016latent, gittens2017skip, allen2019analogies,ethayarajh2018towards,seonwoo2019additive}.
This phenomenon has been well-documented over the past decade in both language models \citep{mikolov2013linguistic,pennington2014glove,arora2016latent, conneau2018you,tenney2019bert, elhage2022superposition,burns2022discovering, tigges2023linear, nanda2023emergent, moschella2022relative, li2023inference, park2023linear, gurnee2023finding,cunningham2024sparse,jiang2024learning} and computer vision \citep{radford2015unsupervised, raghu2017svcca, bau2017network, engel2017latent, kim2018interpretability,chen2020concept, wang2023concept, trager2023linear}.
Our approach actively exploits this tendency by searching for concept representations as linear projections of the embeddings learned by a black-box model.
In contrast to this prior work, our emphasis is on applying these empirical observations for causal disentanglement, compositional generalization, and OOD generation.

\paragraph{OOD generalization}
A growing line of work studies the OOD generalization capabilities of generative models, with the general observation being that existing methods struggle to generalize OOD \citep{xu2022compositional,mo2024compositional,montero2022lost,montero2021role,schott2022visual}.
It is worth noting that most if not all of this work evaluates OOD generalization using reconstruction on held-out OOD samples, as opposed to generation.
For example, a traditional VAE may be able to reconstruct held-out samples, but it is not possible to actively sample OOD.
See Section~\ref{sec:ood} for more discussion.

\section{Identifiability of the proposed model}
\label{app:ident-prop-model}

In this appendix we set up and prove Theorem~4.1.

\subsection{Set-up}

Our setup is the following: Let $\bx = (\bx_1,\ldots,\bx_{\dobs}), \bc=(\bc_1,\ldots,\bc_{\dcon})$ and $\be = (\be_1,\ldots,\be_{\demb})$ be random vectors taking values in $\mathcal{X}, \mathcal{C}$ and $\mathcal{E}$, respectively.
For simplicity, we assume that $\mathcal{X} \subseteq \R^{\dobs}$, $\be\in\mathcal{E}\subseteq\R^{\demb}$, and $\bx = f(\be)$ for some embeddings $\be\in\mathcal{E}$.
We assume that $f$ is injective and differentiable.
This map is allowed to be arbitrarily non-linear: We make no additional assumptions on $f$.

The random variables $\bc_{j}$ will be referred to as ``concepts'': Intuitively they represent abstract concepts such as shape, size, colour, etc.~that may not be perfectly represented by any particular embedding in $\be$.
The \emph{linear representation hypothesis} (Section~\ref{sec:related}) is an empirical observation that abstract concepts can be approximately represented as linear projections of a sufficiently flexible latent space.
This is operationalized by assuming that $\bc_j \approx C_j\be$ (see \ref{model:linrep} below).
The matrix $C_{j}$ is referred to as the \emph{representation} of the concept $\bc_{j}$.
If $\bc_{j}\in\R$ (i.e., $C_{j}\in\R^{1\times \demb}$), then $\bc_{j}$ is referred to as an \emph{atom} and $C_{j}$ its \emph{atomic representation}.
We will often denote atoms by $\ba_{j}$ for notational clarity.

We consider $(\bx,\bc,\be)$ satisfying the following:
\begin{enumerate}[label=(A\arabic*)]
	\item\label{model:linrep} $\bc_j = C_j\be + \eta_j$ for all $j\in[\dcon]$ for some real matrices $C_1\ldots, C_{\dcon}$, where $\eta_{j}\sim \normalN(0,\Lambda_{j})$ for some diagonal matrix
	      $\Lambda_{j}\succ0$.
	\item\label{model:nonlinear} $\bx = f(\be)$ for some injective and differentiable function $f$.
	\item\label{model:scm} There is a DAG $\gr = ([{\dcon}],E)$ such that
	      \begin{align}
		      \label{eq:lin:sem}
		      \bc_{j} = \sum_{k\in\pa_{\gr}(j)}\alpha_{kj}\bc_{k}+\beps_{j},
		      \quad\text{ for all $j\in[{\dcon}]$}
	      \end{align}
	      and where $\beps_1,\ldots, \beps_{\dcon}$ are mutually independent and $\be, \eta$ are independent of $\beps = (\beps_1,\ldots,\beps_{\dcon})$.
\end{enumerate}
Note that these assumptions are not necessary \emph{in practice}---as evidenced by our experiments in Section~\ref{sec:empirical-evaluation} which clearly violate these assumptions---and are merely used here to provide a proof-of-concept identifiability result based on our architectural choices.
The representation \eqref{eq:lin:sem} implies, in particular, the usual factorization
\begin{align*}
	p(\bc_1,\ldots,\bc_{\dcon})
	= \prod_{j=1}^{\dcon}p(\bc_j\given \pa_{\gr}(j)),
\end{align*}
which formalizes the notion of causal disentanglement as in \citet{scholkopf2021toward}.

Our goal is to identify the concept marginal $p(\bc)$ and concept representations $(C_{1},\ldots,C_{\dcon})$ from concept interventions.
Concept interventions are well-defined through the structural causal model defined by the DAG $\gr$ and its structural equations.
We consider single-node concept interventions where $\bc_{j}$ is stochastically set to be centered at $\bmu_{j}$ with variance $\Omega_j\succ0$ (making this precise requires some set-up; see Appendix~\ref{sec:conceptivn} for details).

By identifiability, we mean the usual notion of identifiability up to permutation, shift, and scaling from the literature.
Namely, for any two sets of parameters $(f,C_{1},\ldots,C_{\dcon})$ and $(\wt{f},\wt{C}_{1},\ldots,\wt{C}_{\dcon})$ that generate the same observed marginal $p(\bx)$, there exists permutations $P_{j}$, diagonal scaling matrices $D_j$, and a shift $b \in \R^{\demb}$,  such that
\begin{align}\label{eq:ident1}
	C_j f^{-1}(\bx) =   D_{j} P_{j}\wt{C}_{j}(\wt{f}^{-1}(\bx) + b),
	\quad\text{for all $j$ and $\bx$.}
\end{align}
Moreover, there exists an invertible linear transformation $L$ such that
the concepts satisfy
\begin{align}\label{eq:ident2}
	C_j= P_j\wt{C}_j L^{-1}.
\end{align}
This definition of identifiability, which aligns with similar notions of identifiability that have appeared previously \citep[e.g.][]{seigal2022linear,von2023nonparametric,rajendran2024learning,fokkema2025concepts}, implies in particular that the concepts $\bc_{j}$ are identifiable up to permutation, shift, and scale, and that their representations are identifiable up to a linear transformation.
As a result, the concept marginal $p(\bc)$ is also identifiable (again up to permutation, shift, and scale as well).
The shift ambiguity can of course be resolved by assuming the concepts are centered (zero-mean), and the scale ambiguity can be resolved by assuming concepts are normalized (e.g., unit norm).

We make the following assumptions, which are adapted from \citet{rajendran2024learning} to the present setting.
Note that \citet{rajendran2024learning} do not consider a structural causal model over $\bc$, and thus the notion of a concept intervention there is not well-defined.

\begin{assumption}[Atomic representations]\label{assumption:atomic}
	There exists a set of atomic representations $\mathcal{A} = \{\ba_1, \ldots, \ba_{\natom}\}$ of linearly independent vectors $\ba_j\in\R^{\demb}$ such that the rows of each representation $C_{j}$ are in $\mathcal{A}$.
	We denote the indices of $\mathcal{A}$
	that appear as rows of $C_{j}$ by $S^j$, i.e., $S^{j}=\{i:\ba_{i}\text{ is a row in $C_{j}$}\}$.
	We assume that $\cup_j S^j=\mathcal{A}$, i.e., every atom in $\mathcal{A}$ appears in some concept representation $C_j$.
\end{assumption}

Define a matrix $M\in \R^{\natom\times\dcon}$ to track which concepts use which atoms, i.e.
\begin{align}\label{eq:env_conc}
	M_{ij} = \begin{cases}
		         \frac{1}{\omega_{j}^2} & \text{if $i\in S^j$}
		         \\
		         0                      & \text{otherwise,}
	         \end{cases}
\end{align}
where $\omega_{j}^2$ are the diagonal entries of $\Omega_{j}$.
Similarly, we define a matrix $B\in \R^{\natom\times\dcon}$ given by
\begin{align}\label{eq:env_valuation}
	B_{ij} = \begin{cases}
		         \frac{(\bmu_j)_{k}}{\omega_{j}^2} & \text{if $i\in S^j$ and the $k$th row of $C_{j}$ is $\ba_{i}$,}
		         \\
		         0                                 & \text{otherwise.}
	         \end{cases}
\end{align}
The second assumption ensures that the concepts and associated interventional environments are sufficiently diverse.
\begin{assumption}[Concept diversity]\label{assumption:div}
	For every pair of atoms $\ba_i$ and $\ba_j$ with $i\neq j$ there is a concept $\bc_{k}$ such that $i\in S^k$ and $j\notin S^k$.
	Furthermore, $\rank(M)=\natom$ and there exists $v\in \R^{\dcon}$ such that $v^\top M=0$ and $(v^\top B)_{i}\ne0$ for each coordinate $i$.
\end{assumption}

\subsection{Formal statement of Theorem~4.1}

With this setup, we can now state and prove a more formal version of Theorem~4.1.

\begin{theorem}
	\label{theorem:formal}
	Under Assumptions~\ref{assumption:atomic}-\ref{assumption:div} and given single-node interventions on each concept $\bc_j$, we can identify the representations $C_j$ and the latent concept distribution $p(\bc)$,
	in the sense of (\ref{eq:ident1}-\ref{eq:ident2}).
\end{theorem}

In the context of learning disentangled representations, identifying the concept marginal $p(\bc)$ is particularly relevant, since this implies we also learn certain (in)dependence relationships between the concepts.
In other words, if the ``true'' concept representations are disentangled (i.e., independent), then we identify disentangled concepts.

In relation to existing work, we make the following remarks:
\begin{itemize}
	\item While our proof ultimately relies on techniques from \citet{rajendran2024learning}, Theorem~\ref{theorem:formal} does not trivially follow from their results, which do not cover interventions over the concept marginal $p(\bc)$.
	      Extending their results to the case of concept interventions, including making sure these interventions are well-defined causal objects, is the main technical difficulty in our proof.
	      This is surprisingly tricky since we do not assume a full structural causal model over $(\bx,\bc,\be)$: Interventions are only well-defined over $\bc$, which makes deducing the downstream effects on the observed $\bx$ somewhat subtle.
	\item Various results in CRL \citep[e.g.][see Section~\ref{sec:related} for more references]{buchholz2023learning,von2023nonparametric,varici2024general} prove identifiability results under interventions, however, these results require interventions directly on the embeddings $\be$, as opposed to the concepts $\bc$.
	      Theorem~\ref{theorem:formal}, by contrast, requires only $\dcon \ll \demb$ total interventions and in doing so directly addresses the technical challenges associated with intervening directly on concepts as opposed to embeddings.
	\item \citet{leemann2023post,fokkema2025concepts} study concept identifiability when the nonlinear function $f$ is known.
	      We avoid this simplifying assumption (sometimes called \emph{post-hoc} identifiability), and dealing with the potential nonidentifiability of $f$ is another technical complication our setting must deal with.
	      Indeed, in our setting we can only identify the behaviour of $f$ on the concept subspaces defined by $C_{j}$, which is much weaker than identifying all of $f$ on $\mathcal{E}$.
	      On the other hand, we have left finite-sample aspects to future work.
\end{itemize}

\subsection{Proof of Theorem~4.1}

To prove Theorem~4.1, we begin with some notation.
Let $\mathcal{M}$ denote the collection of all distributions $P$ over $(\bx,\bc,\be)$ satisfying \ref{model:linrep}-\ref{model:scm} for some choice of $C_1,\ldots,C_{\dcon}$, $\eta$, $\gr$, $f_1,\ldots, f_{\dcon}$, and $\beps$.
For a fixed $\gr$, let
\begin{align}
	\mathcal{M}(\gr)
	:=\{P = P(\bx,\bc,\be)\in\mathcal{M} : \text{$P$ satisfies \ref{model:scm} with $\gr$}\}.
\end{align}
The model $\mathcal{M}(\gr)$ is nonempty for any $\gr$, as can be seen by letting $\be, \eta$ and $\beps$ be normally distributed and taking all functions to be linear.
Moreover, $\mathcal{M}=\cup_{\gr}\mathcal{M}(\gr)$.
Finally, let $\mathcal{M}_{\bc}(\gr)$ denote the collection of all marginal distributions over $\bc$ induced by the distributions in $\mathcal{M}(\gr)$.

\subsubsection{Concept interventions}
\label{sec:conceptivn}

In practice, the edge structure of the graph $\gr$ is unknown, and we do not assume this is known in advance.
We work in the setting where we can only observe outcomes of the variables $\bx$.
However, we allow for the possibility to intervene on the variables $\bc_1,\ldots, \bc_{\dcon}$.
We consider consider only interventional distributions with single-node targets $I=\{j\}$ for $j\in[m]$.
Note that, by definition of $\mathcal{M}_{\bc}(\gr)$, if $P\in \mathcal{M}_{\bc}(\gr)$ then there is a distribution $\widetilde{P}\in\mathcal{M}(\gr)$ such that $\widetilde{p}(\bc) = p(\bc)$ for all $\bc\in\mathcal{C}$.
Although immediate from these definitions, we record this as a lemma for future use:
\begin{lemma}
	\label{lem:mpd2jpd}
	If $P\in \mathcal{M}_{\bc}(\gr)$ then there is a distribution $\widetilde{P}\in\mathcal{M}(\gr)$ such that $\widetilde{p}(\bc) = p(\bc)$ for all $\bc\in\mathcal{C}$.
\end{lemma}

For the target $I=\{j\}$ and distribution $P\in\mathcal{M}_{\bc}(\gr)$ we say that $P^{j}$ is an \emph{interventional distribution} for $j$ and $P$ if in $P^{j}$ we have the following identities:
\begin{equation}
	\label{eqn:intervention}
	\begin{split}
		\bc_{i} & = \sum_{k\in\pa_{\gr}(i)}\alpha_{ki}\bc_{k}+\beps_{i}, \\
		\bc_{j} & = \bmu_{j} + \eta_j',
		\quad\WHERE \eta_j'\sim\normalN(0,\omega_j^2).
	\end{split}
\end{equation}
That is, the structural equations defining $\bc_i$ for $i\ne j$ are equal to those defining $\bc_i$ in the observational distribution $P$, while the distribution of $\bc_j$ is manipulated to be $\normalN(\bmu_j,\omega_j^2)$, independent of the other concepts.

The following is obvious from our assumptions:
\begin{lemma}
	\label{lem:invmarkov}
	For any distribution $P\in\mathcal{M}_{\bc}(\gr)$, the interventional distribution $P^{j}$ defined by \eqref{eqn:intervention} is also an element of $\mathcal{M}_{\bc}(\gr)$, i.e., $P^{j}\in\mathcal{M}_{\bc}(\gr)$.
\end{lemma}

Thus, we have the usual, well-defined notion of intervention over the concepts $\bc$.
The next step is to translate the effect of these interventions onto $\be$ and $\bx$.

The following proposition shows that an interventional distribution $P^{j}\in\mathcal{M}_{\bc}(\gr)$ for $P\in\mathcal{M}_{\bc}(\gr)$ always extends to a well-defined post-interventional joint distribution over $(\bx,\bc,\be)$ that can be interpreted as an interventional distribution for the appropriate choice of joint observational distribution over $(\bx,\bc,\be)$.
\begin{proposition}
	\label{prop: ivnextend}
	Let $P^{j}$ be an interventional distribution for $j$ and $P\in\mathcal{M}_{\bc}(\gr)$.
	Then there exists a distribution $\widetilde{P}^{j}\in\mathcal{M}(\gr)$ such that
	\begin{enumerate}
		\item $\widetilde{p}^{j}(\bc) = p^{j}(\bc)$ for all $\bc\in\mathcal{C}$,
		\item $\widetilde{p}^{j}(\be) = \widetilde{p}(\be)$ for all $\be\in\mathcal{E}$,
		\item If $\widetilde{P}$ is induced by the functional relation $\bx = f(\be)$, and similarly $\widetilde{P}^{j}$ with $\bx =f^{j}(\be)$ then $f^{j} = f$.
	\end{enumerate}
\end{proposition}

\begin{proof}
	Lemma~\ref{lem:invmarkov} implies that $P^{j}\in\mathcal{M}_{\bc}(\gr)$.
	Thus Lemma~\ref{lem:mpd2jpd} implies the existence of joint distributions $\wt{P},\wt{P}^{j}\in\mathcal{M}(\gr)$ such that
	\begin{align}
		\label{prop: tautology:1}
		\widetilde{p}(\bc) = p(\bc)
		\quad\AND\quad
		\widetilde{p}^{j}(\bc) = p^{j}(\bc)
		\text{ for all }\bc\in\mathcal{C}.
	\end{align}
	(1) follows from the definition of $\widetilde{P}^{j}\in\mathcal{M}(\gr)$ for $P^{j}\in \mathcal{M}_{\bc}(\gr)$ from Lemma~\ref{lem:mpd2jpd} above; i.e., \eqref{prop: tautology:1} above.

	(2) follows from the definition of an interventional distribution and the factorization
	\[
		\widetilde{p}(\bx,\bc,\be) = \widetilde{p}(\bx \mid \be) \widetilde{p}(\bc \mid \be)\widetilde{p}(\be),
	\]
	satisfied by any distribution $P\in \mathcal{M}(\gr)$.
	Namely, the interventional distribution $P^{j}$ satisfies
	\[
		\begin{split}
			\widetilde{p}^{j}(\bx,\bc,\be)
			 & = \widetilde{p}^{j}(\bx \mid \be) \widetilde{p}^{j}(\bc \mid \be)\widetilde{p}^{j}(\be), \\
			 & = \widetilde{p}^{j}(\bx \mid \be) {p}^{j}(\bc \mid \be)\widetilde{p}^{j}(\be),           \\
		\end{split}
	\]
	according to the definition of $\widetilde{P}$.
	Since the intervention only perturbs the conditional factors $\widetilde{p}(\bc_j\mid \bc_{\textrm{pa}_\gr(j)}, \be)$ in $\widetilde{p}(\bc\mid \be)$ for $j$, we can always choose $\widetilde{P}$ and $\widetilde{P}^{j}$ such that $\widetilde{p}^{j}(\be) = \widetilde{p}(\be)$.

	(3) follows similarly to (2).
	Namely, since the intervention only perturbs the conditional factors specified above, we can always choose $f^{j} = f$.
\end{proof}

We additionally have the following:
\begin{proposition}
	\label{prop: multiple interventions}
	Let $I_j:=\{j\}$ be the collection of single-node intervention targets, and consider the interventional distributions $P^{1},\ldots, P^{\dcon}$ for $P\in\mathcal{M}_{\bc}(\gr)$.
	Then there exists $\widetilde{P},\widetilde{P}^{1},\ldots, \widetilde{P}^{\dcon}\in\mathcal{M}(\gr)$ such that
	\[
		\widetilde{p}^{j}(\be)
		= \widetilde{p}(\be)
		\qquad \textrm{ for all $j\in [\dcon]$}
	\]
	and $f^{j}=f$ for all $j$.
\end{proposition}

\begin{proof}
	Fix a choice of $\widetilde{P}\in \mathcal{M}(\gr)$ for $P$, which specifies a function $f$.
	For the chosen $\widetilde{P}$, we construct $\widetilde{P},\widetilde{P}^{1},\ldots, \widetilde{P}^{\dcon}$ according to Proposition~\ref{prop: ivnextend}.
	The result follows by applying Proposition~\ref{prop: ivnextend} for every $j$.
\end{proof}

Proposition~\ref{prop: multiple interventions} will allow us (below) to define more formally the environments that are used to identify concepts.
Intuitively, each environment corresponds to single-node interventions on a single concept $\bc_{j}$, however, we can only observe the effect this intervention has on the observed $\bx$.
The purpose of Propositions~\ref{prop: ivnextend}-\ref{prop: multiple interventions} are to trace the dependence between $\bx$ and $\bc$---via the embeddings $\be$---according to the model \ref{model:linrep}-\ref{model:scm}.

Formally, let $I_{j}=\{j\}$ with $\eta_j'\sim\normalN(0,\omega_{j}^{2})$ and $\bc_j = \bmu_j+\eta_j'$ according to \eqref{eqn:intervention}, i.e., each environment corresponds to a single-node intervention on the $j$th concept $\bc_{j}$.
By Proposition~\ref{prop: multiple interventions}, there exist post-intervention joint distributions $\wt{P}^{j}\in\mathcal{M}(\gr)$ over $(\bx,\bc,\be)$ such that
\[
	\widetilde{p}^{j}(\be)
	= \widetilde{p}(\be)
	\qquad \textrm{ for all $j\in [\dcon]$}
\]
and $f^{j}=f$ for all $j$.
Then the $j$th environment corresponds to sampling $\bx=f(\be)$ where $\be\sim p_{j}(\be) := \wt{p}^{j}(\be\given\bc_{j}=\bmu_{j})$.

\subsubsection{Reduction to concept identification}

We begin by computing the log-likelihood ratio between the observational and interventional environments.
The idea is that the concept representations are revealed through fluctuations in this likelihood ratio.
Using Proposition~\ref{prop: multiple interventions}, we have
\begin{align}
	\log p_{0}(\be) - \log p_{j}(\be)
	 & \propto \log p_{0}(\be) - \log \wt{p}^{j}(\bc_{j}=\bmu_{j}\given \be) - \log \wt{p}^{j}(\be) \\
	 & \propto -\tfrac1{2}(\bmu_{j} - C_{j}\be)^T\Omega_{j}^{-1}(\bmu_{j} - C_{j}\be)               \\
	 & \propto \sum_{i=1}^{n}\frac12 M_{ij}(\ba_{i}^{T}\be)^{2} - B_{ij}\ba_{i}^{T}\be + O(1).
\end{align}
From here, we proceed as in \citet{rajendran2024learning} (note that in their notation, $\be=\bz$).
For completeness, we sketch the main steps here.
The basic idea is to analyze the system of $\dcon$ equations induced by the log-likelihood ratios for each concept intervention.
First, we reduce the system to a standard form by transforming the atoms into the standard basis via a change of coordinates.
After this change of coordinates, the system can be rewritten as
\begin{align}
	h_{j}(\be)
	:= \log p_{0}(\be) - \log p_{j}(\be)
	 & \propto \sum_{i=1}^{n}\frac12 M_{ij}\be_{i}^{2} - B_{ij}\be_{i} + O(1).
\end{align}
These functions are convex, which allows us to identify the $|S_{T}|$ where $S_{T}=\cup_{j\in T}S_{j}$ by minimizing the convex function $\sum_{j\in T}h_{j}(\be)$.
Then a similar induction argument identifies the matrices $M$ and $B$.
Now, given two distinct representations of $p(\bx)$ via nonlinear mixing functions $f$ and $\wt{f}$, define $\phi=\wt{f}^{-1}\circ f$.
Let $\mathbf{h}=(h_{1},\ldots,h_{\natom})$ and write $\be^{c}=(\be_{1},\ldots,\be_{\natom})$ and $\be=(\be^{c},\be^{\perp})\in\R^{\natom\times(\dcon-\natom)}$.
Let $\iota^{\perp}(\be^{c})=(\be^{c},0)$, $\pi^{c}(\be)=\be^{c}$, and $\phi^{\perp}:\R^{\natom}\to\R^{\natom}$ by $\phi^{\perp}(\be^{c})_{i}=\phi(\be,0)_{i}$.
Note that via the change of variables formula, we can write $\mathbf{h}(\be)=\mathbf{H}(\bx)$ for some function $\mathbf{H}$.
Then
\begin{align}
	\mathbf{h}(\be^{c},0)
	= \mathbf{H}(f(\be^{c},0))
	= \mathbf{H}(\wt{g}(\phi^{\perp}(\be^{c})))
	= \mathbf{h}(\phi^{\perp}(\be^{c})),
\end{align}
which is the crucial relation used in the proof of Theorem~1 of \citet{rajendran2024learning}.
From here, identifiability follows from a straightforward but lengthy linear algebraic argument based on the first-order conditions for minimizing $\mathbf{h}$ and the identifiability of $M,B$ proved above.
The proof of Theorem~\ref{theorem:formal} follows.

\section{\texttt{quad}: A semi-synthetic benchmark for compositional generation}
\label{app:quad}

\begin{figure}[t]
	\centering
	\includegraphics[width=\textwidth]{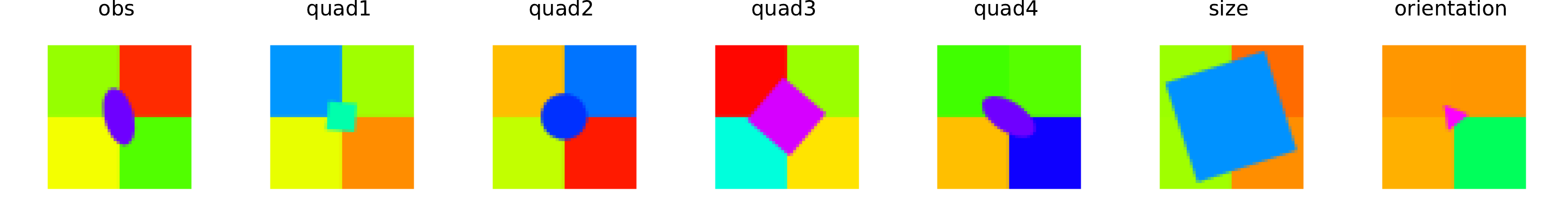}
	\caption{Example images obtained from the \texttt{quad} dataset.
		Each subfigure corresponds to a different context, indicated by the title (\texttt{obs} = observational; the others indicate a single-node intervention).
		For example, in \texttt{quad1}, the first (top left) quadrant has been manipulated from orange-green hues to blue-red hues.}
	\label{fig:quad_single_examples}
\end{figure}

\begin{figure}[t]
	\centering
	\includegraphics[width=\textwidth]{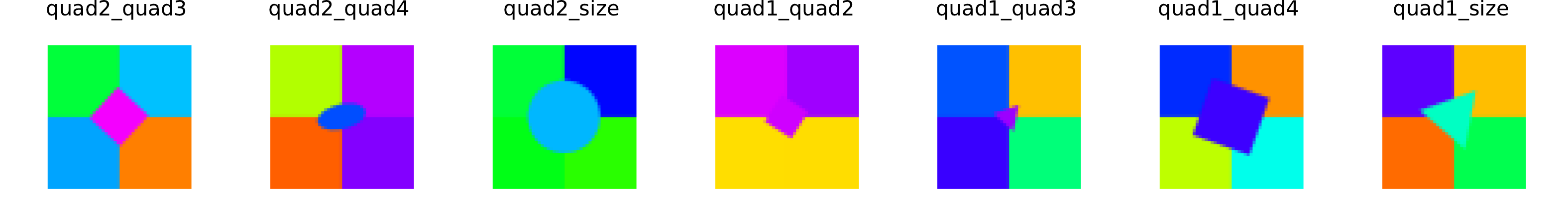}
	\caption{Example images obtained from the \texttt{quad} dataset.
		Each subfigure corresponds to a different double-concept intervention context, indicated by the title.
		For example, in \texttt{quad2\_quad3}, the second (top-right) and third (bottom-left) quadrants have been manipulated from orange-green hues to blue-red hues.}
	\label{fig:quad_double_examples}
\end{figure}

To provide a controllable, synthetic test bed for evaluating composition in a visual environment, we developed a simple semi-synthetic benchmark, visualized in Figures~\ref{fig:quad_single_examples}--\ref{fig:quad_double_examples}.
\texttt{quad} is a visual environment defined by 8 concepts:
\begin{enumerate}[itemsep=0pt]
	\item \texttt{quad1}: The colour of the first quadrant;
	\item \texttt{quad2}: The colour of the second quadrant;
	\item \texttt{quad3}: The colour of the third quadrant;
	\item \texttt{quad4}: The colour of the fourth quadrant;
	\item \texttt{size}: The size of the center object;
	\item \texttt{orientation}: The orientation (angle) of the center object;
	\item \texttt{object}: The colour of the center object;
	\item \texttt{shape}: The shape of the center object (circle, square, pill, triangle).
\end{enumerate}
With the exception of \texttt{shape}, which is discrete, the remaining concepts take values in $[0,1]$.
This environment has the following appealing properties:
\begin{itemize}
	\item Composing multiple concepts is straightforward and perceptually distinct;
	\item Concepts are continuous, which allows a variety of soft and hard interventions;
	\item Sampling is fast and easy, and requires only a few lines of Python code and no external software;
	\item Creating OOD hold-out validation datasets of any size is straightforward;
	\item This can be simulated on any $n\times n$ grid (as long as $n$ is large enough to distinguish different shapes; $n\ge 16$ is large enough in practice).
	      In our experiments, we used $n=64$.
\end{itemize}

These properties make \texttt{quad} especially useful for both quantitative and qualitative evaluation; indeed, it was created precisely for this reason.
Example images are shown in Figures~\ref{fig:quad_single_examples}-\ref{fig:quad_double_examples}.

To describe the experimental setting used in the \texttt{quad} (independent) experiments, let $\bc_{j}$ indicate the $j$th concept as listed above.
We used seven different contexts, including an observational context and one context each for single-concept interventions on $(\bc_{1},\ldots,\bc_{6})$, i.e., \texttt{quad1}, \texttt{quad2}, \texttt{quad3}, \texttt{quad4}, \texttt{size}, and \texttt{orientation}.
We use a sample size of 50\,000 images per context.
In each context, $\bc_{7}$ (\texttt{object}) and $\bc_{8}$ (\texttt{shape}) are sampled uniformly at random.
The remaining six concepts were sampled as follows:
\begin{itemize}
	\item The observational context was generated by randomly sampling $(\bc_{1},\ldots,\bc_{6})$ independently from the range $[0,0.5]$,
	\item The interventional contexts were generated by isolating a particular concept and sampling it uniformly from $[0.5,1]$, while sampling the rest from $[0,0.5]$.
\end{itemize}
For example, colours in $[0,0.5]$ include greens, yellows, and reds, and colours in $[0.5,1]$ include blues, purples, and magenta.
Thus, the object colour takes on any value in every context, whereas the four quadrant colours are restricted in the observational context.
When we intervene on a particular quadrant, we see that the colour palette noticeably shifts from greens to blues (Figure~\ref{fig:quad_single_examples}).

Similarly, we can compose multiple interventions together, as in Figure~\ref{fig:quad_double_examples}.
This facilitates evaluation of the concept composition capabilities of models trained on single interventions, providing a ground truth distribution to compare against generated OOD samples like in Figure~\ref{fig:quad-ood-full}.

We also made use of the easy-to-generate and perceptually distinct concept compositions in this dateset by evaluating models on held-out contexts (each containing 10\,000 images) with double-concept interventions, including \texttt{quad2\_quad3}, \texttt{quad2\_quad4}, \texttt{quad2\_size}, \texttt{quad1\_quad2}, \texttt{quad1\_quad3}, \texttt{quad1\_quad4}, and \texttt{quad1\_size}.
Examples can be seen in Figure~\ref{fig:quad_double_examples}.

For comparison, see Figure~\ref{app:ood-generation:-eye} for examples of generated multi-concept compositions (OOD) from a simple 3-layer convolutional model.
The training data did not contain any multi-concept compositions.

For \texttt{quad} dependent, a random SEM is generated using the \texttt{sempler} package~\citep{gamella2022characterization}, which also allows for data generation under shift interventions.

\section{Experimental details}
\label{app:addit-exper}

In this appendix we provide additional details on our experimental protocol and set-up.
We have used Snakemake~\citep{molder2025sustainable} for reproducibility.

\subsection{Evaluation metrics}
\label{app:evaluation-metrics}

We evaluated each model on the following metrics:
\begin{itemize}
	\item \textbf{Validation ELBO loss}: This is the standard ELBO loss used for validating VAEs.
	      We report this in bits per dimension (bpd), which is the loss scaled by a factor of \(\frac{1}{\mathrm{num\_pixels}\cdot\ln 2}\).
	\item \textbf{\(p\)-sliced Wasserstein (SW) distance}: Unlike the loss above, which evaluates \emph{individual reconstructed} images with respect to their corresponding ground truth images, this metric~\citep{bonneel2015sliced} evaluates a \emph{generated distribution} $\mu_{\textrm{gen}}$ with respect to a ground truth distribution $\mu$.
	      Specifically, we use the implementation of \citet{flamary2021pot} to compute a Monte-Carlo approximation of $\mathcal{SWD}_p(\mu_{\textrm{gen}}, \mu)$, where
	      \begin{equation}
		      \label{eq:sw}
		      \mathcal{SWD}_p(\mu, \nu) = \underset{\theta \sim \mathcal{U}(\mathbb{S}^{d-1})}{\mathbb{E}}\left(\mathcal{W}_p^p(\theta_\# \mu, \theta_\# \nu)\right)^{\frac{1}{p}},
	      \end{equation}
	      and \(\theta_\# \mu\) stands for the pushforwards of the projection \(X \in \mathbb{R}^d \mapsto \langle \theta, X \rangle\) and \(p=2\).
\end{itemize}

When reporting values for these metrics in tables, we show the mean, averaged over different runs/seeds, followed by the standard error.
We generally use a 70/30 training/validation split.

\subsection{Datasets}

In addition to the \texttt{quad} benchmark introduced in Appendix~\ref{app:quad}, we used the following benchmark datasets.

\paragraph{3DIdent}
\label{app:3dident}

The 3DIdent dataset~\citep{zimmermann2021contrastive} consists of a 3D object, rendered using the  Blender rendering engine, under different lighting conditions and orientations.
We generated 50\,000 samples per context corresponding to single-node interventions on background colour, spotlight colour, and object colour, in addition to the observational context.

\paragraph{MNIST}
\label{app:mnist}

We use a variation of the classic MNIST dataset~\citep{lecun1998gradient} known as Morpho-MNIST~\citep{castro2019morpho} along with additional affine transformations of MNIST (e.g., as in \citet{simard2003best}).
In addition to the observational context (standard MNIST images), the dataset contains the five contexts corresponding to single-concept interventions on concepts \texttt{scaled}, \texttt{shear}, \texttt{swel}, \texttt{thic}, and \texttt{thin}.
The datasets contain 60\,000 images per context.

\subsection{Architectures}
\label{app:arch}
We tested our context module end-to-end using two black-box models:
\begin{enumerate}
	\item (lightweight) \(\beta\)-VAE: A standard 3-layer convolutional network with latent dimension \(\dim(\mathbf{z})=128\).
	      This is used for the \texttt{quad} experiments in Section~\ref{sec:empirical-evaluation} and the \texttt{quad} and MNIST experiments in Section~\ref{app:lw-vae}.
	\item NVAE: A deep hierarchical VAE; see \cite{vahdat2020nvae} for details.
	      This is used for the MNIST and 3DIdent experiments in Sections~\ref{sec:empirical-evaluation} and \ref{app:nvae}.
\end{enumerate}
\noindent
The lightweight-VAE was used for comprehensive ablations and experiments on simple datasets, totaling more than 350 models overall, with 5 per column per table in Appendix~\ref{app:lw-vae}.
Since NVAE requires substantially more resources and time to train, this was used to train 12 total models, shown in Table~\ref{tab:one-table-to-rule-them-all}, the three CM variations plus three ablations on both MNIST and 3DIdent.

\subsection{Additional ablations}

We conducted additional ablations to stress test our model and isolate the effect of the context module.
This included additional experiments on different models, regularization, and expressivity.

\paragraph{Models} We ran ablations over different base VAEs augmented with the context module, denoted CM (end-to-end)---this is our module attached to the black-box, making full use of the different contexts, trained end-to-end.
To evaluate the proposed module, we compared its performance to three natural baselines:

\begin{enumerate}
	\item \textbf{\ablation}: CM (end-to-end) but trained with a single context on the full \emph{pooled} dataset---this is our module attached to the black-box given all the data but not making use of context-specific information, providing an ablation that circumvents our intervention layer (Section~\ref{sec:details}) while maintaining the same architectural complexity.
	\item \textbf{base}: Base VAE (without the context module) trained only on a single observational context---this is the black-box when denied all (non-observational) context information and data.
	\item \textbf{base pooled}: Base VAE (without the context module) trained with a single context on the full pooled dataset---this is the black-box when given all data but no explicit context information.
\end{enumerate}

Results for \(\beta\)VAE as the base applied to \texttt{quad} and MNIST are shown in Section~\ref{sec:empirical-evaluation} and Appendix~\ref{app:lw-vae}.
Results for NVAE as the base applied to 3DIdent are shown in Section~\ref{sec:empirical-evaluation}.

\paragraph{Regularization} We run CM (end-to-end) with \(\beta\)-VAE base on MNIST, independently trying out two different regularizers, varying the regularization weight \(\lambda\) for each: group lasso and \(\ell_2\).
These are applied to the weights of the reduced-form SEM \(A_0\) in \ref{eq:concept:sem}.
We also try varying \(\beta\) (as in \(\beta\)-VAE).
Results are shown in Appendix~\ref{app:mnist-diff-reg}.

\paragraph{Expressivity} We run CM (end-to-end) on MNIST, jointly varying three parameters that control expressiveness of our context module:
expressive input width \(\width_{\mathrm{exp}}\) (which controls latent dimension \(\dim(\mathbf{z})\)),
depth of the expressive layer \(h_{\mathrm{exp}}\), and
concept width ($\width_c=\dim(\bc_{j}) = \width_{\eps}=\dim(\beps_{j})$ as in Section~\ref{sec:details}).
Results are shown in Appendix~\ref{app:vary-expr-conc}.

\subsection{Additional baselines}
\label{sec:additional-baselines}

In addition to the above base models and ablations, we include comparisons against three disentanglement methods, including the unsupervised \(\beta\)-TCVAE~\citep{chen2018isolating}, the weakly supervised Ada-GVAE~\citep{adagvae}, and the supervised TVAE~\citep{tvae}.
We gratefully acknowledge the open source implementation of these within the \texttt{disent} Python package~\citep{Michlo2021Disent}.
In addition to comparisons against the \texttt{quad} ablations, these baselines are run on MNIST and 3DIdent in Appendix~\ref{app:mnist-ablations}.

\section{Additional results}
\label{app:additional-results}

In this section, we present complete results, including experimental settings described in Appendix~\ref{app:addit-exper} for our context module with the lightweight-VAE black-box architecture as well as additional results on MNIST and 3DIdent for our context module with NVAE.

\subsection{lightweight-VAE}
\label{app:lw-vae}

\subsubsection{Ablations (MNIST and \texttt{quad})}
\label{app:mnist-ablations}

A description of the architecture and experiment is given in Appendix~\ref{app:addit-exper}.
Tables~\ref{tab:mnist-ablation}--\ref{tab:3didnt-disent} are already summarized in the main text in Table~\ref{tab:one-table-to-rule-them-all}, but here we use the extra space here to additionally present standard errors and show a breakdown of the results over the different contexts.

\TableMnistAblation
\TableQuadAblation
\TableQuadDependentAblation
\TableThreeDIdentDisent

\clearpage
\subsubsection{Examples of concept composition (\texttt{quad})}
\label{app:ood-generation:-eye}

Figure~\ref{fig:quad-ood-full} depicts samples generated from the context module appended to the lightweight-VAE described in Appendix~\ref{app:arch}.

\begin{figure}[ht]
	\scriptsize
	\floatconts
	{fig:quad-ood-full}
	{\caption{Example generated OOD images from a run of \texttt{quad}.}}
	{%
		\subfigure{%
			\includegraphics[width=0.42\linewidth]{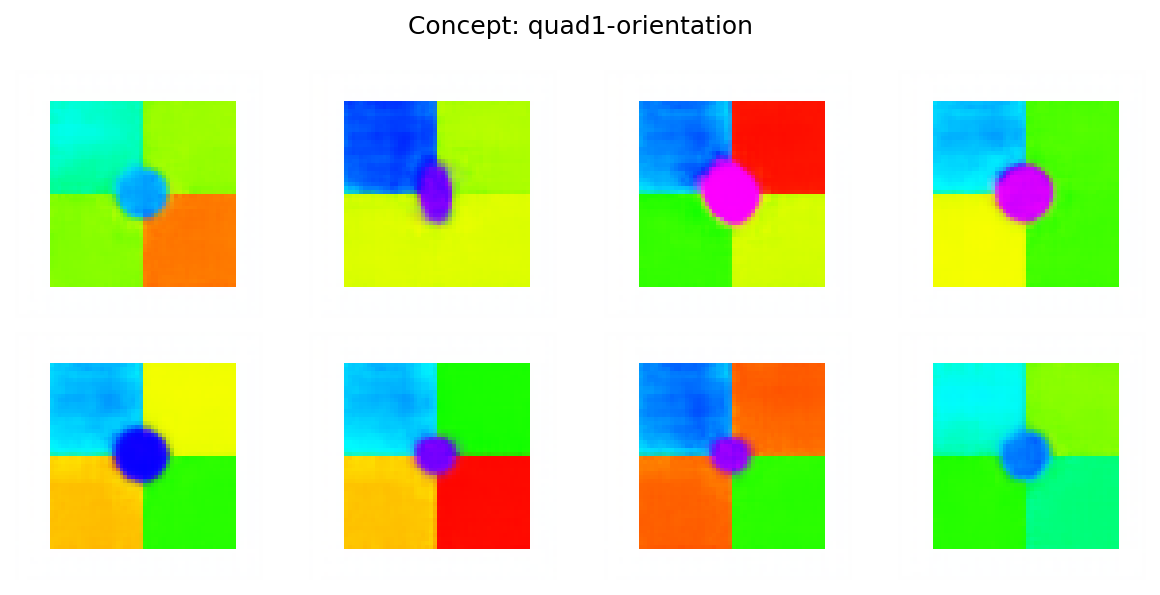}%
		}%
		\hfill
		\subfigure{%
			\includegraphics[width=0.42\linewidth]{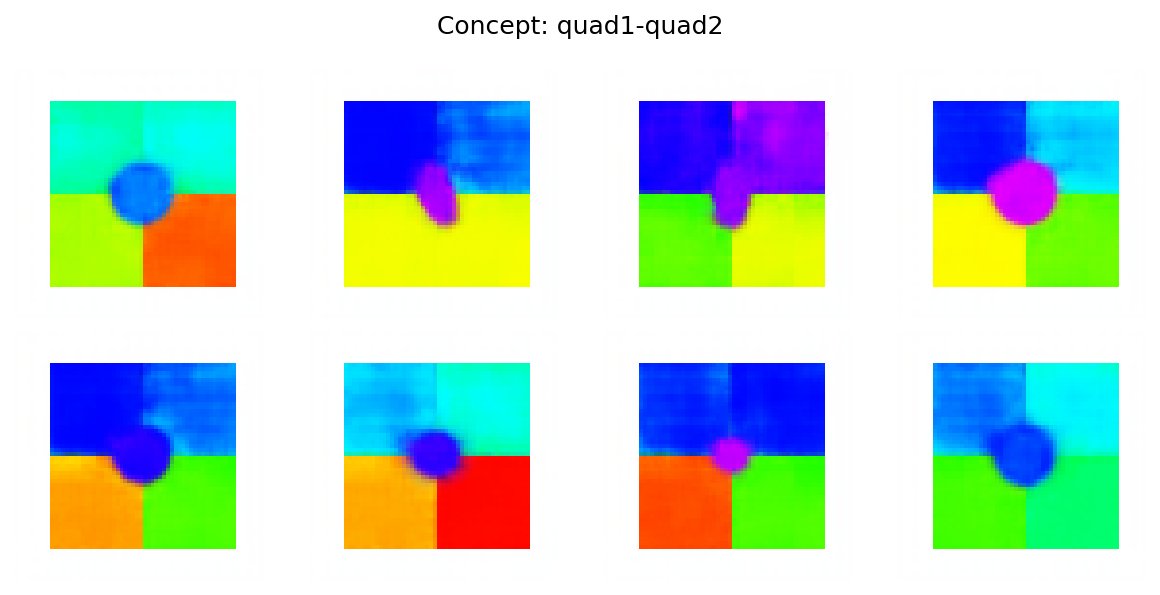}%
		}\\
		\subfigure{%
			\includegraphics[width=0.42\linewidth]{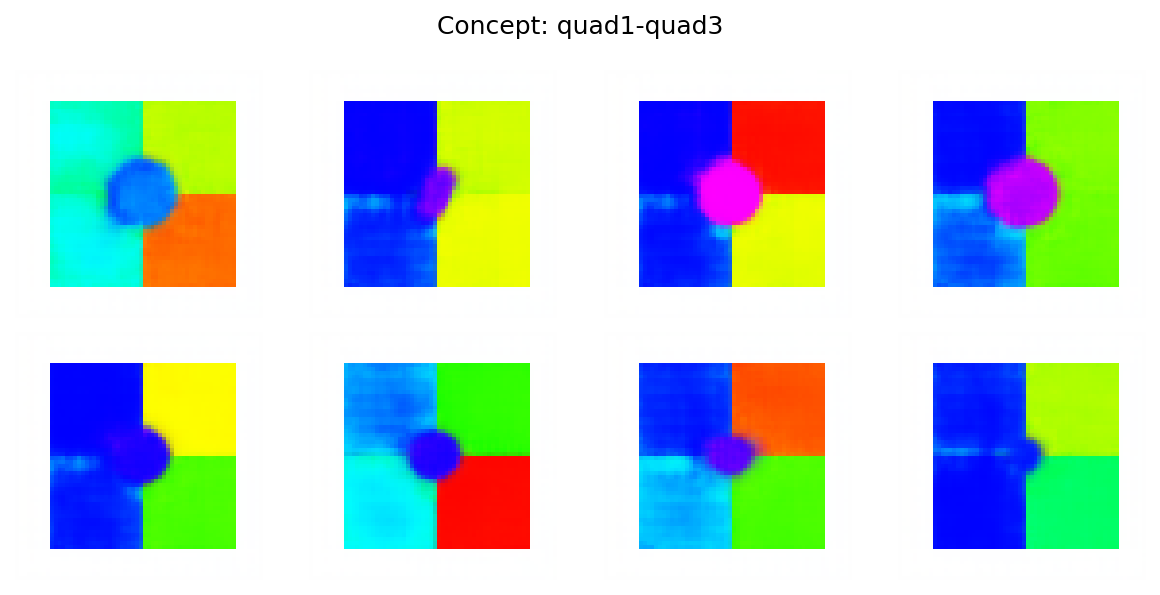}%
		}%
		\hfill
		\subfigure{%
			\includegraphics[width=0.42\linewidth]{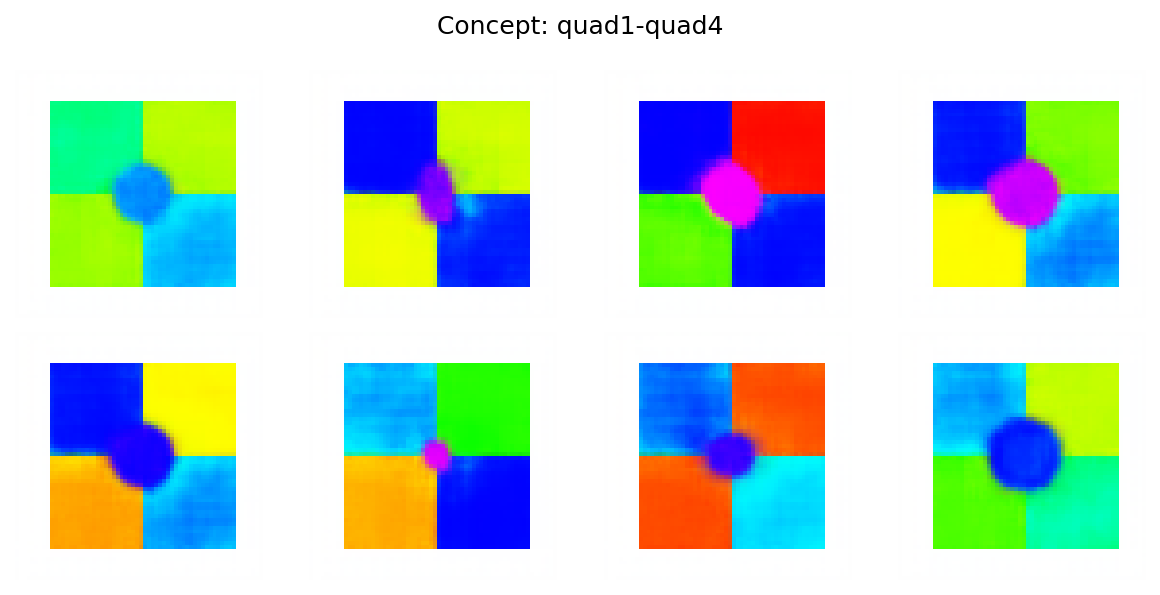}%
		}\\
		\subfigure{%
			\includegraphics[width=0.42\linewidth]{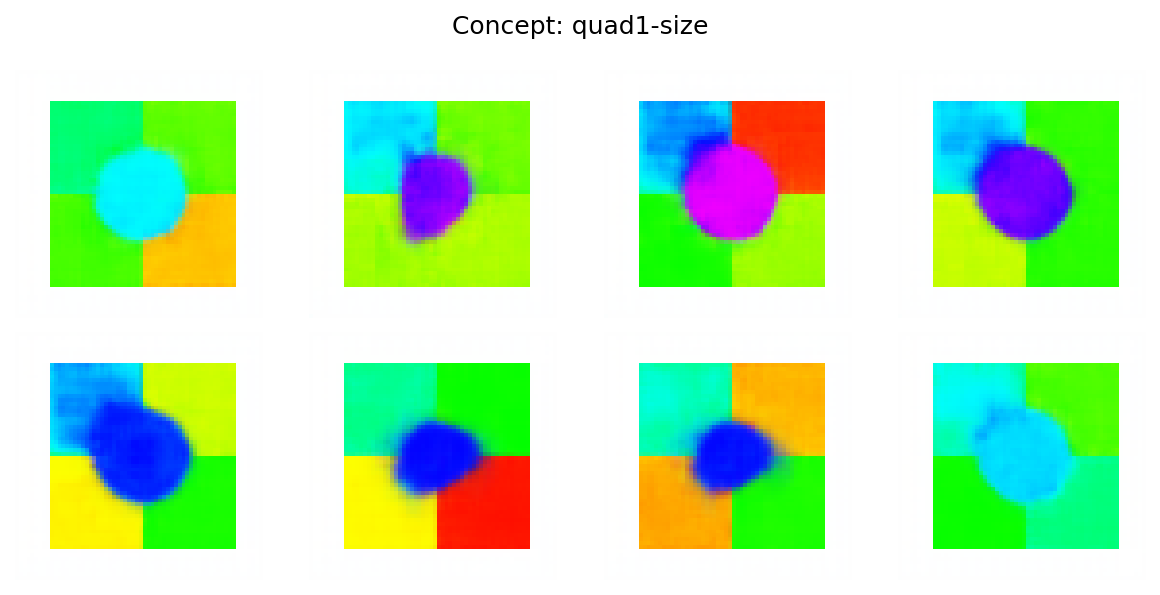}%
		}%
		\hfill
		\subfigure{%
			\includegraphics[width=0.42\linewidth]{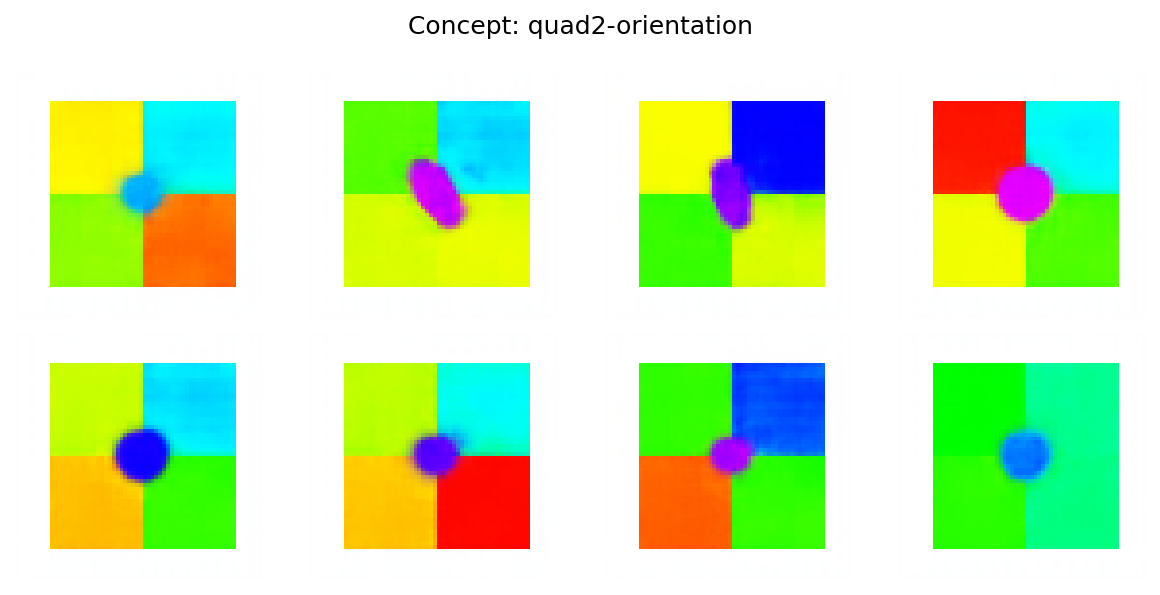}%
		}\\
		\subfigure{%
			\includegraphics[width=0.42\linewidth]{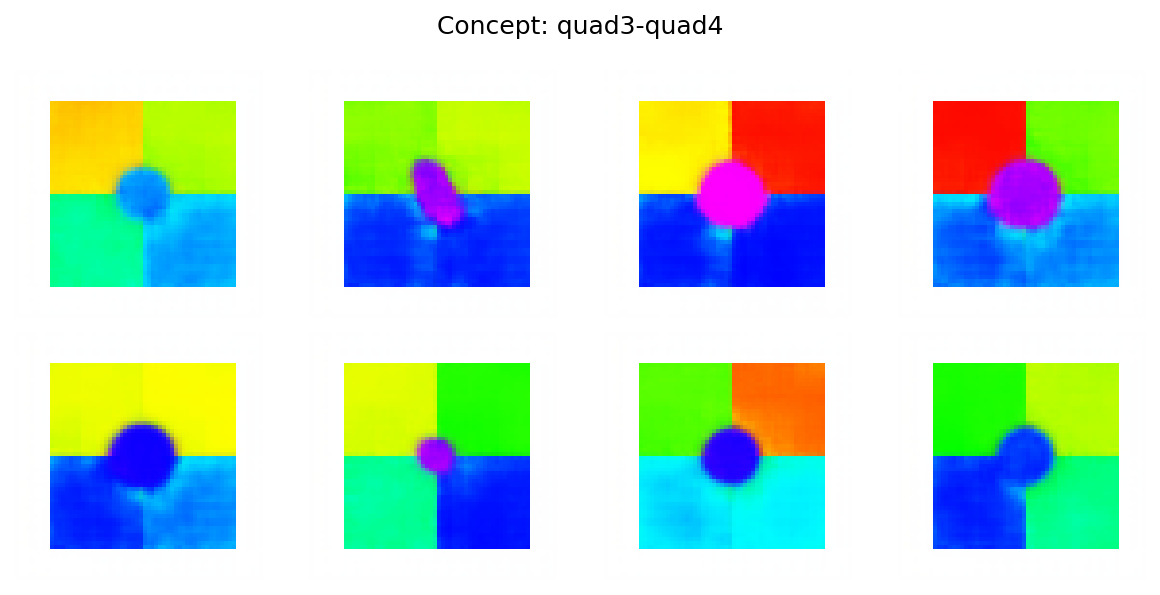}%
		}%
		\hfill
		\subfigure{%
			\includegraphics[width=0.42\linewidth]{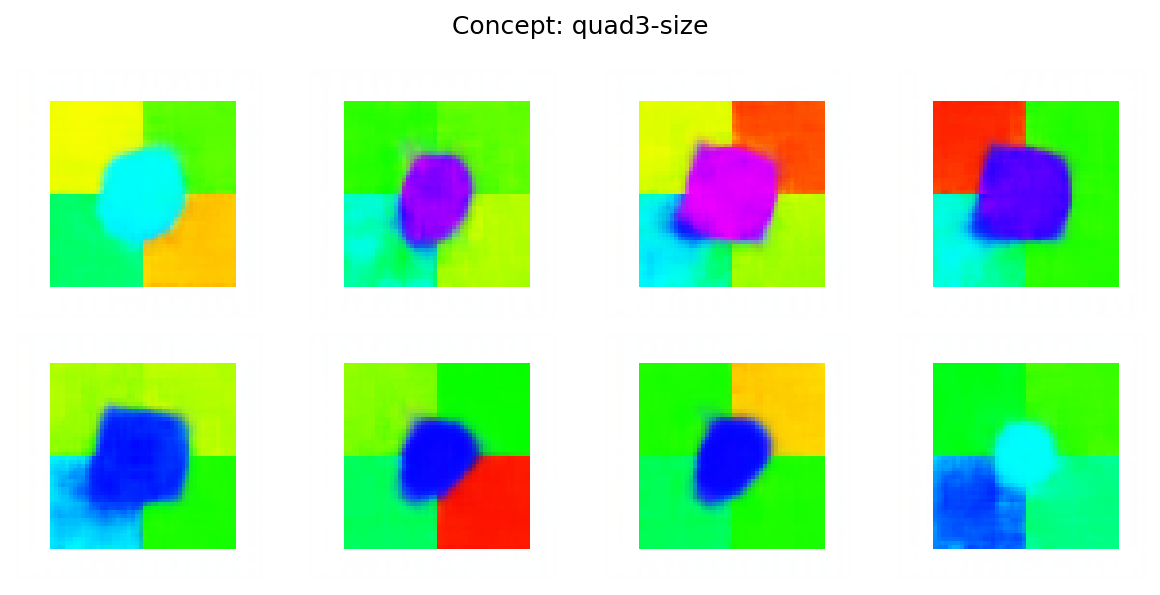}%
		}\\
		\subfigure{%
			\includegraphics[width=0.42\linewidth]{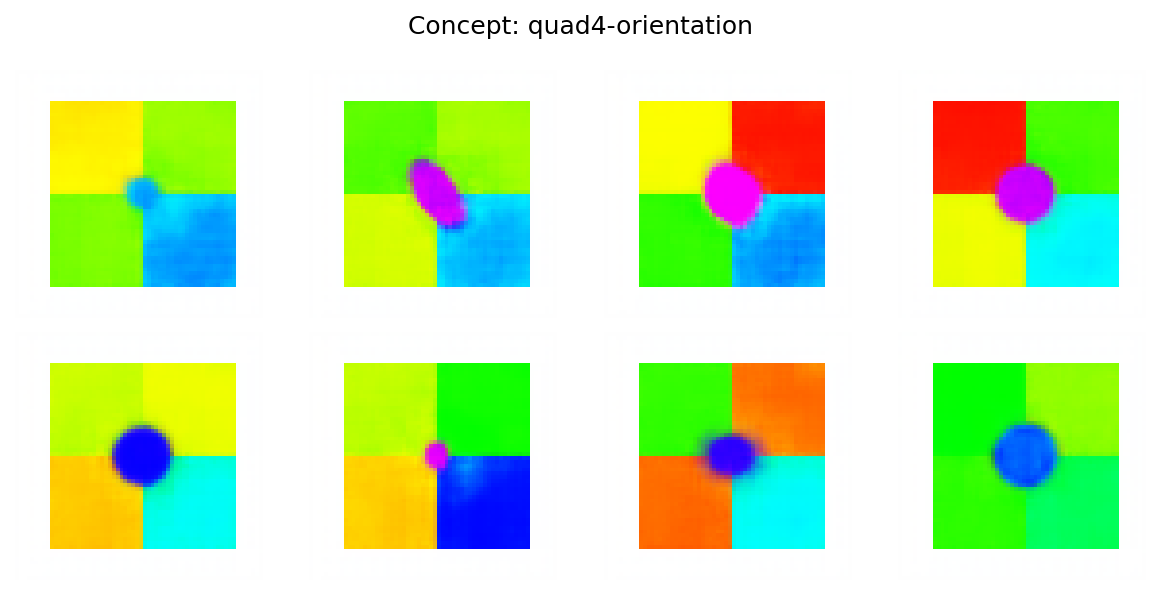}%
		}%
		\hfill
		\subfigure{%
			\includegraphics[width=0.42\linewidth]{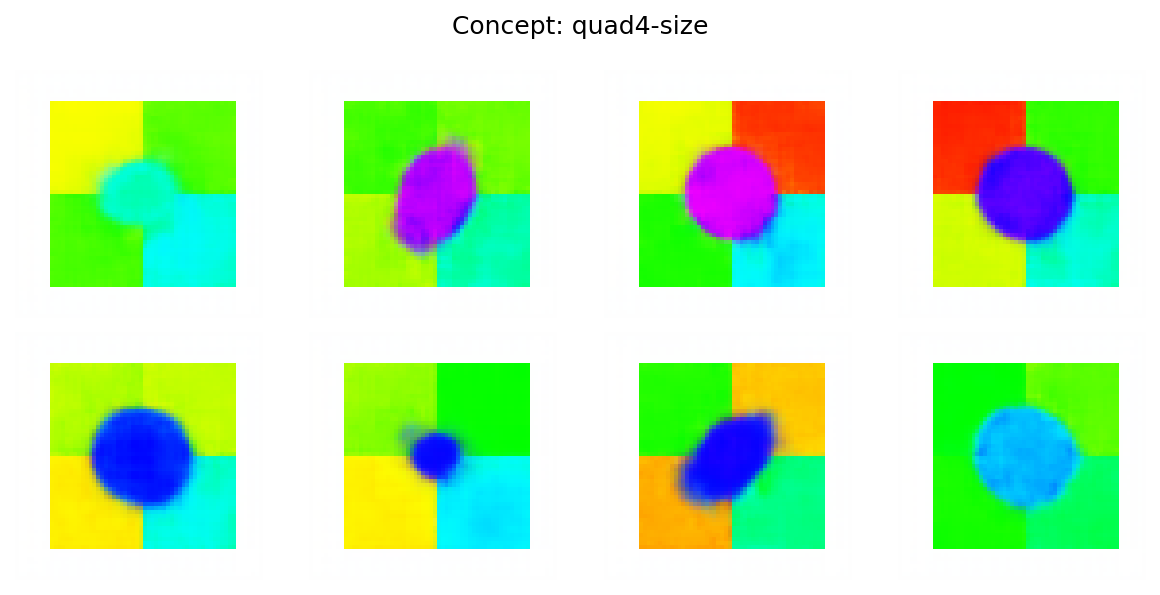}%
		}%
	}
\end{figure}
\clearpage

\subsubsection{Regularization (MNIST)}
\label{app:mnist-diff-reg}

A description of the architecture and experiment is given in Appendix~\ref{app:addit-exper}.
Tables~\ref{tab:mnist-groupnorm}~and~\ref{tab:mnist-l2norm} respectively show results of group lasso \citep{yuan2006model} and \(L_2\) regularization, while Table~\ref{tab:mnist-beta} shows the effects regularizing the latent space of the VAE by varying \(\beta\) \citep{higgins2017beta}.\footnote{All other MNIST experiments in Appendix~\ref{app:lw-vae} are run with \(\lambda=0\) for group lasso and \(L_2\) and with \(\beta=1\).}
Altogether, the results are relatively stable performance across regularization strengths, with slight increase in reconstruction performance for higher group lasso or \(L_2\) regularization strength but no other statistically significant differences.

\vspace{2em}
\TableMnistGroupNorm
\TableMnistEllTwoNorm
\TableMnistBeta
\clearpage

\subsubsection{Regularization (\texttt{quad})}
\label{app:quad-diff-reg}

This is like Appendix~\ref{app:mnist-diff-reg} but applied to the \texttt{quad} dataset rather than MNIST.
A description of the architecture and experiment is given in Appendix~\ref{app:addit-exper}.
Tables~\ref{tab:quad-groupnorm}~and~\ref{tab:quad-l2norm} respectively show results of group lasso \citep{yuan2006model} and \(L_2\) regularization, while Table~\ref{tab:quad-beta} shows the effects regularizing the latent space of the VAE by varying \(\beta\) \citep{higgins2017beta}.\footnote{All other \texttt{quad} experiments in Appendix~\ref{app:lw-vae} are run with \(\lambda=0\) for group lasso and \(L_2\) and with \(\beta=1\).}

\vspace{2em}
\TableQuadGroupNorm
\TableQuadEllTwoNorm
\TableQuadBeta
\clearpage

\subsubsection{Expressivity (MNIST)}
\label{app:vary-expr-conc}

A description of the architecture and experiment is given in Appendix~\ref{app:addit-exper}.
Tables~\ref{tab:mnist-capacity-2}~and~\ref{tab:mnist-capacity-4} show varying widths for an expressive layer depth respectively of 2 and 5: the tuples in the column headers indicate \((\width_{\mathrm{exp}},\ \width_c)\).\footnote{All other experiments on MNIST in Appendix~\ref{app:lw-vae} are run with \(h_{\mathrm{exp}}=2\), \(\width_{\mathrm{exp}}=15\), and \(\width_c=5\).}
These results suggest relatively stable results across different configurations for concept learning while the best reconstruction performance is obtained from a moderate bottleneck, with expressive layer input width \(\width_{\mathrm{exp}}=22\) and a concept width of \(\width_c=5\).

\vspace{2em}
\TableMnistCapacityTwo
\vspace{2em}
\TableMnistCapacityFour
\clearpage

\subsubsection{Expressivity (\texttt{quad})}
\label{app:vary-expr-conc-quad}

This is like Appendix~\ref{app:vary-expr-conc} but applied to the \texttt{quad} dataset rather than MNIST.
A description of the architecture and experiment is given in Appendix~\ref{app:addit-exper}.
Tables~\ref{tab:quad-capacity-2}~and~\ref{tab:quad-capacity-4} show varying widths for an expressive layer depth respectively of 2 and 5: the tuples in the column headers indicate \((\width_{\mathrm{exp}},\ \width_c)\).\footnote{All other experiments on \texttt{quad} in Appendix~\ref{app:lw-vae} are run with \(h_{\mathrm{exp}}=2\), \(\width_{\mathrm{exp}}=32\), and \(\width_c=8\).}
These results suggest relatively stable results across different configurations.

\vspace{2em}
\TableQuadCapacityTwo
\TableQuadCapacityFour
\clearpage

\subsection{Additional NVAE Results}
\label{app:nvae}

\subsubsection{Additional examples of concept composition}

\begin{figure}[!ht]
	\centering
	\includegraphics[height=0.3\linewidth]{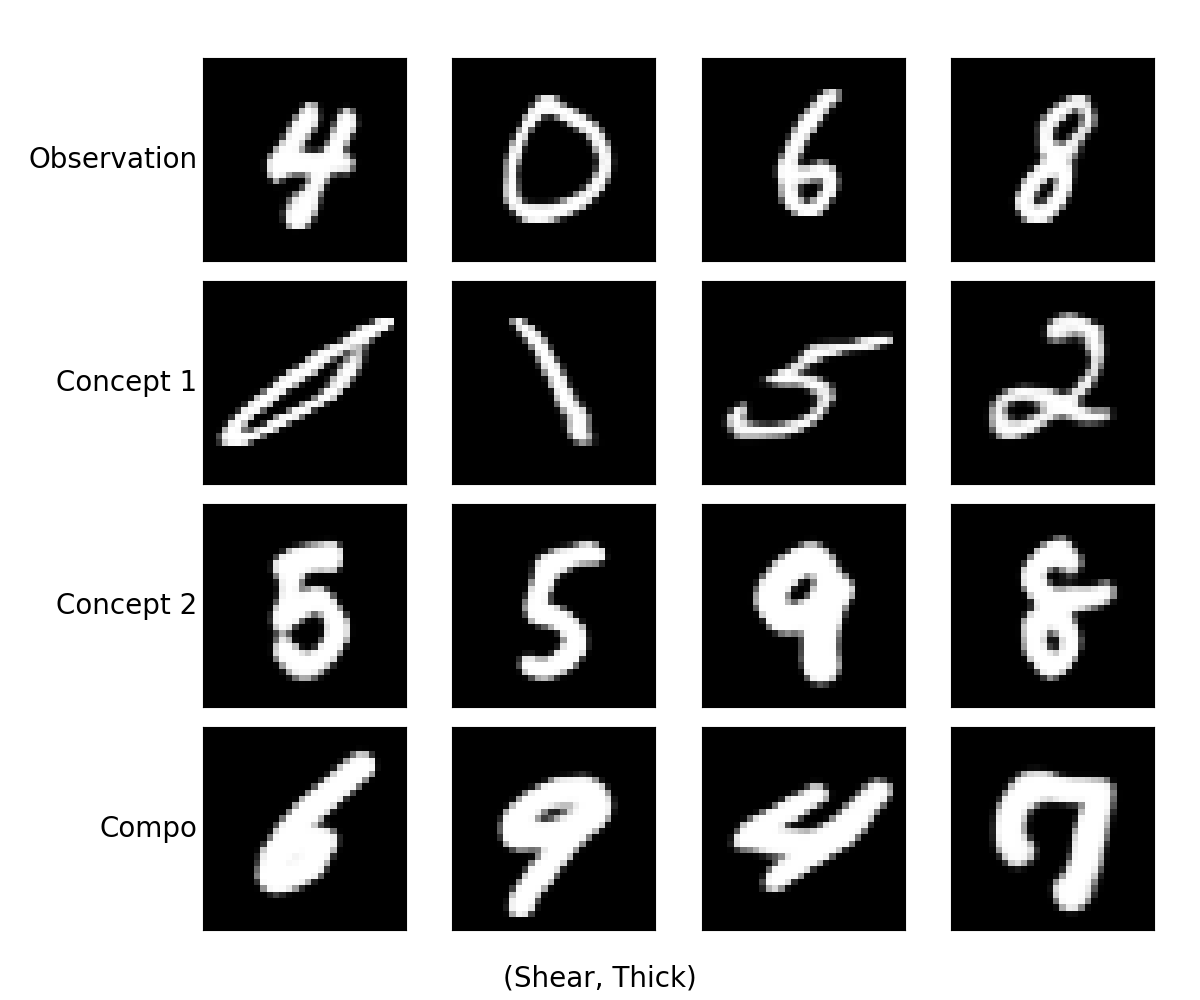}
	\includegraphics[height=0.3\linewidth]{nvae/3DIdent/combo_comparison/background-object-new2}\\
	\includegraphics[height=0.3\linewidth]{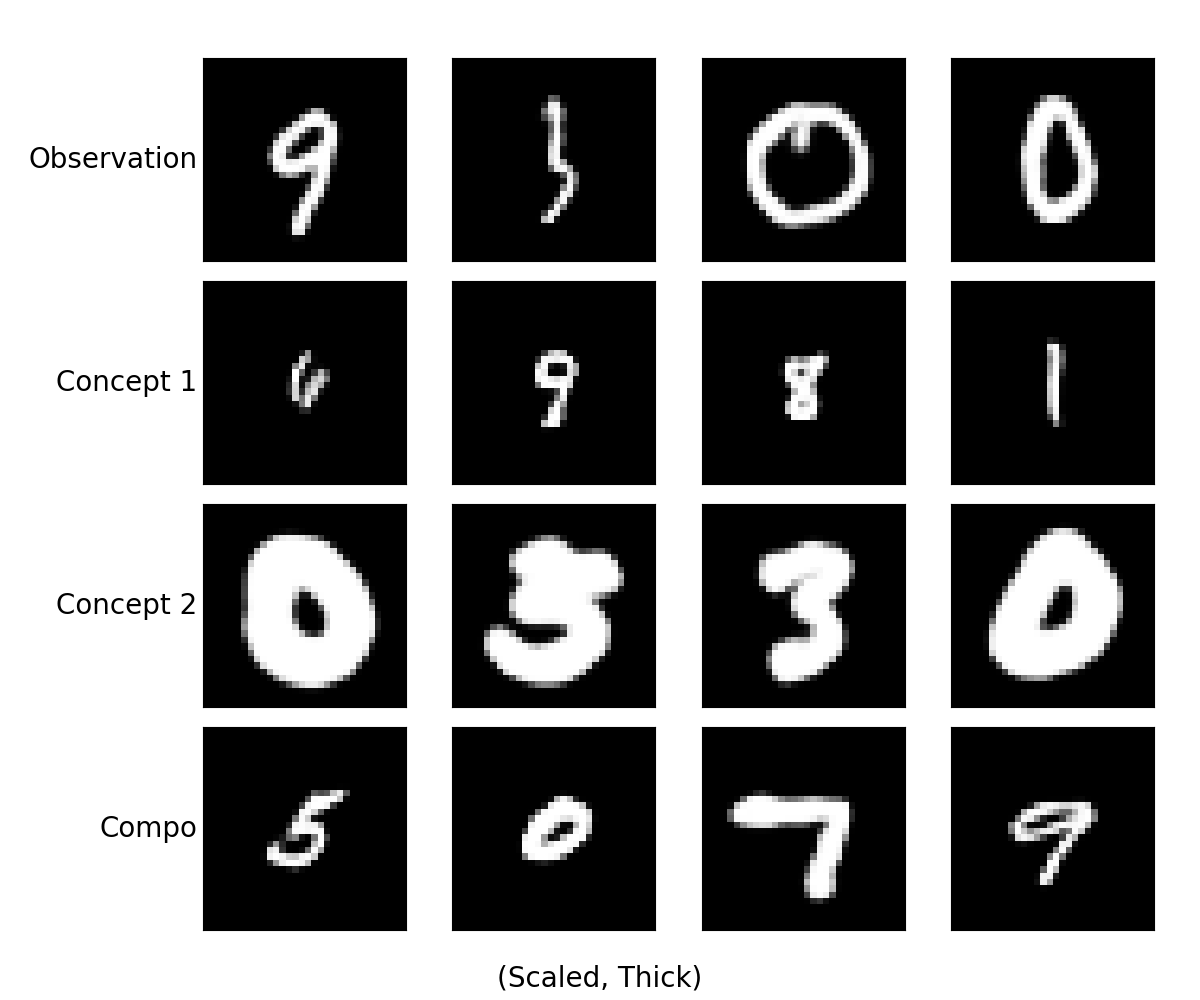}
	\includegraphics[height=0.3\linewidth]{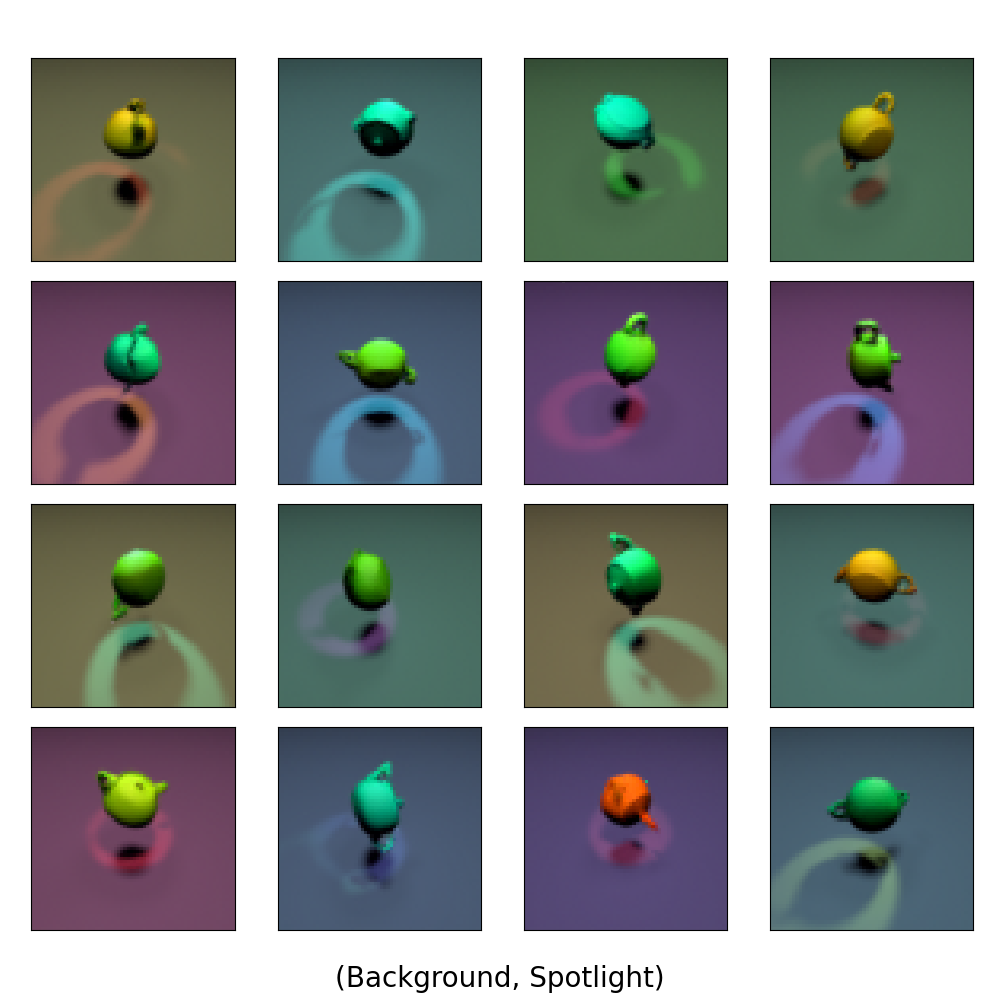}\\
	\includegraphics[height=0.3\linewidth]{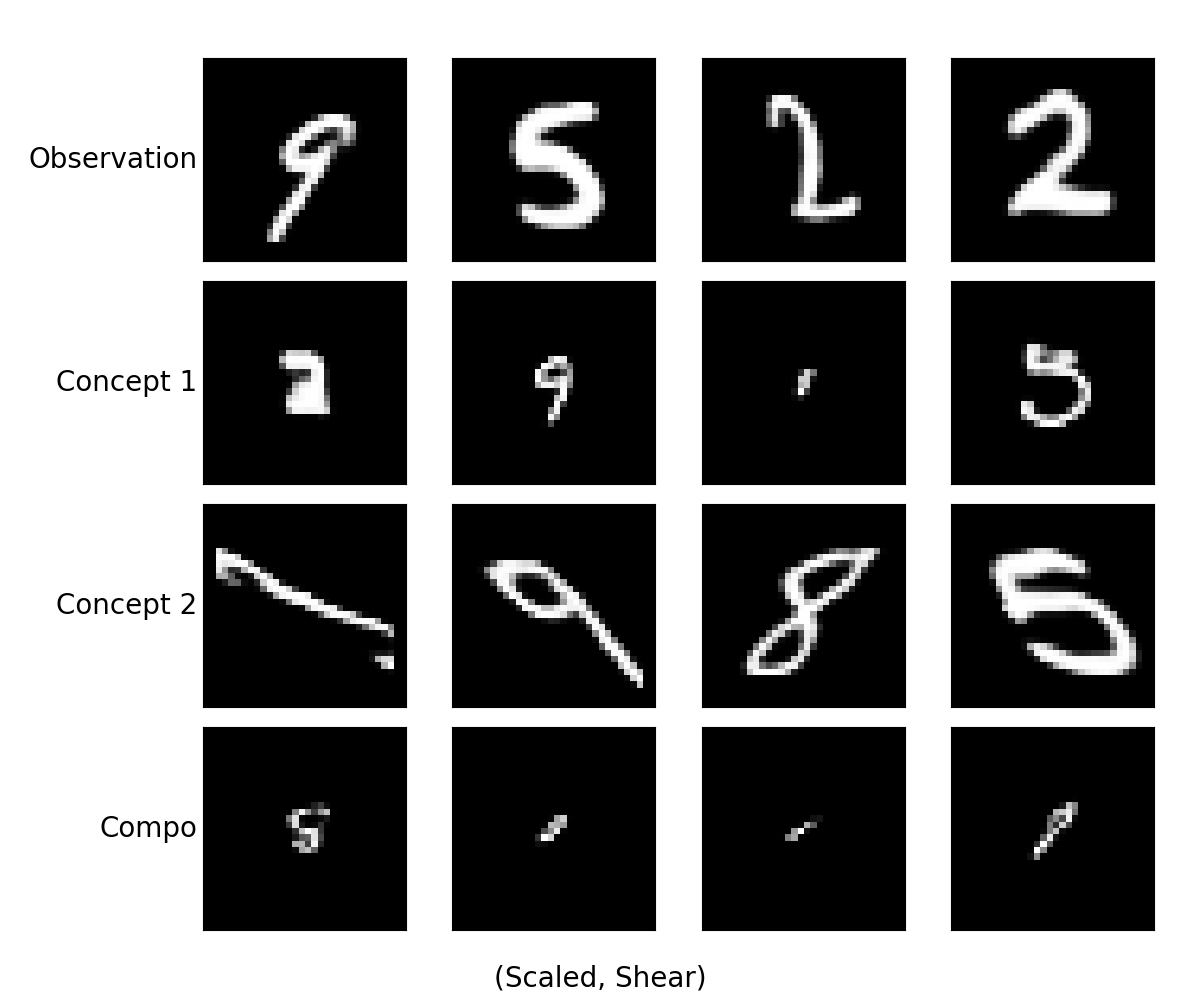}
	\includegraphics[height=0.3\linewidth]{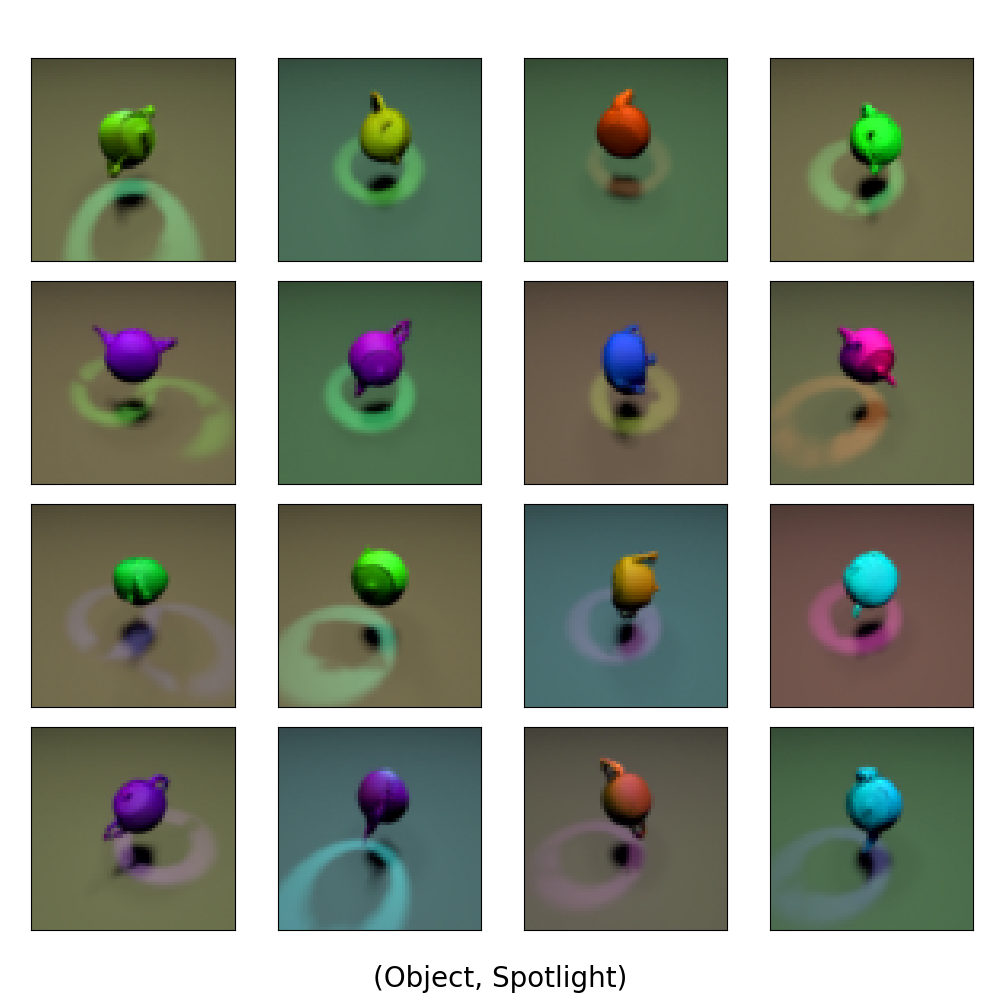}\\
	\caption{Additional examples of concept composition in MNIST (left) and 3DIdent (right).}
	\label{fig:app_combo_comparison}
\end{figure}

Figure~\ref{fig:app_combo_comparison} contains additional examples of composing concepts from the NVAE-based context module.
Each row corresponds to actual generated samples from the trained model in different contexts:
\begin{enumerate}[itemsep=0pt]
	\item The first row shows generated observational samples;
	\item The second and third row contain generated single-node interventions;
	\item The final row shows generated samples where two distinct concepts are composed together.
\end{enumerate}
The final row of compositional generations is genuinely OOD, in that the training data does not contain any examples where both concepts have been intervened upon.

Additional examples of both concept learning (sampling after single-target interventions) and composition (sampling after multi-target interventions) are provided in Appendices~\ref{app:addit-mnist-figur}--\ref{app:addit-3did-figur}.

\vspace{0.5em}

\clearpage
\subsubsection{Additional MNIST figures}
\label{app:addit-mnist-figur}
\begin{figure}[!ht]
	\centering
	\includegraphics[height=0.3\linewidth]{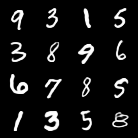}
	\includegraphics[height=0.3\linewidth]{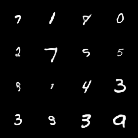}
	\includegraphics[height=0.3\linewidth]{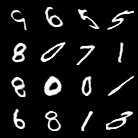}
	\includegraphics[height=0.3\linewidth]{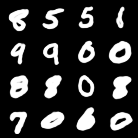}
	\includegraphics[height=0.3\linewidth]{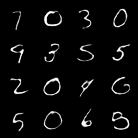}
	\includegraphics[height=0.3\linewidth]{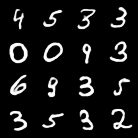}
	\caption{Examples of learned concepts (sampling after single-target intervention) in MNIST.
		(top) observational, scaled, shear (bottom) thic, thin, swel.}
	\label{fig:mnist_samples}
\end{figure}

\begin{figure}[!ht]
	\centering
	\includegraphics[height=0.24\linewidth]{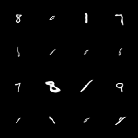}
	\includegraphics[height=0.24\linewidth]{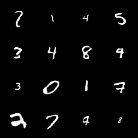}
	\includegraphics[height=0.24\linewidth]{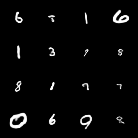}
	\includegraphics[height=0.24\linewidth]{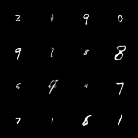}
	\includegraphics[height=0.24\linewidth]{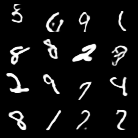}
	\includegraphics[height=0.24\linewidth]{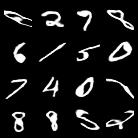}
	\includegraphics[height=0.24\linewidth]{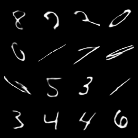}
	\includegraphics[height=0.24\linewidth]{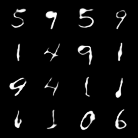}
	\caption{Examples of OOD concept composition (sampling after multi-target intervention) in MNIST.}
	\label{fig:mnist_sample_combo}
\end{figure}

See Figure~\ref{fig:mnist_samples} for concept learning and Figure~\ref{fig:mnist_sample_combo} for composition.
\vspace{1em}

\vspace{1em}

\clearpage
\subsubsection{Additional 3DIdent figures}
\label{app:addit-3did-figur}
\begin{figure}[!ht]
	\centering
	\includegraphics[height=0.3\linewidth]{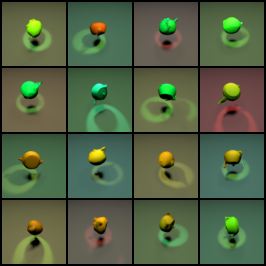}
	\includegraphics[height=0.3\linewidth]{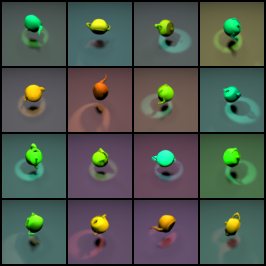}\\
	\includegraphics[height=0.3\linewidth]{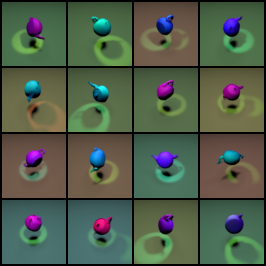}
	\includegraphics[height=0.3\linewidth]{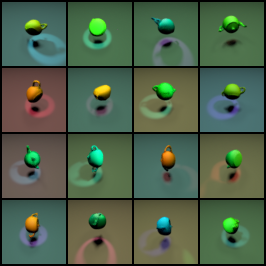}
	\caption{Examples of learned concepts (sampling after single-target intervention) in 3DIdent.
		(top) observational, background (bottom) object, spotlight.}
	\label{fig:3dident_samples}
\end{figure}

\begin{figure}[!ht]
	\centering
	\includegraphics[height=0.3\linewidth]{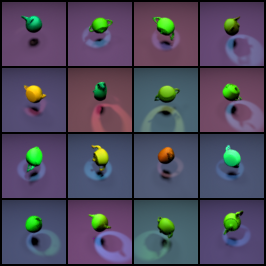}
	\includegraphics[height=0.3\linewidth]{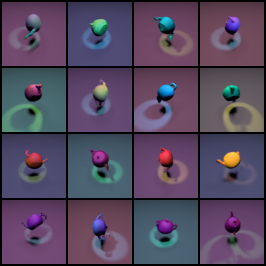}
	\includegraphics[height=0.3\linewidth]{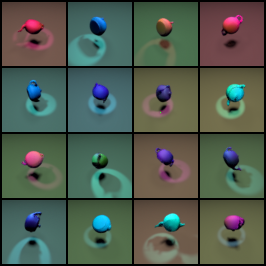}
	\caption{Examples of OOD concept composition (sampling after multi-target intervention) in 3DIdent.}
	\label{fig:3dident_sample_combo}
\end{figure}

See Figure~\ref{fig:3dident_samples} for concept learning and Figure~\ref{fig:3dident_sample_combo} for composition.
\vspace{2em}

\vspace{2em}

\end{document}